\documentclass{article}

\usepackage[preprint]{neurips_2026}
\makeatletter
\renewcommand{\@noticestring}{}
\makeatother

\usepackage[utf8]{inputenc} % allow utf-8 input
\usepackage[T1]{fontenc}    % use 8-bit T1 fonts
\usepackage{hyperref}       % hyperlinks
\usepackage{url}            % simple URL typesetting
\usepackage{booktabs}       % professional-quality tables
\usepackage[most]{tcolorbox}
\usepackage{fvextra}
\usepackage{amsfonts}       % blackboard math symbols
\usepackage{amssymb}
\usepackage{nicefrac}       % compact symbols for 1/2, etc.
\usepackage{microtype}      % microtypography
\usepackage{xcolor}         % colors
\usepackage{listings}
\usepackage{graphicx}
\usepackage{subcaption}
\usepackage{wrapfig}
\usepackage{multirow}
\usepackage{comment}
\usepackage{cleveref}
\usepackage{algorithm}
\usepackage{algpseudocode}
\usepackage{amsmath,amsfonts,bm}

\usepackage{xspace}
\usepackage{xcolor}         % colors
\definecolor{darkred}{RGB}{150,0,0}
\definecolor{darkgreen}{RGB}{0,150,0}
\definecolor{darkblue}{RGB}{0,0,200}

\def\eqref#1{equation~\ref{#1}}
\def\1{\bm{1}}

\DeclareMathAlphabet{\mathsfit}{\encodingdefault}{\sfdefault}{m}{sl}
\SetMathAlphabet{\mathsfit}{bold}{\encodingdefault}{\sfdefault}{bx}{n}

\usepackage{tikz}
\usetikzlibrary{patterns}
\usepackage{lineno}

\definecolor{darkblue}{rgb}{0, 0, 0.5}
\hypersetup{colorlinks=true, citecolor=darkblue, linkcolor=darkblue, urlcolor=darkblue}

\lstdefinestyle{pythoncode}{
    language=Python,
    basicstyle=\ttfamily\small,
    keywordstyle=\color{blue},
    commentstyle=\color{gray},
    stringstyle=\color{teal},
    showstringspaces=false,
    breaklines=true,
    frame=single,
    columns=fullflexible
}

\lstdefinestyle{prompttext}{
    language={},
    basicstyle=\ttfamily\small,
    keywordstyle=\color{black},
    commentstyle=\color{black},
    stringstyle=\color{black},
    identifierstyle=\color{black},
    showstringspaces=false,
    breaklines=true,
    frame=single,
    columns=fullflexible,
    keepspaces=true
}

\title{Evolutionary Multi-Task Optimization for LLM-Guided Program Discovery}

\author{%
  \quad Halil Alperen Gozeten$^1$\And 
  \quad Xuechen Zhang$^1$ \And 
  \quad\quad\quad M. Emrullah Ildiz$^1$\quad\quad\quad\And 
  \hspace*{10pt}Ege Onur Taga$^1$\And
  \hspace*{-105pt}Tara Javidi$^2$\\\\
  \hspace*{-80pt}1: University of Michigan - Ann Arbor\\
  \hspace*{-80pt}\texttt{\{alperen,zxuechen,eildiz,egetaga,oymak\}@umich.edu} \vspace{5pt}\\
 \hspace*{-80pt}2: University of California San Diego\\
  \hspace*{-80pt}\texttt{tara@ece.ucsd.edu} \And
  \hspace*{-85pt}Samet Oymak$^1$  
}

\makeatletter
\@ifpackageloaded{tcolorbox}{}{%
  \usepackage{tcolorbox} % no options -> avoids option clash
}
\makeatother
\tcbuselibrary{skins,breakable}

\definecolor{abstractgray}{RGB}{240,240,240}

\newlength{\AbsPad}
\newlength{\AbstractBoxWidth}
\RenewDocumentEnvironment{abstract}{+b}
{%
  \vskip 0.075in%
  \begin{center}
  {\large\bfseries Abstract}\par\vspace{1em}
  \begin{tcolorbox}[
    enhanced,
    colback=abstractgray,
    colframe=abstractgray,
    boxrule=0pt,
    arc=10pt,
    boxsep=0pt,
    left=\AbsPad,
    right=\AbsPad,
    top=12pt,
    bottom=12pt,
    width=\AbstractBoxWidth,
    before skip=0pt,
    after skip=0pt
  ]
  #1
  \end{tcolorbox}
  \end{center}
  \vskip 1ex%
}
{}

\begin{document}

\maketitle

\begingroup
\renewcommand\thefootnote{\fnsymbol{footnote}}
\footnotetext[1]{Code: \url{https://github.com/alperengozeten/emo}}
\endgroup

\vspace{-0.5cm}
\begin{abstract}
Recent LLM-guided evolutionary search methods have shown that iterative program mutation can discover strong algorithms, but they typically optimize each task independently, even when related tasks share reusable structure. We study evolutionary multitask optimization (EMO) for LLM-guided program discovery and propose EMO-STA (Shared-Then-Adapt), a two-stage framework that first evolves a shared archive of executable programs across a task family and then adapts selected shared candidates to each target task. Within EMO-STA, we explore multiple adaptation strategies, including warm-starting from the shared archive, adapting the best average shared program, and adapting the shared program that performs best on each target task. Across eight task families spanning continuous optimization, geometric construction, modeling, and algorithmic optimization, EMO-STA improves over single-task evolution under matched evolutionary-iteration budgets in most settings, with STA Best-Local providing the strongest in-distribution adaptation and STA Best-Shared yielding robust transfer to unseen tasks. Iteration-allocation experiments show that devoting a substantial fraction of the family-level iteration budget to shared evolution is effective, with roughly balanced shared and aggregate adaptation budgets often yielding the best performance. ARC and time-series feature-engineering case studies further suggest that shared evolution can mitigate overfitting in low-evidence settings by favoring programs that perform well across related tasks rather than exploiting brittle task-specific artifacts. 
\end{abstract}
%\vspace{-0.5cm}
\section{Introduction}\label{sec:intro}

Recent LLM-guided discovery methods such as FunSearch and AlphaEvolve have shown that evolutionary code search can discover new mathematical constructions and improve scientific or algorithmic procedures. Besides applications in scientific discovery, evolutionary search (ES) has major impact in a variety of domains including agentic AI systems. For instance, ES can be used for iteratively generating code until test cases pass, prompt optimization to maximize language model performance or, more generally, optimizing the prompts and information flow/graph of agentic systems.
However, the existing methods typically run evolution independently, focused on one target problem at a time, even when there are multiple tasks with structural similarities \citep{romera-paredes2024funsearch,novikov2025alphaevolve}. Importantly, evolutionary search with LLMs is an expensive process often requiring tens of rounds to achieve satisfactory results, which highlights a need for more efficient procedures~\citep{romera-paredes2024funsearch,novikov2025alphaevolve,fernando2023promptbreederselfreferentialselfimprovementprompt,agrawal2025gepa}. Although evolutionary multitasking provides a natural precedent for jointly optimizing related tasks, much of this literature focuses on shared-population search in a fixed representation space, such as unified decision-vector encodings or genetic-programming representations \citep{gupta2016multifactorial,scott2017multitask,cai2021hybrid}. 

This motivates a natural question for LLM-guided program discovery: how to conduct evolutionary search over LLM-generated executable programs across related tasks, while still allowing task-specific adaptation under each task's objective?

\begin{figure*}[t]
    \centering

    \begin{subfigure}[t]{0.76\textwidth}
        \centering
        \includegraphics[width=\linewidth]{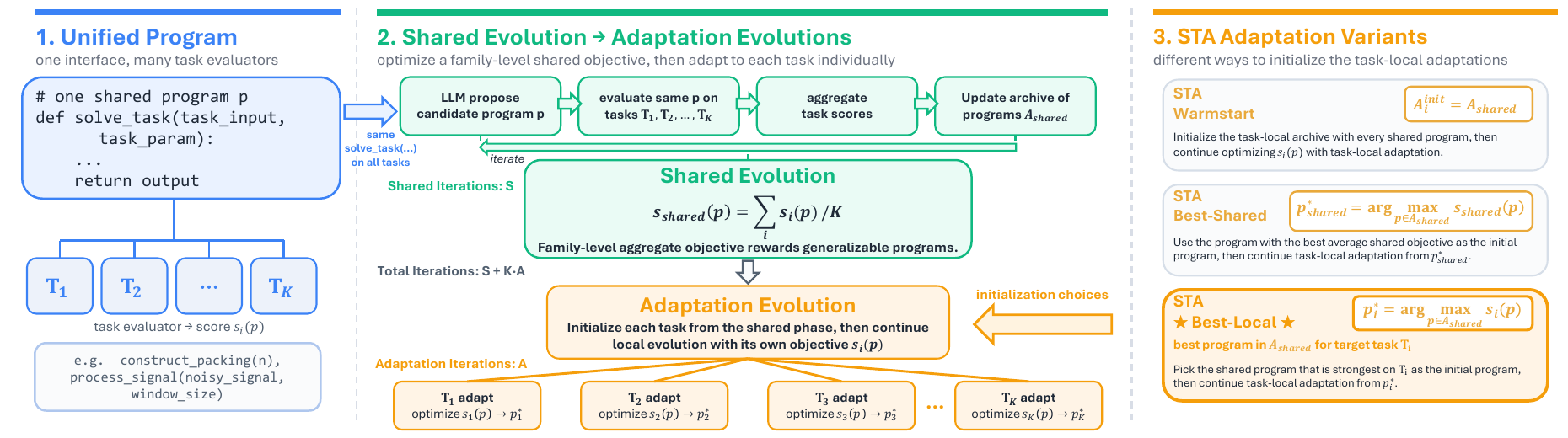}
        \caption{\small{\textbf{EMO-STA shared-then-adapt framework.}}}
        \label{fig:emo_sta_framework}
    \end{subfigure}
    \hspace{0.015\textwidth}
    \begin{subfigure}[t]{0.21\textwidth}
        \centering
        \includegraphics[width=\linewidth]{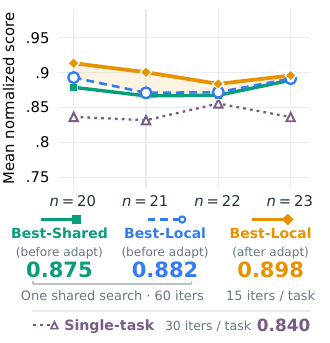}
        \caption{\small{\textbf{Gain breakdown.}}}
        \label{fig:circle-packing-seed-mean-compact}
    \end{subfigure}

    \vspace{-5pt}
    \caption{\small{\textbf{EMO-STA shared-then-adapt framework and adaptation behavior.}
    \textbf{(a):} A unified candidate-program interface lets the same evolving program \(p\) be evaluated across a related task family \(\mathcal{T}_1,\ldots,\mathcal{T}_K\). EMO-STA first runs shared evolution with aggregate objective \(s_{\mathrm{shared}}(p)=\frac{1}{K}\sum_{i=1}^{K} s_i(p)\), producing a shared archive \(\mathcal{A}_{\mathrm{shared}}\) of reusable programs. It then initializes task-specific adaptation from this archive using one of three STA variants: Warmstart, Best-Shared, or \textbf{Best-Local}. The \textbf{Best-Local} variant selects, for each target task, the shared program with the highest target-task score before adaptation and serves as our primary strategy for in-distribution task-local adaptation. \textbf{(b):} A Sonnet 4.5 circle-packing rectangle example decomposes gains into Best-Local selection from shared archive and subsequent adaptation.}}
    \label{fig:emo_sta}
    \vspace{-15pt}
\end{figure*}

In this paper, we tackle the multitask optimization challenge with LLMs by introducing a set of algorithms under the \emph{Shared-Then-Adapt (STA)} umbrella. EMO-STA first runs a single shared evolutionary search over the full task family, where each candidate is evaluated by a shared family-level objective that aggregates performance across tasks and produces an archive of candidate programs. Then, for each target task, EMO-STA starts a task-specific adaptation run from the shared archive using one of three initializations as detailed in Figure \ref{fig:emo_sta}: 

(i) \emph{STA Warmstart}: transfers the entire shared evolution program archive into the task-local archive and continues evolution from that population; 

(ii) \emph{STA Best-Shared}: uses the program with the best average shared score across the family as the initial program for task-local evolution; and 

(iii) \emph{STA Best-Local}: selects the program in the shared archive that performs best on the target task and uses it as the initial program for adaptation.

Building on this formalism, our work makes three contributions toward more efficient and generalizable LLM-guided evolutionary optimization: \textbf{First}, we introduce EMO-STA, a shared-then-adapt framework that evolves a reusable program archive across a task family before adapting selected candidates to individual tasks. By design, EMO-STA biases the search toward a general-purpose solution as it requires the LLM to write a single program that applies to all tasks simply by varying task-specific parameters. \textbf{Second}, across diverse continuous-optimization, geometric, modeling, and algorithmic task families, we show that EMO-STA improves solution quality over single-task evolution under matched iteration budgets in most settings (see Section~\ref{sec:empirical}). Our ablations show that STA Best-Local is the most reliable in-distribution adaptation strategy, while allocating a substantial—and often roughly balanced—fraction of the iteration budget to shared evolution yields strong performance. \textbf{Finally}, we show that shared evolution can improve generalization: \emph{STA Best-Shared} provides robust transfer to held-out tasks (Section~\ref{sec held out}), and in low-evidence settings such as ARC and time-series forecasting, shared evolution helps mitigate overfitting by favoring programs that succeed across related tasks rather than exploiting spurious task-specific artifacts (specifically, overfitting to the training dataset due to insufficient sample size, see Section~\ref{sec:shared-evolution-overfitting}). These out-of-distribution results are consistent with a useful structural
bias (e.g., a better search space) introduced by shared evolution when single-task search reinforces spurious task-specific solutions. The iteration-matched comparisons do not, however, isolate this effect from differences in evaluator work. 

Overall, EMO-STA offers an effective multi-task evolutionary search method that converts shared structure across related problems into stronger and more generalizable LLM-discovered programs.
%\input{sections/scratch_introduction}
%\vspace{-10pt}
\section{EMO-STA: Multitask Evolution with \emph{Shared-Then-Adapt}}

EMO-STA is applicable whenever related tasks can be expressed through a shared executable interface, so the same LLM-generated candidate program can be evaluated by task-specific evaluators during shared evolution and then adapted to individual tasks. We enforce this compatibility by defining a unified entry-point function and prompting the LLM to implement that format. For example, circle-packing tasks with \(n=20,22,24,26\) can share \texttt{construct\_packing(n)}, even though AlphaEvolve studies a fixed \(n=26\). Similarly, signal-processing tasks can vary the kinds of noisy signals they use---for example, signals that mix multiple frequencies, change frequency over time, or contain sudden jumps---while keeping the same interface \texttt{process\_signal(noisy\_signal, window\_size)}.

\subsection{Problem Setup}
We study whether a fixed family-level evolutionary-iteration budget can yield stronger solutions across related discovery tasks by first discovering reusable program structure jointly, then adapting it to each task, rather than solving each task independently. Formally, let $\mathfrak{T}=\{\mathcal{T}_1,\ldots,\mathcal{T}_K\}$ be a family of related discovery tasks. In EMO-STA, we assume that all tasks in a family are written so that the same candidate program can be executed on each of them. Equivalently, the family shares one program space \(\mathcal{P}\) and one executable interface, while each task \(\mathcal{T}_i\) has its own evaluator and scalar score \(s_i(\cdot)\), which may aggregate performance over multiple evaluation instances. When raw objectives differ in scale across tasks, we define \(s_i\) as a task-normalized score, so that the shared objective averages scores on a common scale rather than being biased toward tasks with larger numerical ranges. Given a total family-level iteration budget \(B_{\mathrm{tot}}\), our overall goal is to produce task-specific solutions \(p_1,\ldots,p_K \in \mathcal{P}\) that maximize average family performance,
\[
\max_{p_1,\ldots,p_K \in \mathcal{P}}
\frac{1}{K}\sum_{i=1}^{K} s_i(p_i),
\qquad
\text{subject to total iteration budget } \le B_{\mathrm{tot}},
\]
where \(s_i(\cdot)\) is the scalar, possibly normalized, score on task \(\mathcal{T}_i\).

\subsection{Shared Evolution Phase}
In the shared phase, we evolve a single archive against an aggregated family-level objective by evaluating each candidate program \(p\) on every task in the family:
\[
s_{\mathrm{shared}}(p)
=
\frac{1}{K}\sum_{i=1}^{K} s_i(p).
\]
Similar aggregate evaluation across multiple instances or benchmark
problems is used in prior LLM-based heuristic-design methods
\citep{eoh,llamea,reevo}. Here, we use this objective to evolve a
shared archive \(\mathcal{A}^{\mathrm{shared}}\) of programs that perform
well across the task family. Rather than requiring one shared program
to solve every task optimally, this archive provides reusable starting
points for the task-specific adaptation phase.

\subsection{Adaptation Evolution Phase}
Starting from the shared archive \(\mathcal{A}^{\mathrm{shared}}\), EMO-STA constructs task-local initialization checkpoints for each target task. For each target task \(\mathcal{T}_i\), we convert shared candidates into task-local checkpoint entries through a rescoring projection, denoted by \(\Pi_i : \mathcal{A}^{\mathrm{shared}} \rightarrow \mathcal{A}^{\mathrm{init}}_i\). For a shared program \(p\), \(\Pi_i(p)\) is the same executable program paired with its task-local evaluation state, obtained by re-evaluating the code under the evaluator for \(\mathcal{T}_i\). Using this projection, we consider three ways to initialize the task-local adaptation evolution.

\noindent\textit{\textbf{STA Warmstart.}}
This method transfers the entire shared program archive into the task-local archive and continues adaptation evolution from the resulting population:
\begin{align*}
\mathcal{A}^{\mathrm{init}}_{i,\mathrm{warm}}
=
\{\Pi_i(p) : p \in \mathcal{A}^{\mathrm{shared}}\}.
\end{align*}
\noindent\textit{\textbf{STA Best-Shared.}}
This method uses the program with the best shared family-level score as the initial program for adaptation evolution:
\begin{align*}
p_{\mathrm{bs}}^\star
=
\arg\max_{p \in \mathcal{A}^{\mathrm{shared}}} s_{\mathrm{shared}}(p),
\qquad
\mathcal{A}^{\mathrm{init}}_{i,\mathrm{bs}}
=
\{\Pi_i(p_{\mathrm{bs}}^\star)\}.
\end{align*}
\noindent\textit{\textbf{STA Best-Local.}}
This method uses the program in the shared archive that performs best on the target task \(\mathcal{T}_i\) as the initial program for task-local evolution:
\begin{align*}
p_{i,\mathrm{bl}}^\star
=
\arg\max_{p \in \mathcal{A}^{\mathrm{shared}}} s_i(p),
\qquad
\mathcal{A}^{\mathrm{init}}_{i,\mathrm{bl}}
=
\{\Pi_i(p_{i,\mathrm{bl}}^\star)\}.
\end{align*}
Together, these variants represent different ways of transferring the shared search state into task-local adaptation. \emph{STA Warmstart} preserves archive diversity, \emph{STA Best-Shared} starts from the strongest average family-level program, and \emph{STA Best-Local} starts from the shared candidate that already performs best on the target task.

Each initialized archive \(\mathcal{A}^{\mathrm{init}}_{i,v}\), \(v \in \{\mathrm{warm},\mathrm{bs},\mathrm{bl}\}\), is then evolved independently on task \(\mathcal{T}_i\) with the same adaptation budget. Thus, the three EMO-STA variants share the same shared phase, evaluator family, and budget; only the task-local initialization differs. Algorithm~\ref{alg:emo-sta} in Appendix~\ref{app:emo-sta-pseudocode} summarizes the full shared-then-adapt procedure.

\textbf{Reported scores and iteration-budget matching.}
For each task \(\mathcal{T}_i\), we report four scores: the three adapted EMO-STA variants initialized from \(\mathcal{A}^{\mathrm{init}}_{i,\mathrm{warm}}\), \(\mathcal{A}^{\mathrm{init}}_{i,\mathrm{bs}}\), and \(\mathcal{A}^{\mathrm{init}}_{i,\mathrm{bl}}\), respectively, and the \emph{Single-task} baseline, obtained by evolving \(\mathcal{T}_i\) from scratch without access to a shared archive. Section~\ref{sec:empirical} explains how we match iteration
budgets across methods and specifies how many iterations are used for
shared evolution, adaptation, and the single-task baseline.

%\begin{figure}[!t]
    %\centering
    %\includegraphics[width=0.7\linewidth]%{figures/k_module_problem_balanced_fixed_b30_budget_sweep.pdf}
    %\caption{Budget sweep for MT-STS on the K-module problem with the single-task %baseline fixed at 30 iterations. The leftmost bar corresponds to the single-task %setting $0 / 30 / 30$, while the remaining bars show MT-STS adapted performance for %different \textit{Shared / Adapt / Baseline} budgets. Each bar reports the mean %score across five models: Claude Haiku-4.5, Sonnet-4.5, Sonnet-4.6, Opus-4.5, and %Opus-4.6. The bold score annotation marks the best MT-STS configuration, $40 / 20 / %30$.}
%    \label{fig:kmb-balanced-fixed-b30-budget-sweep}
%\end{figure}

% Colors
\definecolor{singleTaskPeach}{RGB}{246,200,184}
\definecolor{staMethodColor}{RGB}{169,216,200}

% Legend box dimensions
\newcommand{\LegendBoxW}{0.52}
\newcommand{\LegendBoxH}{0.20}
\newcommand{\LegendRaise}{-0.1ex}

% Single-task legend box
\newcommand{\legendboxsingle}{%
  \raisebox{\LegendRaise}{%
    \tikz{
      \filldraw[draw=black, fill=singleTaskPeach, line width=0.35pt]
        (0,0) rectangle (\LegendBoxW,\LegendBoxH);
    }%
  }%
}

% STA Warmstart legend box
\newcommand{\legendboxwarmstart}{%
  \raisebox{\LegendRaise}{%
    \tikz{
      \filldraw[draw=black, fill=staMethodColor, line width=0.35pt]
        (0,0) rectangle (\LegendBoxW,\LegendBoxH);
      \begin{scope}
        \clip (0,0) rectangle (\LegendBoxW,\LegendBoxH);
        \foreach \x in {-0.35,-0.20,-0.05,0.10,0.25,0.40,0.55} {
          \draw[black, solid, line width=0.20pt] (\x,0) -- ++(0.34,0.34);
        }
      \end{scope}
    }%
  }%
}

% STA Best-Local legend box
\newcommand{\legendboxbestlocal}{%
  \raisebox{\LegendRaise}{%
    \tikz{
      \filldraw[draw=black, fill=staMethodColor, line width=0.35pt]
        (0,0) rectangle (\LegendBoxW,\LegendBoxH);
    }%
  }%
}

% STA Best-Shared legend box
\newcommand{\legendboxbestshared}{%
  \raisebox{\LegendRaise}{%
    \tikz{
      \filldraw[draw=black, fill=staMethodColor, line width=0.35pt]
        (0,0) rectangle (\LegendBoxW,\LegendBoxH);
      \begin{scope}
        \clip (0,0) rectangle (\LegendBoxW,\LegendBoxH);
        \foreach \x in {-0.35,-0.20,-0.05,0.10,0.25,0.40,0.55} {
          \draw[black, solid, line width=0.18pt] (\x,0) -- ++(0.34,0.34);
          \draw[black, solid, line width=0.18pt] (\x,\LegendBoxH) -- ++(0.34,-0.34);
        }
      \end{scope}
    }%
  }%
}

% Examples(\xz{find evidence. working on. not sure about what I write}): 

% \begin{itemize}
%     \item Cross-Language Evolution: Python → Rust → C++ optimization chains (listed in openevolve Roadmap, Research Directions) A "lesson learned" about loop unrolling in C++ could be abstracted into a natural-language strategy for a similar Python-based task.
%     \item Case Studies in SkyDiscover. Category 1: Mathematical Discoveries. AdaEvolve refines promising candidates using constrained optimization (SLSQP). Example 2: Max/Min Pairwise Distance Ratio in 3D (n = 14). EvoX discovers the best solution via constrained numerical optimization (SLSQP). SLSQP is shareable?
%     \item Also compare Example 1 and 2. both use Pairwise distance computation. Directly shareable
%     \item Example 2: Max/Min Pairwise Distance Ratio in 3D (n = 14). EvoX discovers the best solution via constrained numerical optimization (SLSQP), starting from three carefully chosen geometric seeds. The three carefully chosen geometric seeds can be shareable among all Max/Min Pairwise Distance Ratio in 3D problem?
% \end{itemize}

\vspace{-5pt}
\section{Empirical Evaluation of EMO-STA}\label{sec:empirical}
We evaluate EMO-STA on eight task families, each defined by a small set of related subtasks, spanning continuous optimization, geometric construction, modeling, and algorithmic tasks: function minimization, circle packing, circle-packing rectangles, Heilbronn triangle, signal processing, SLDBench-3D, Rust adaptive sort, and K-module. Our experiments address five questions: (i) whether STA improves over direct single-task optimization under matched evolutionary-iteration budgets, (ii) how the gains distribute across tasks within a family, (iii) how a fixed total task-family budget should be divided between shared and adaptation evolutions, (iv) whether the evolved programs generalize to held-out task sizes in OOD evaluation, and (v) how task relatedness affects EMO-STA's advantage over single-task evolution. For the main task-family comparisons, we report results for five models: Claude Haiku-4.5, Sonnet-4.5, Sonnet-4.6, Opus-4.5, and Opus-4.6. For these comparisons, we first average scores over the in-distribution tasks in the corresponding family for each run, then report \(\text{mean} \pm \text{std}\) over five independent runs, with all reported scores higher-is-better. Task-level figures report individual subtask scores separately. 

%We first report in-distribution family-level results and task-level transfer gains, then study the shared-versus-adaptation compute trade-off, then evaluate OOD transfer on nearby held-out task sizes, and finally present a separate exploratory ARC study. \textcolor{red}{update this part later}

\begin{table*}[t]
\centering
\caption{\small{Comparison of standard single-task and EMO-STA optimization for continuous optimization families. Each score cell reports \textit{mean $\pm$ std}; the first line is \textit{STA Best-Local / STA Warmstart}, and the second line is \textit{STA Best-Shared / Single-task}. Bold marks the largest mean among the four scores in each cell.}}
\label{tab:mt-sts-main-seed-adaptation-constructive}
\footnotesize
\setlength{\tabcolsep}{3pt}
\renewcommand{\arraystretch}{1.05}
\setlength{\aboverulesep}{0.5ex}
\setlength{\belowrulesep}{0.5ex}

\providecommand{\mtstscell}[1]{\begin{tabular}[c]{@{}c@{}}#1\end{tabular}}

\providecommand{\mtstsscorecell}[4]{%
\begin{tabular}[c]{@{}c@{}}
$#1$ / $#2$\\[-1pt]
$#3$ / $#4$
\end{tabular}}

\resizebox{\textwidth}{!}{%
\begin{tabular}{lcccc}
\toprule
\multirow{4}{*}{Model}
& \mtstscell{Function}
& \mtstscell{Circle}
& \mtstscell{Circle packing}
& \mtstscell{Heilbronn} \\
& \mtstscell{minimization}
& \mtstscell{packing}
& \mtstscell{rectangles}
& \mtstscell{triangle} \\[-1pt]
& {\scriptsize STA Best-Local / STA Warmstart}
& {\scriptsize STA Best-Local / STA Warmstart}
& {\scriptsize STA Best-Local / STA Warmstart}
& {\scriptsize STA Best-Local / STA Warmstart} \\[-2pt]
& {\scriptsize STA Best-Shared / Single-task}
& {\scriptsize STA Best-Shared / Single-task}
& {\scriptsize STA Best-Shared / Single-task}
& {\scriptsize STA Best-Shared / Single-task} \\
\midrule

\textbf{Haiku-4.5}
& \mtstsscorecell{\mathbf{.952 \pm .04}}{.949 \pm .05}{.941 \pm .06}{.888 \pm .05}
& \mtstsscorecell{.934 \pm .03}{.926 \pm .03}{\mathbf{.940 \pm .02}}{.865 \pm .03}
& \mtstsscorecell{.861 \pm .02}{\mathbf{.865 \pm .02}}{.845 \pm .01}{.832 \pm .01}
& \mtstsscorecell{\mathbf{.650 \pm .06}}{.628 \pm .05}{.628 \pm .06}{.547 \pm .03} \\
\midrule

\textbf{Sonnet-4.5}
& \mtstsscorecell{\mathbf{.925 \pm .02}}{.917 \pm .02}{.904 \pm .03}{.891 \pm .05}
& \mtstsscorecell{\mathbf{.965 \pm .02}}{.964 \pm .02}{.947 \pm .03}{.927 \pm .02}
& \mtstsscorecell{\mathbf{.898 \pm .03}}{.890 \pm .03}{.892 \pm .04}{.840 \pm .02}
& \mtstsscorecell{\mathbf{.622 \pm .04}}{.596 \pm .05}{.619 \pm .05}{.548 \pm .04} \\
\midrule

\textbf{Opus-4.5}
& \mtstsscorecell{\mathbf{.969 \pm .03}}{.942 \pm .07}{.941 \pm .09}{.914 \pm .05}
& \mtstsscorecell{\mathbf{.940 \pm .01}}{.926 \pm .01}{.930 \pm .01}{.912 \pm .01}
& \mtstsscorecell{\mathbf{.951 \pm .01}}{.943 \pm .01}{.943 \pm .01}{.912 \pm .01}
& \mtstsscorecell{\mathbf{.741 \pm .03}}{.732 \pm .04}{.704 \pm .04}{.622 \pm .06} \\
\midrule

\textbf{Sonnet-4.6}
& \mtstsscorecell{.988 \pm .02}{.973 \pm .03}{\mathbf{.991 \pm .02}}{.901 \pm .02}
& \mtstsscorecell{\mathbf{.997 \pm .00}}{\mathbf{.997 \pm .00}}{\mathbf{.997 \pm .00}}{.957 \pm .03}
& \mtstsscorecell{\mathbf{.986 \pm .01}}{.985 \pm .01}{.985 \pm .01}{.967 \pm .02}
& \mtstsscorecell{.862 \pm .04}{.809 \pm .04}{\mathbf{.865 \pm .07}}{.678 \pm .05} \\
\midrule

\textbf{Opus-4.6}
& \mtstsscorecell{\mathbf{.945 \pm .03}}{.943 \pm .03}{.932 \pm .04}{.895 \pm .04}
& \mtstsscorecell{\mathbf{.984 \pm .01}}{.972 \pm .02}{.979 \pm .02}{.963 \pm .01}
& \mtstsscorecell{\mathbf{.967 \pm .01}}{.957 \pm .01}{.962 \pm .01}{.944 \pm .01}
& \mtstsscorecell{.863 \pm .03}{.844 \pm .03}{\mathbf{.877 \pm .03}}{.744 \pm .04} \\

\bottomrule
\end{tabular}%
}
\vspace{-5pt}
\end{table*}

\subsection{Experimental setup}
This subsection summarizes the implementation, task construction, scoring, and iteration-budget matching.

\textbf{Implementation and task interface unification.}
All experiments are implemented on top of OpenEvolve, an open-source framework for AlphaEvolve-style evolutionary code search~\citep{openevolve,novikov2025alphaevolve}. We keep its underlying program-evolution loop, including island-based population management, MAP-Elites context selection, and checkpointing, and implement EMO-STA on top of that infrastructure. Here, each OpenEvolve iteration consists of proposing or editing one
candidate program and evaluating it. We apply EMO-STA to task families adaptable to a common program interface, so the same evolving program can be evaluated across all subtasks in the family. For example, in circle packing the same evolving program is used across all task sizes through \texttt{construct\_packing(n)}, with the evaluator varying only the current task parameter \(n\). Representative prompts specifying these shared interfaces are provided in Appendix~\ref{app:shared-interface-prompts}.

\noindent\textbf{Task families, baselines, and iteration budgets.}
Most task families contain four in-distribution training tasks, while
SLDBench-3D contains two. Within each family, the shared phase, evaluator family, and adaptation budget are identical across \emph{STA Warmstart}, \emph{STA Best-Shared}, and \emph{STA Best-Local}; only the task-local adaptation initialization differs. Thus, these variants ablate how the shared archive is used to initialize adaptation. The \emph{Single-task} baseline evolves each task independently from scratch, using the same program interface but without shared evolution or shared-archive access.

For a fair comparison, we match the family-level evolutionary-iteration budget. Let \(S\) denote shared iterations, \(A\) adaptation iterations per task, \(K\) the number of tasks, and \(B\) iterations per task for the
\emph{Single-task} baseline. We choose \(B\) such that \(S+KA=KB\),
giving both methods the same total number of evolutionary iterations.
This matches candidate-generation iterations, not task-level
evaluations, token usage, monetary cost, or wall-clock time: each shared
iteration evaluates its candidate across \(K\) tasks, whereas a
task-local iteration evaluates one.
Appendix~\ref{app:iteration-accounting} details this distinction.

We report budgets as \emph{Shared / Per-task Adapt / Total}, where
\emph{Total} is the common family-level iteration budget. The main
tables use \(120\) iterations for circle packing, circle packing
rectangles, Heilbronn triangle, and K-module; \(100\) for function
minimization, signal processing, and Rust adaptive sort; and \(80\)
for SLDBench-3D (\(K=2\)). Similar qualitative trends hold across
different total iteration budgets
(Appendix~\ref{app:additional-compute-allocation-results}).

\noindent \textbf{Score normalization and evaluation design.} For the three geometric families---circle packing, circle-packing rectangles, and Heilbronn triangle---different task sizes have different objective scales. To make shared evolution meaningful, we normalize each task score by its corresponding reference value, so that the shared objective averages comparable normalized scores rather than being dominated by larger-\(n\) tasks. As concrete examples, for circle-packing families we divide by fixed per-\(n\) reference sums of radii, and for Heilbronn triangle we divide by fixed per-\(n\) reference minimum triangle areas. Under this convention, a score of \(1\) corresponds to matching the reference value for that task size, while values below or above \(1\) indicate performance below or above that reference. These normalized scores are also the values reported in the tables for those three families. 

Additional details on task-family construction, score normalization, held-out task sizes, and iteration-budget allocation settings are provided in Appendix~\ref{app:task-family-design}.

\begin{table*}[t]
\centering
\caption{\small{Comparison of standard single-task and EMO-STA optimization for modeling and algorithmic optimization families. Each score cell reports \textit{mean $\pm$ std}; the first line is \textit{STA Best-Local / STA Warmstart}, and the second line is \textit{STA Best-Shared / Single-task}. Bold marks the largest mean among the four scores in each cell.}}
\label{tab:mt-sts-main-seed-adaptation-modeling}
\footnotesize
\setlength{\tabcolsep}{3pt}
\renewcommand{\arraystretch}{1.05}
\setlength{\aboverulesep}{0.5ex}
\setlength{\belowrulesep}{0.5ex}

\providecommand{\mtstscell}[1]{\begin{tabular}[c]{@{}c@{}}#1\end{tabular}}

\providecommand{\mtstsscorecell}[4]{%
\begin{tabular}[c]{@{}c@{}}
$#1$ / $#2$\\[-1pt]
$#3$ / $#4$
\end{tabular}}

\resizebox{\textwidth}{!}{%
\begin{tabular}{lcccc}
\toprule
\multirow{4}{*}{Model}
& \mtstscell{Signal}
& \multirow{2}{*}{SLDBench-3D}
& \mtstscell{Rust adaptive}
& \multirow{2}{*}{K-module (Balanced)} \\
& \mtstscell{processing}
&
& \mtstscell{sort}
& \\[-1pt]
& {\scriptsize STA Best-Local / STA Warmstart}
& {\scriptsize STA Best-Local / STA Warmstart}
& {\scriptsize STA Best-Local / STA Warmstart}
& {\scriptsize STA Best-Local / STA Warmstart} \\[-2pt]
& {\scriptsize STA Best-Shared / Single-task}
& {\scriptsize STA Best-Shared / Single-task}
& {\scriptsize STA Best-Shared / Single-task}
& {\scriptsize STA Best-Shared / Single-task} \\
\midrule

\textbf{Haiku-4.5}
& \mtstsscorecell{\mathbf{.600 \pm .05}}{.584 \pm .06}{.597 \pm .04}{.569 \pm .01}
& \mtstsscorecell{\mathbf{.958 \pm .02}}{.953 \pm .02}{.949 \pm .02}{.951 \pm .01}
& \mtstsscorecell{.533 \pm .02}{.535 \pm .02}{.509 \pm .03}{\mathbf{.539 \pm .02}}
& \mtstsscorecell{.567 \pm .06}{.567 \pm .04}{\mathbf{.575 \pm .07}}{.550 \pm .03} \\
\midrule

\textbf{Sonnet-4.5}
& \mtstsscorecell{\mathbf{.587 \pm .01}}{.578 \pm .02}{.582 \pm .02}{.576 \pm .01}
& \mtstsscorecell{\mathbf{.976 \pm .01}}{.971 \pm .01}{.971 \pm .02}{.959 \pm .01}
& \mtstsscorecell{.481 \pm .03}{.484 \pm .03}{.457 \pm .03}{\mathbf{.528 \pm .01}}
& \mtstsscorecell{.617 \pm .03}{\mathbf{.650 \pm .02}}{.567 \pm .06}{.617 \pm .05} \\
\midrule

\textbf{Opus-4.5}
& \mtstsscorecell{.620 \pm .03}{\mathbf{.635 \pm .03}}{.625 \pm .02}{.568 \pm .01}
& \mtstsscorecell{\mathbf{.983 \pm .00}}{.972 \pm .01}{.981 \pm .00}{.973 \pm .01}
& \mtstsscorecell{.515 \pm .05}{\mathbf{.520 \pm .05}}{.483 \pm .05}{.497 \pm .02}
& \mtstsscorecell{.617 \pm .03}{\mathbf{.675 \pm .03}}{.592 \pm .03}{.567 \pm .05} \\
\midrule

\textbf{Sonnet-4.6}
& \mtstsscorecell{\mathbf{.628 \pm .04}}{.626 \pm .04}{.613 \pm .05}{.608 \pm .03}
& \mtstsscorecell{.969 \pm .01}{.968 \pm .01}{\mathbf{.969 \pm .01}}{.955 \pm .01}
& \mtstsscorecell{.659 \pm .01}{\mathbf{.663 \pm .01}}{.656 \pm .01}{.616 \pm .03}
& \mtstsscorecell{.617 \pm .09}{\mathbf{.700 \pm .07}}{.575 \pm .03}{.675 \pm .05} \\
\midrule

\textbf{Opus-4.6}
& \mtstsscorecell{.713 \pm .05}{.707 \pm .04}{\mathbf{.716 \pm .04}}{.648 \pm .03}
& \mtstsscorecell{\mathbf{.975 \pm .01}}{.973 \pm .01}{.967 \pm .01}{.964 \pm .02}
& \mtstsscorecell{.616 \pm .02}{\mathbf{.625 \pm .02}}{.612 \pm .02}{.531 \pm .05}
& \mtstsscorecell{.725 \pm .02}{\mathbf{.800 \pm .05}}{.692 \pm .05}{.758 \pm .08} \\

\bottomrule
\end{tabular}%
}
\vspace{-10pt}
\end{table*}

\subsection{In-Distribution Results and Task-Level Transfer}

\begin{figure*}[!t]
    \centering

    \begin{center}
    \small
    \legendboxsingle\ Single-task
    \hspace{1.0em}
    \legendboxwarmstart\ STA Warmstart
    \hspace{1.0em}
    \legendboxbestlocal\ STA Best-Local
    \hspace{1.0em}
    \legendboxbestshared\ STA Best-Shared
    \end{center}

    \begin{subfigure}[t]{0.49\textwidth}
        \centering
        \includegraphics[width=\linewidth]{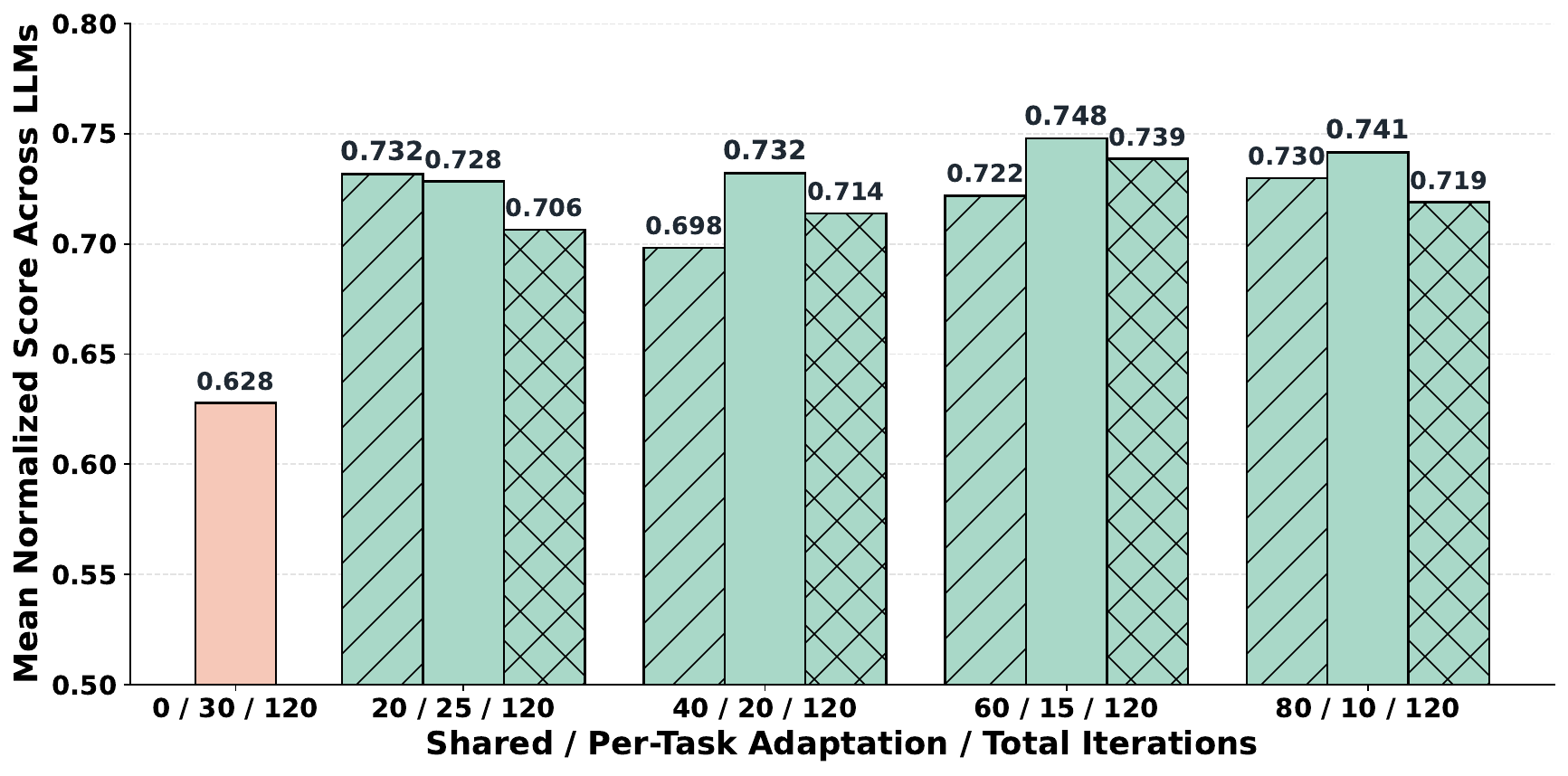}
        \caption{\small{\textit{Heilbronn triangle} family with 120 total iterations. The best setting is \textit{STA Best-Local} at $60 / 15 / 120$.}}
        \label{fig:heilbronn-fixed-b120-seed-adaptation-budget-sweep}
    \end{subfigure}
    \hfill
    \begin{subfigure}[t]{0.49\textwidth}
        \centering
        \includegraphics[width=\linewidth]{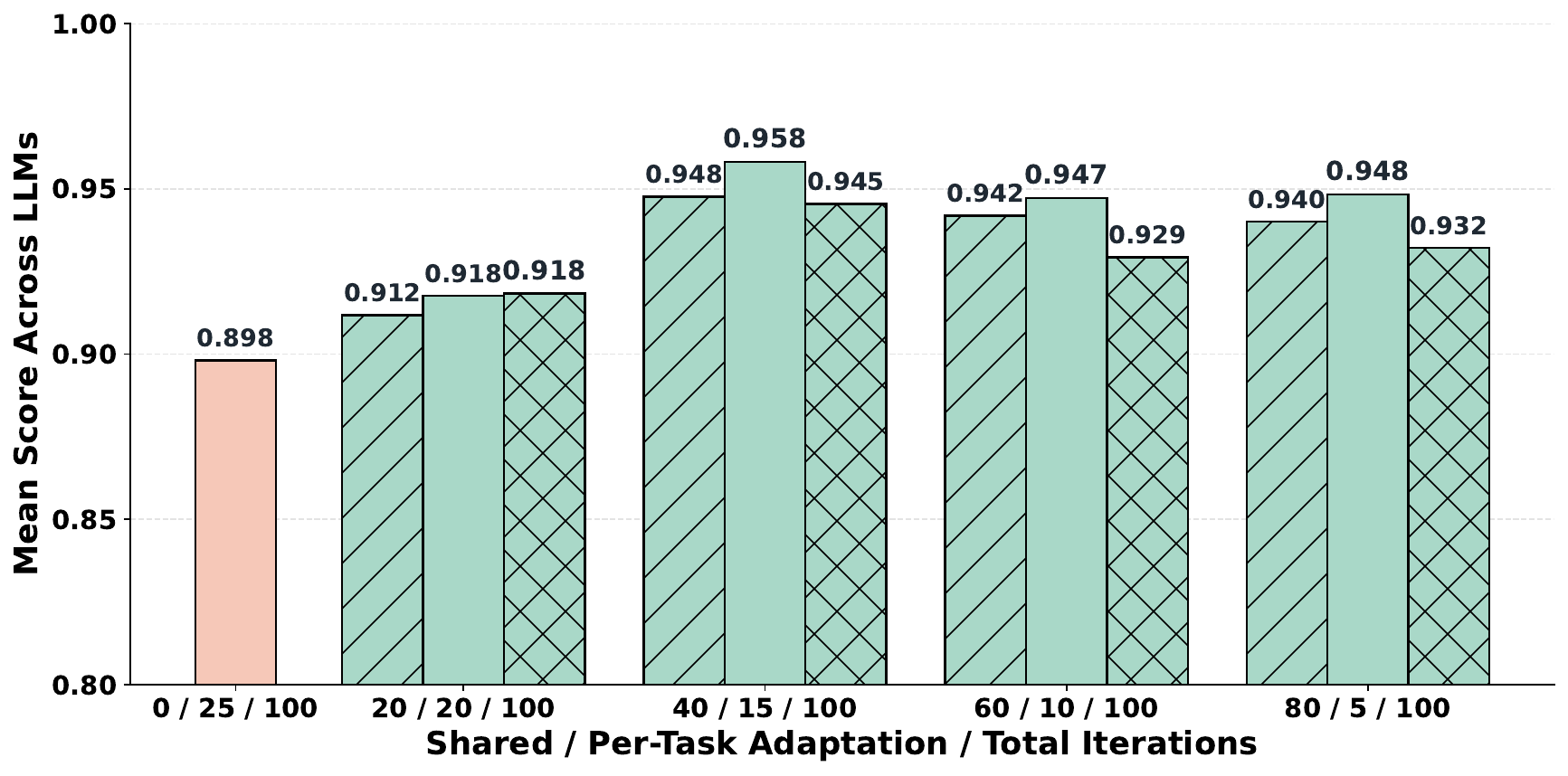}
        \caption{\small{\textit{Function minimization} family with 100 total iterations. The best setting is \textit{STA Best-Local} at $40 / 15 / 100$.}}
        \label{fig:function-minimization-fixed-b100-seed-adaptation-budget-sweep}
    \end{subfigure}

    \caption{
    \small{Iteration-budget allocation results for EMO-STA on two \(K=4\) task families.
    Grouped bars compare \textit{STA Warmstart}, \textit{STA Best-Local}, and
    \textit{STA Best-Shared} across \emph{Shared / Per-task Adapt / Total}
    settings, giving shared-evolution, per-task adaptation, and total
    family-level iteration counts. The leftmost bar is the direct single-task
    baseline with allocation \(0/B/KB\). Bars show mean scores averaged across
    five Claude models.}
    }
    \label{fig:emo-sta-budget-sweep-side-by-side}
    \vspace{-15pt}
\end{figure*}

We begin with the in-distribution comparison against direct single-task evolution, then inspect how improvements vary across individual subtasks.

\textbf{EMO-STA consistently improves over single-task optimization.}
Tables~\ref{tab:mt-sts-main-seed-adaptation-constructive} and
\ref{tab:mt-sts-main-seed-adaptation-modeling} compare EMO-STA against single-task evolution under matched evolutionary-iteration budgets. Across both tables, at least one EMO-STA variant exceeds the iteration-matched single-task baseline in \(38/40\) model--family settings. The improvements are especially large on geometric tasks such as Heilbronn triangle, where performance improves from \(0.547\) to \(0.650\) for Haiku-4.5, from \(0.622\) to \(0.741\) for Opus-4.5, and from \(0.744\) to \(0.877\) for Opus-4.6. Importantly, EMO-STA also improves when the single-task baseline is already competitive, e.g., from \(0.957\) to \(0.997\) on circle packing with Sonnet-4.6, from \(0.944\) to \(0.967\) on circle-packing rectangles with Opus-4.6, and from \(0.964\) to \(0.975\) on SLDBench-3D with Opus-4.6.

\textbf{\emph{STA Best-Local} most frequently achieves the highest in-distribution mean score.}
Across in-distribution results, \emph{STA Best-Local} achieves the highest average score, is strongest in \(23/40\) settings, and outperforms the \emph{Single-task} baseline in \(35/40\) settings. This is intuitive because the shared archive contains multiple useful candidates, and the best starting point for a task is often not the program with the highest average family-level score, but the one that already performs best on that task. By average score, \emph{STA Warmstart} is next, followed by \emph{STA Best-Shared}; both remain competitive, with \emph{STA Best-Shared} outperforming the \emph{Single-task} baseline in \(33/40\) settings.

\textbf{Some task structures favor archive-level transfer.}
In the balanced K-module family, \emph{STA Warmstart} achieves the highest mean score for four of five models. This suggests that retaining multiple shared candidates can aid task-local adaptation even when overall task-score correlation is zero. In Rust adaptive sort, weaker models struggle to translate shared evolution into task-specific gains, while stronger models benefit from the shared archive with \emph{STA Warmstart} and improve over the single-task baseline.

\begin{wrapfigure}{r}{0.6\textwidth}
    \centering
    \vspace{-12pt}
    \includegraphics[width=0.6\textwidth]{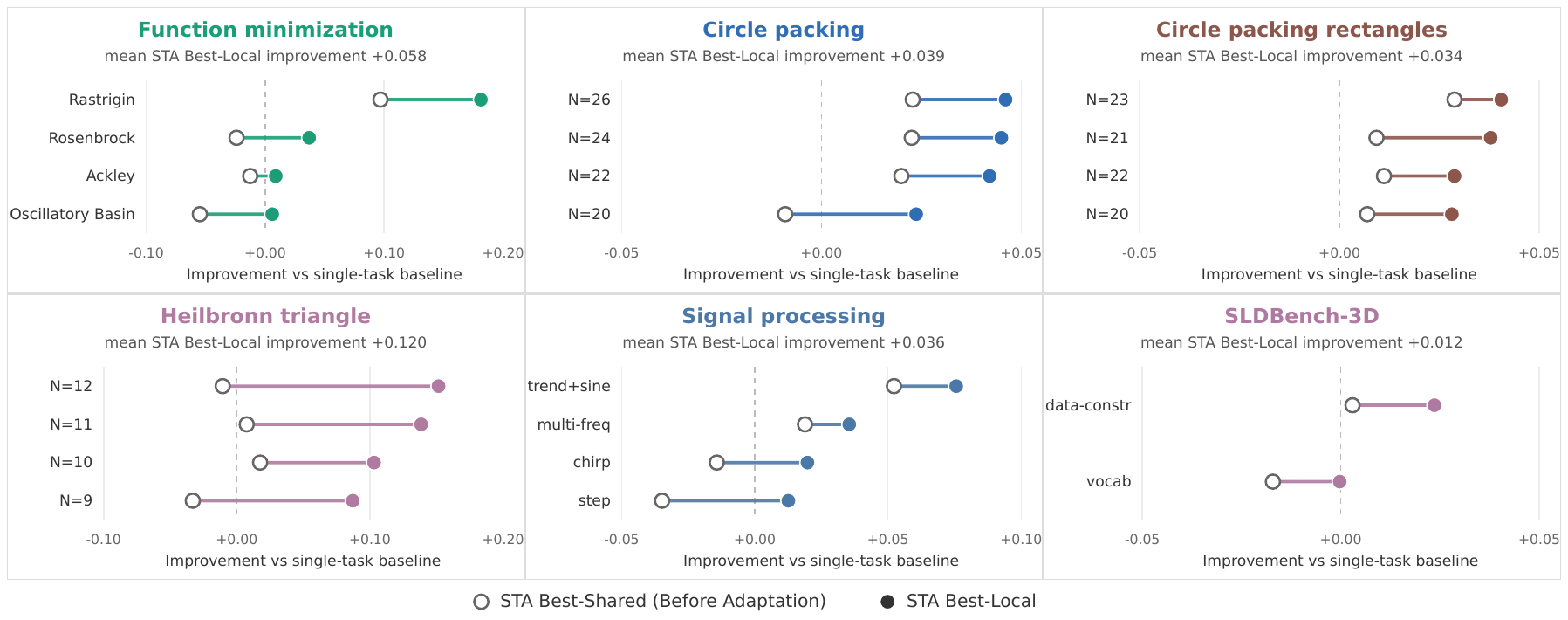}
    \caption{\small{Task-level transfer gains for \textit{STA Best-Local}. Each panel shows one task family and each row one in-distribution task. Open markers show the pre-adaptation shared score, filled markers show the final \textit{STA Best-Local} score, and the x-axis reports improvement over the single-task baseline.}}
    \label{fig:emo-sta-best-task-seed-gain-profile}
    \vspace{-12pt}
\end{wrapfigure}

\textbf{Task-level gains confirm the family-level improvement trends.}
Figure~\ref{fig:emo-sta-best-task-seed-gain-profile} breaks down the \emph{STA Best-Local} results by individual tasks, showing how gains vary across them. The open markers show the pre-adaptation best shared program score and the filled markers show the final adapted score, both measured relative to the single-task baseline. Across the subtasks shown in Figure~\ref{fig:emo-sta-best-task-seed-gain-profile}, the filled markers move to the right of the open markers and lie above zero, showing that task-local adaptation consistently improves the shared candidate and turns family-level structure into task-level gains. This indicates that EMO-STA is not merely improving family-level averages through one outlier task; rather, shared search followed by task-local adaptation improves many subtasks within each family. Also, Appendix~\ref{app:emo-sta-program-examples} gives concrete program examples of this behavior, where shared evolution finds a reusable solver structure and \emph{STA Best-Local} refines it for a target task.

\begingroup
\noindent\textbf{Additional ablations.}
In Appendix~\ref{app:search-ablations}, we further compare EMO-STA
with several ablations using Qwen3.8-27B, including shared-only search,
adaptation initialized from candidates found by searching a single
source task, and adaptation from candidates collected across multiple
short single-task searches. STA achieves the highest mean score among the iteration-matched methods in these experiments, supporting the benefit of shared search followed by task-local adaptation.
\par
\endgroup

\begin{figure*}[t]
    \centering

    \begin{center}
    \small
    \legendboxsingle\ Single-task
    \hspace{1.0em}
    \legendboxwarmstart\ STA Warmstart
    \hspace{1.0em}
    \legendboxbestlocal\ STA Best-Local
    \hspace{1.0em}
    \legendboxbestshared\ STA Best-Shared
    \end{center}

    \begin{subfigure}[t]{0.32\textwidth}
        \centering
        \includegraphics[width=\linewidth]{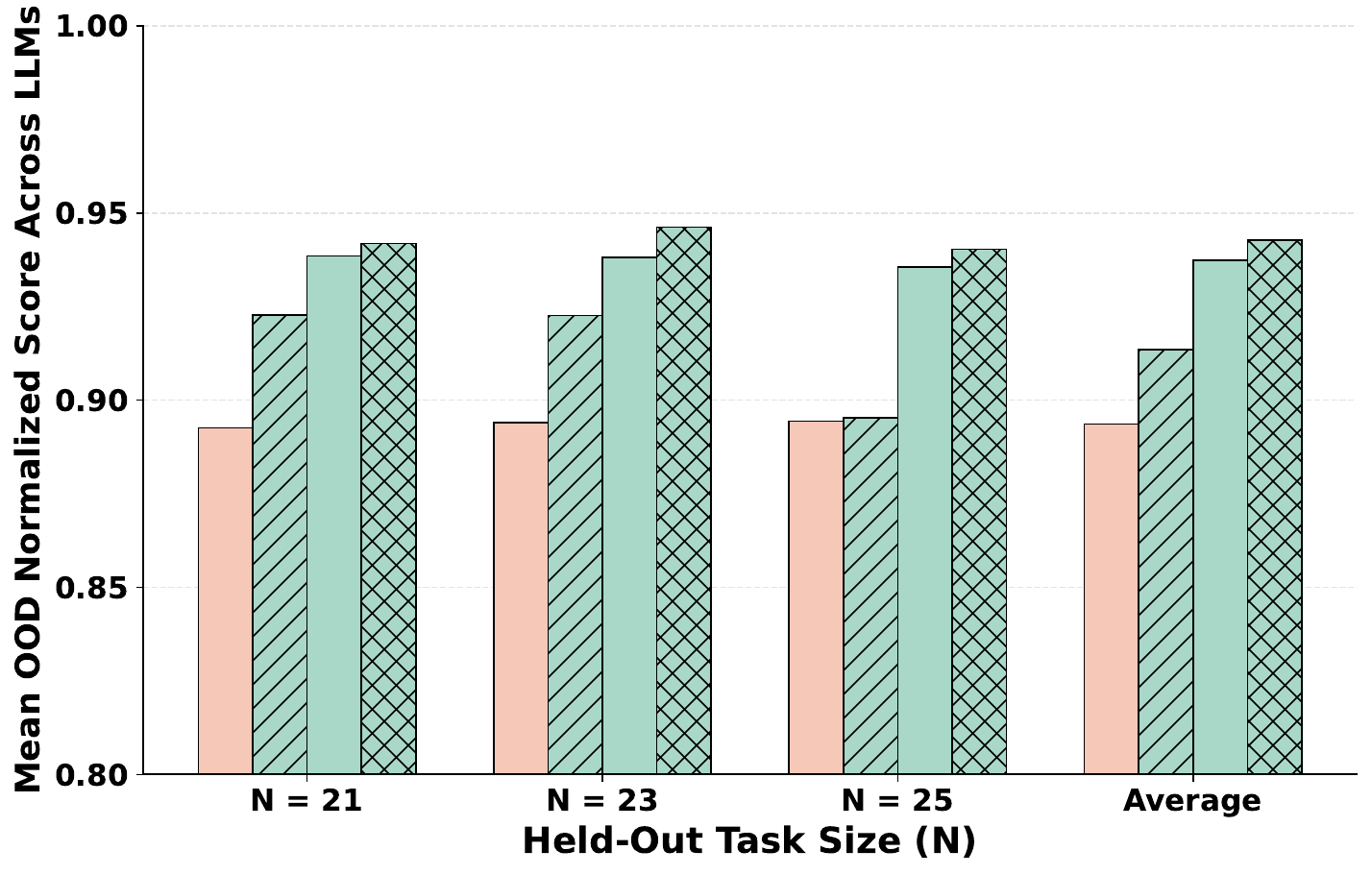}
        \caption{\small{\textit{Circle packing}.}}
        \label{fig:cp-ood-s60-a15-b30-all-methods}
    \end{subfigure}
    \hfill
    \begin{subfigure}[t]{0.32\textwidth}
        \centering
        \includegraphics[width=\linewidth]{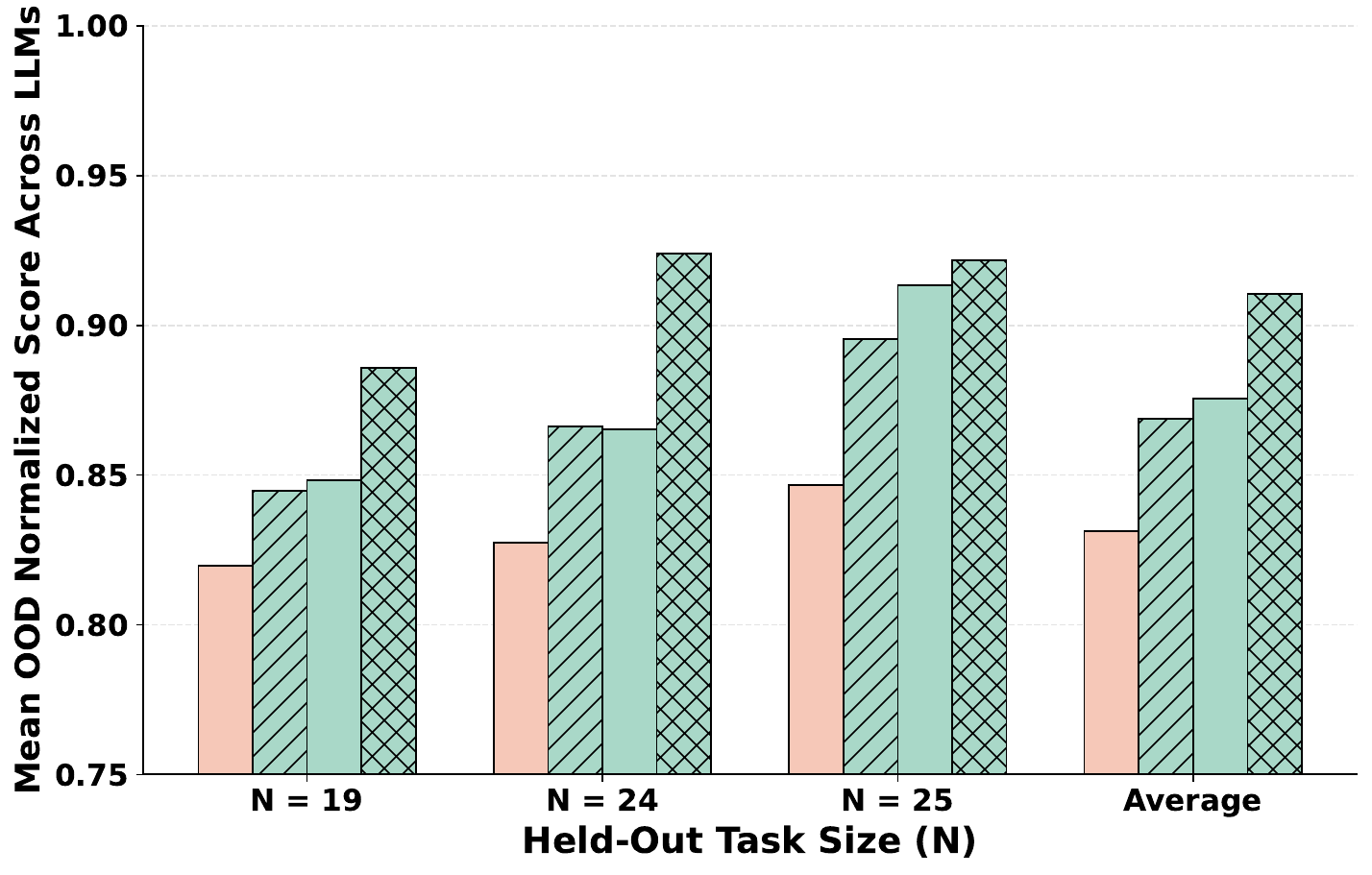}
        \caption{\small{\textit{Circle packing rectangle}.}}
        \label{fig:cp-rect-ood-s60-a15-b30-all-methods}
    \end{subfigure}
    \hfill
    \begin{subfigure}[t]{0.32\textwidth}
        \centering
        \includegraphics[width=\linewidth]{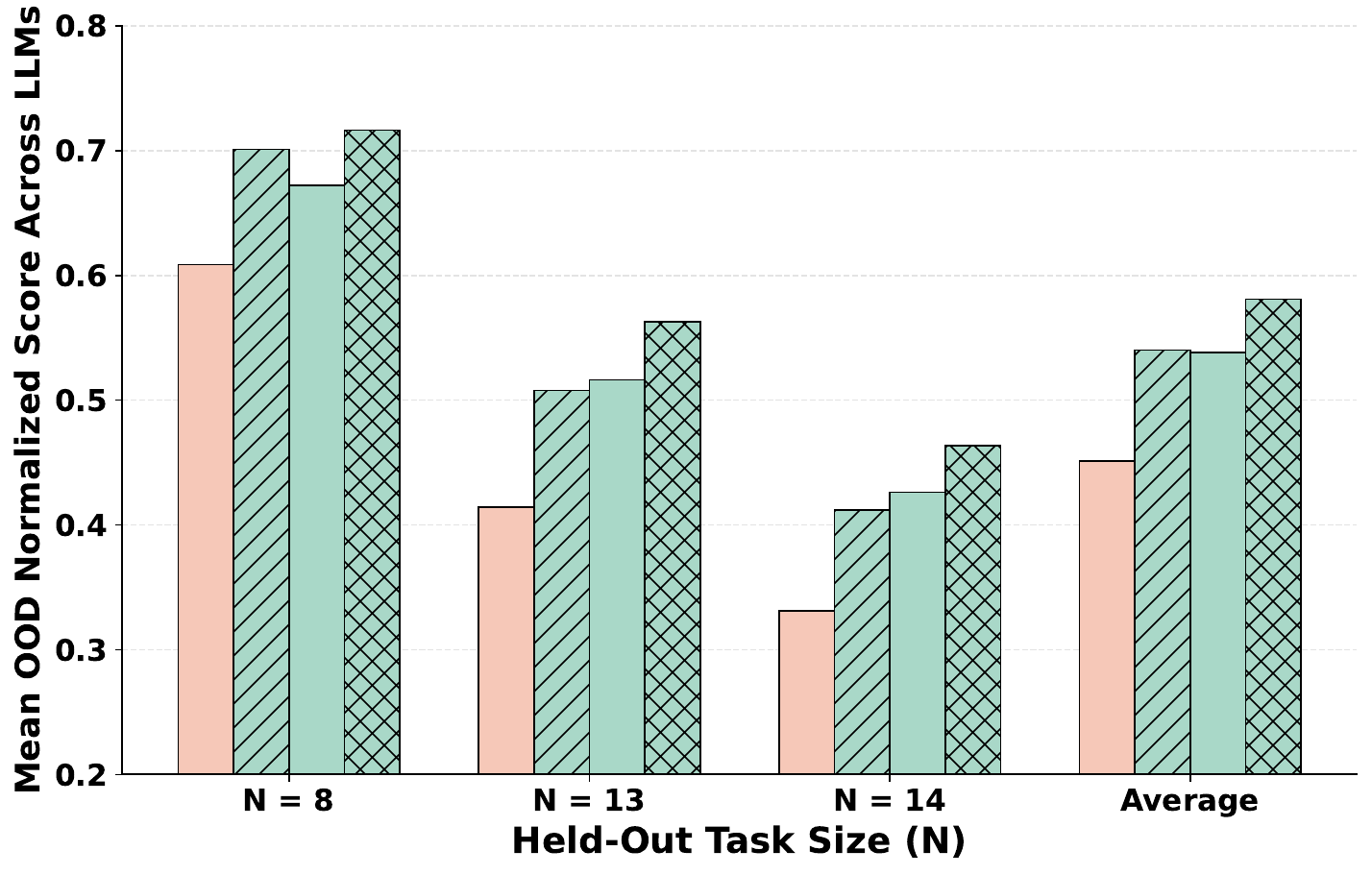}
        \caption{\small{\textit{Heilbronn triangle}.}}
        \label{fig:heilbronn-ood-s60-a15-b30-all-methods}
    \end{subfigure}

    \caption{\small{Held-out task-size evaluation at fixed \textit{60 / 15 / 120} iteration-budget allocation across three domains: \textit{circle packing}, \textit{circle packing rectangle}, and \textit{Heilbronn triangle}. For each STA variant, programs are adapted to each in-distribution task size and evaluated on each held-out size; bars report mean OOD score across adaptation source tasks and models. The comparison includes the \textit{Single-task} baseline and the STA variants: \textit{STA Warmstart}, \textit{STA Best-Local}, and \textit{STA Best-Shared}.}}
    \label{fig:ood-s60-a15-b30-all-methods}
    \vspace{-15pt}
\end{figure*}

\subsection{Iteration Allocation: Shared vs.\ Adaptation Evolutions}
Figure~\ref{fig:emo-sta-budget-sweep-side-by-side} and
Appendix~\ref{app:additional-compute-allocation-results} compare how a
fixed family-level iteration budget is divided between shared evolution and
per-task adaptation. Across all three EMO-STA variants, nonzero shared
budgets consistently outperform the direct single-task baseline.

\textbf{Equal shared/adaptation splits are consistently strong.}
A particularly effective recipe is to split the iteration budget roughly evenly between shared evolution and aggregate task-local adaptation, i.e., \(S \approx K A\). In Figure~\ref{fig:heilbronn-fixed-b120-seed-adaptation-budget-sweep}, the equal-split setting \(60/15/120\) is the best setting for Heilbronn triangle, while Figure~\ref{fig:function-minimization-fixed-b100-seed-adaptation-budget-sweep} shows that function minimization peaks nearby at \(40/15/100\). The equal-split point or a nearby allocation provides a strong shared/adaptation trade-off across all three \textit{STA} variants. Additional results in Appendix~\ref{app:additional-compute-allocation-results} show the same budget-allocation pattern for the two circle-packing families. Intuitively, this balance gives the shared phase enough budget to discover reusable family-level structure, while preserving sufficient task-local iterations to specialize that structure to each task.

\begin{wrapfigure}{r}{0.38\linewidth}
    \centering
    \vspace{-4em}
    \includegraphics[width=\linewidth]{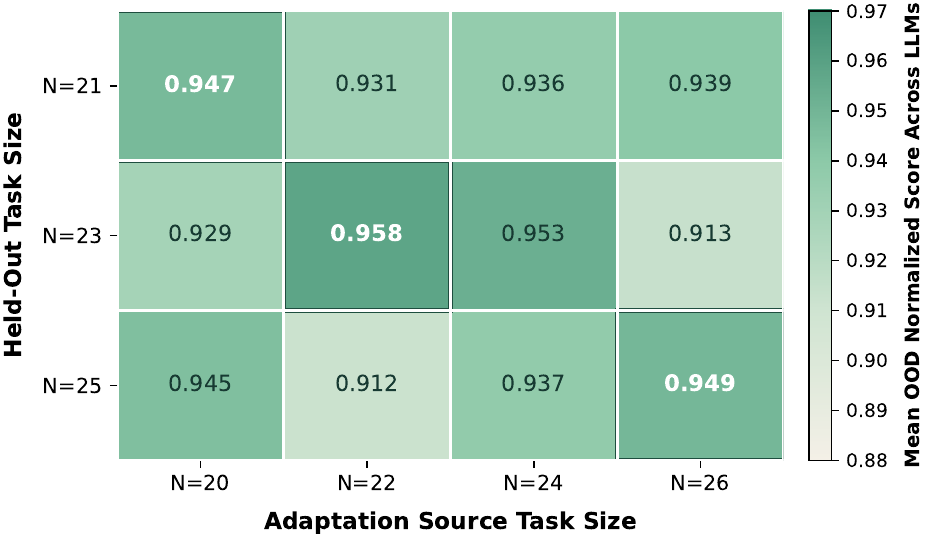}
    \caption{\small{OOD circle-packing results for \textit{STA Best-Local} at budget $60 / 15 / 120$. Rows are held-out sizes, columns are source tasks, and cells report mean OOD normalized score across LLMs and seeds.}}
    \label{fig:circle-packing-best-local-ood-transfer-heatmap}
    \vspace{-2.5em}
\end{wrapfigure}

%\vspace{-10pt}
\subsection{Generalization to Held-Out Task Sizes}\label{sec held out}
We evaluate out-of-distribution (OOD) transfer in geometric families. For each EMO-STA variant, we take programs adapted to each in-distribution task, evaluate them on each held-out task, and in Figure~\ref{fig:ood-s60-a15-b30-all-methods} report the mean across adaptation source tasks. The in-distribution / held-out sizes are \(N\!=\! \{20,22,24,26\}/\{21,23,25\}\) for circle packing, \(N\!=\!\{20,21,22,23\}/\{19,24,25\}\) for circle packing in rectangles, and \(N\!=\!\{9,10,11,12\}/\{8,13,14\}\) for Heilbronn triangle. Additional OOD results in Appendix~\ref{app:additional-ood-evaluation-details} break down the held-out evaluations across shared/adaptation iteration allocations, showing similar overall trends.

\textbf{\emph{STA Best-Shared} gives the strongest held-out transfer.}
Figure~\ref{fig:ood-s60-a15-b30-all-methods} evaluates in-distribution programs on held-out task sizes for the three geometric families. Across circle packing, circle packing in rectangles, and Heilbronn triangle, \emph{STA Best-Shared} is the strongest OOD variant on average. This suggests that the best shared evolution program often captures the most transferable family-level structure, even when \emph{STA Best-Local} is stronger for in-distribution adaptation. The gap is especially clear for circle packing in rectangles and Heilbronn triangle, where \emph{STA Best-Shared} outperforms both \emph{STA Warmstart} and \emph{STA Best-Local} across held-out sizes.

\textbf{Nearby task sizes transfer best.} The circle-packing heatmap in Figure~\ref{fig:circle-packing-best-local-ood-transfer-heatmap} shows how \emph{STA Best-Local} transfer depends on the source task used for adaptation. Nearby task sizes tend to transfer best as adaptation on \(N=20\) gives the strongest result for held-out \(N=21\) (\(0.947\)), adaptation on \(N=22\) is best for held-out \(N=23\) (\(0.958\)), and adaptation on \(N=26\) is best for held-out \(N=25\) (\(0.949\)). This illustrates that EMO-STA learns reusable family-level programs, while task-local adaptation still tunes the solution toward a particular region of the task family.

\begin{figure*}[t]
    \centering

    \begin{subfigure}[t]{0.34\textwidth}
        \vspace{0pt}
        \centering
        \includegraphics[width=\linewidth]{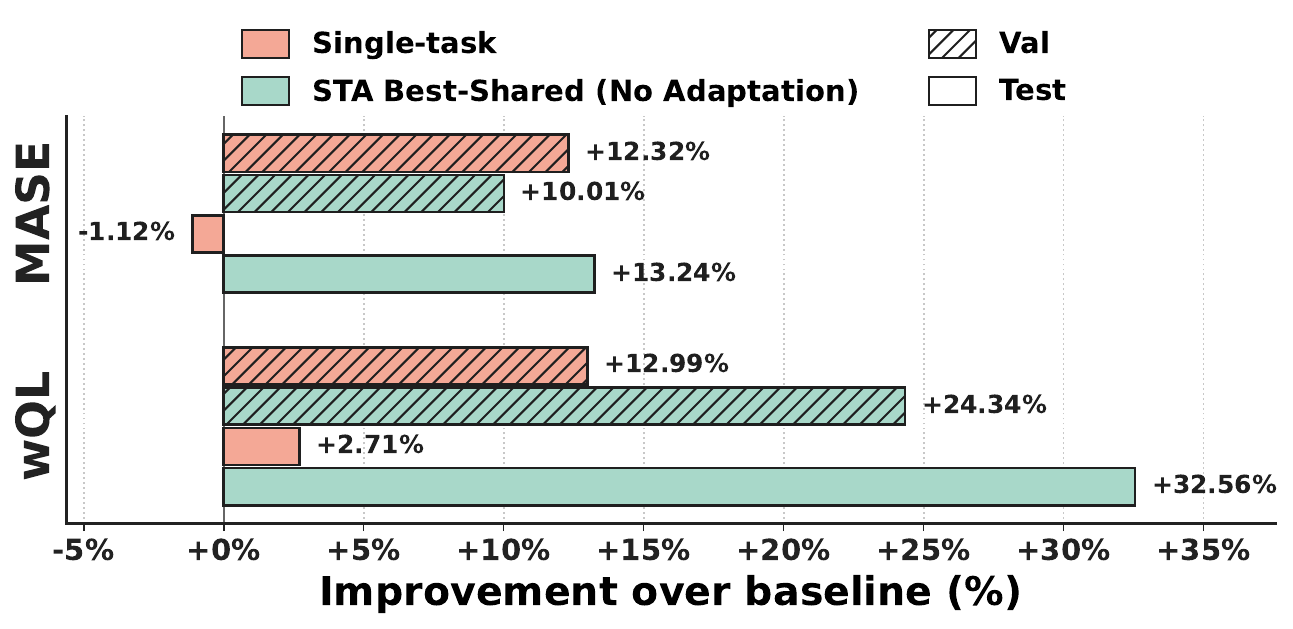}
        \caption{\small{\textit{COVID Deaths}. Single-task evolution transfers poorly to held-out test windows, while \textit{STA Best-Shared} achieves stronger test performance.}}
        \label{fig:covid-deaths}
    \end{subfigure}\hfill
    \begin{subfigure}[t]{0.34\textwidth}
        \vspace{0pt}
        \centering
        \includegraphics[width=\linewidth]{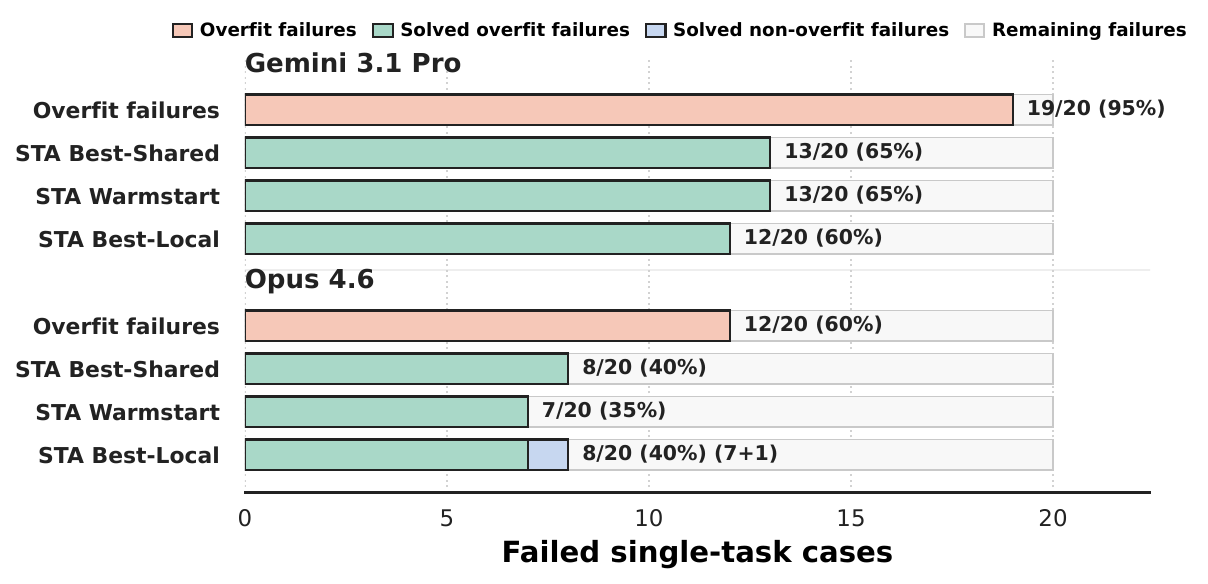}
        \caption{\small{\textit{ARC}. Across both models, EMO-STA variants recover selected single-task failures through shared evolution over transformed task variants, especially for overfit failures.}}
        \label{fig:arc_overfitting}
    \end{subfigure}\hfill
    \begin{subfigure}[t]{0.27\textwidth}
        \vspace{0pt}
        \centering
        \includegraphics[width=\linewidth]{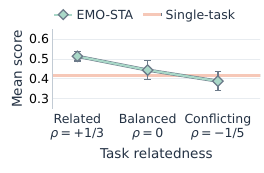}
        \caption{\small{\textit{Task relatedness}. On controlled K-module tasks, EMO-STA's advantage decreases as relatedness weakens. }}
        \label{fig:k-module-relatedness}
    \end{subfigure}

    \vspace{-5pt}
    \caption{\small{\textbf{Generalization and task relatedness.}
    (a--b) Shared evolution mitigates overfitting in two low-sample settings. In time-series feature engineering, one shared transformation improves held-out test performance over per-series evolution. In ARC, transformation-based shared evolution primarily recovers overfit failures. (c) A controlled K-module experiment evaluates how EMO-STA's advantage depends on task relatedness.}}
    \label{fig:shared-evolution-overfitting}
    \vspace{-15pt}
\end{figure*}

\begingroup

\vspace{-4pt}
\subsection{Effect of Task Relatedness on EMO-STA}
\label{sec:task-relatedness}
EMO-STA relies on related tasks sharing reusable structure, raising a natural question: when does sharing help, and when can it cause negative transfer? To isolate task relatedness from task difficulty, we construct controlled K-module families with the same search space, scoring rule, per-task score distribution, and optimal score, while varying overlap among their hidden target configurations. Starting from the balanced construction used in the main K-module experiments, we create related and conflicting variants by varying target overlap, and quantify relatedness by the mean pairwise Pearson correlation \(\rho\) between task scores over all configurations.

Figure~\ref{fig:k-module-relatedness} compares \emph{STA Best-Local} against single-task using Qwen3.8-27B under matched 120-iteration family budgets. EMO-STA's advantage is largest for related tasks (\(\rho=1/3\)), smaller for balanced tasks (\(\rho=0\)), and becomes negative relative to the single-task reference for conflicting tasks (\(\rho=-1/5\)). For this model and allocation, greater target overlap strengthens EMO-STA's advantage, while conflicting targets can induce negative transfer. Appendix~\ref{app:k-module-relatedness} provides details.
\par
\endgroup

% \vspace{-5pt}
\section{Mitigating Overfitting with Shared Evolution}
\label{sec:shared-evolution-overfitting}
\vspace{-4pt}

We next examine generalization in low-evidence settings, complementing fixed-iteration solution-quality comparisons. When each individual task provides limited or noisy evidence, independent evolution can overfit the task-specific objective itself, so additional single-task evolution may reinforce spurious solutions rather than improve generalization. We investigate the benefit of shared evolution across two different settings: Abstraction and Reasoning Corpus (ARC) ~\citep{chollet2019measure,xu2024llmsabstractionreasoningcorpus} and Time-series feature engineering. In ARC, this appears as programs that fit the few training examples but fail the hidden test grid; in time-series forecasting, it appears as validation improvements that do not transfer to held-out test windows. Across both settings, shared evolution acts as a regularizer by favoring programs or transformations that work across related tasks rather than only one sparse target.

\vspace{-3pt}
\subsection{ARC}
\vspace{-2pt}
We evaluate EMO-STA on the Abstraction and Reasoning Corpus (ARC), a natural setting for studying overfitting in program evolution. Because each ARC task has only a few input--output training pairs, single-task search can find programs that fit the examples without capturing the underlying rule. We use the ARC Prize 2025 ARC-AGI evaluation split. To construct a controlled multi-task setting, we generate transformed variants of each ARC task using invertible spatial and symbolic transformations, such as rotations, flips, and color permutations. These transformations change the surface appearance of the grids while preserving the same reasoning rule up to the corresponding transformation. Additional setup details are provided in \Cref{app:arc_setting}.

\textbf{Overfitting in Single-Task Evolution.}
We first run the single-task evolution baseline on ARC tasks separately for each primary model and select the first 20 failed run--task instances in evaluation-split order for each model. As shown in \Cref{fig:arc_overfitting}, overfitting is the dominant failure mode, especially for Gemini-3.1-Pro-Preview: \(19/20\) Gemini failures and \(12/20\) Claude Opus 4.6 failures fit all training examples but fail to generalize to the held-out test input.

\textbf{Multi-Task Evolution with EMO-STA Variants.}
Motivated by this diagnosis, we evaluate whether structured task diversity can reduce overfitting on the same selected failure cases. For each original ARC task, EMO-STA jointly evolves over the original task and its transformed variants, allowing information to be shared across tasks that follow the same underlying rule. We evaluate all three archive-initialization variants under the same shared phase and adaptation budget. 

As shown in \Cref{fig:arc_overfitting}, all EMO-STA variants recover a substantial fraction of failed single-task cases. For Gemini-3.1-Pro-Preview, \emph{STA Best-Shared} and \emph{STA Warmstart} each solve \(13/20\) cases, while \emph{STA Best-Local} solves \(12/20\). For Claude Opus 4.6, \emph{STA Best-Shared} and \emph{STA Best-Local} each solve \(8/20\), while \emph{STA Warmstart} solves \(7/20\). Since almost all recovered cases come from overfit failures, these results suggest that transformation-based shared evolution provides useful regularization to obtain solutions that generalize beyond sparse training examples.

\vspace{-10pt}
\subsection{Time-series Feature Engineering}
\label{subsec:time-series-feature-engineering}

We next use the EFE-Time time-series feature engineering
setting~\citep{taga2026evolutionaryfeatureengineeringstructured}
as a data-science testbed for overfitting in evolutionary optimization. EFE-Time evolves
preprocessing programs for time-series forecasting, where each candidate transformation is
selected by validation forecasting performance over the entire dataset level. 
%\begin{wrapfigure}{r}{0.35\linewidth}
%    \centering
%    \includegraphics[width=\linewidth]{figures/covid_deaths.pdf}
%    \caption{\small Single-task evolution
%    improves validation metrics but transfers poorly to held-out test windows, while STA
%   Best-Shared applies one shared transformation without task-local adaptation and
%    achieves stronger test gains.}
%    \label{fig:covid-deaths}
%    \vspace{-0.5em}
%\end{wrapfigure} 
This setting is an ideal testbed to study
overfitting: an evolution run can discover transformations that improve
the validation windows of an individual series while failing to capture structure that
transfers to held-out windows. 
We evaluate this effect on GIFT-Eval's COVID Deaths task, a daily healthcare benchmark
with 266 univariate series and a 30-step prediction horizon~\citep{aksu2024giftevalbenchmarkgeneraltime}.
The target is confirmed COVID-19 deaths, whose daily counts are noisy and heterogeneous
because of underreporting, cross-country reporting differences, reporting delays, and
day-of-week effects. 

\textbf{Shared vs.\ per-series evolution.} We treat each series as a related task.
The single-task baseline evolves a separate EFE-Time transformation for each series, while
STA Best-Shared evolves one family-level transformation across all series and applies it
directly, with no task-local adaptation. We note that this matches the standard dataset-level evolution in EFE-Time~\citep{taga2026evolutionaryfeatureengineeringstructured}; here, we investigate whether this shared dataset-level evolution improves over evolving transformations separately for each series. We match iteration budgets by running each single-task
evolution for 5 iterations, for $266 \times 5 = 1330$ total iterations, and running STA
Best-Shared for 1330 shared iterations. Each shared candidate is evaluated across all series, so evaluator work is not matched. Because MASE and wQL are lower-is-better error metrics, we report improvement as percentage error reduction relative to the baseline; positive values therefore indicate lower error.

\looseness=-1
\textbf{Validation--test generalization.} Figure~\ref{fig:covid-deaths} shows a clear validation--test gap. Single-task evolution improves validation MASE and wQL by 12.32\% and 12.99\%, respectively, but these gains
do not transfer: test MASE worsens by 1.12\% and test wQL improves by only 2.71\%.
In contrast, STA Best-Shared without adaptation improves test MASE and wQL by 13.24\%
and 32.56\%. Per-series evolution improves validation performance but yields limited or negative gains on the test windows, while the shared transformation improves both test metrics. This is consistent with reduced overspecialization, although the comparison also differs in evaluation workload and per-run search length.
\vspace{-5pt}
\section{Related Work}

\begingroup

\noindent \textbf{LLM-based evolutionary optimization and program search.}
Evolutionary program search has a long history, with genetic programming and Cartesian genetic programming showing that executable or graph-structured programs can be optimized directly against a fitness function~\citep{koza1992genetic,miller2000cartesian}. Recent LLM-based systems revisit this idea by using language models to evolve programs in evaluation-driven optimization loops. FunSearch demonstrated this paradigm for mathematical discovery~\citep{romera-paredes2024funsearch}, while AlphaEvolve broadened it to executable code across mathematical, scientific, and systems problems~\citep{novikov2025alphaevolve}. OpenEvolve provides an open-source implementation of this style of search~\citep{openevolve}.

Related automated heuristic design methods such as EoH and LLaMEA evolve a single heuristic or metaheuristic using performance aggregated over collections of training instances or benchmark problems~\citep{eoh,llamea}, providing a close precedent for our shared family-level objective. EoH-S goes further by evolving a complementary set of heuristics for heterogeneous instances~\citep{eohs}. %while MTHS jointly evolves task-agnostic metaheuristics and task-specific programs across multiple heuristic-design tasks~\citep{mths}. 
GEPA also preserves candidates that excel on different validation instances through Pareto-based selection~\citep{agrawal2025gepa}. EMO-STA differs in its explicit \emph{shared-then-adapt} organization: it first evolves a shared archive under a family-level objective, then uses that archive to initialize further evolution under each target task's objective. This separation lets us study how shared discovery and task-specific adaptation complement one another, including initialization, budget allocation, held-out transfer, and dependence on task relatedness.
\par
\endgroup

\noindent \textbf{Evolutionary multitasking and transfer.}
Evolutionary multitasking studies how related optimization tasks can be solved jointly by transferring information across tasks. Multifactorial evolution formalizes this idea through a shared population and implicit genetic transfer~\citep{gupta2016multifactorial}, while multitask Cartesian genetic programming extends program evolution to multiple related program-synthesis tasks in a shared representation~\citep{scott2017multitask}. A central challenge is controlling transfer so that tasks exchange useful information without inducing negative transfer. MFEA-II addresses this by estimating inter-task transfer parameters online~\citep{bali2020mfea2}, while other approaches learn explicit mappings or hybrid transfer mechanisms between tasks~\citep{feng2019autoencoding,cai2021hybrid}. EMO-STA shares this motivation but studies transfer through an archive of LLM-generated executable programs, with shared search followed by separate task-local evolution. Our controlled task-relatedness experiment examines how the benefit of sharing changes as task objectives become less aligned, including settings with negative transfer.

\noindent \textbf{Warm-starting and transfer in optimization.}
Warm-starting and population seeding provide a direct precedent for EMO-STA's initialization strategies. Prior work studies how initial populations affect evolutionary search and how problem-dependent seeding can improve optimization~\citep{Kazimipour,friedrich2014seedinginitialpopulationmultiobjective}; warm-start CMA-ES similarly exploits information from related optimization problems~\citep{nomura2020warmstartingcmaeshyperparameter}. EMO-STA applies this principle to executable programs: rather than starting each task from scratch, it initializes task-local evolution with candidates from a shared family-level archive. Its initialization strategies vary whether adaptation receives the full archive, its best average candidate, or the candidate that performs best on the target task.

Further discussion of automatic heuristic design, shared representations, Bayesian optimization, and test-time adaptation is provided in Appendix~\ref{sec:further_related}.

\vspace{-5pt}
\section{Discussion}\label{sec:discussion}
%EMO-STA is a framework for LLM-guided evolutionary program discovery that evolves a shared archive across related tasks and uses it to initialize task-local adaptation. Across our experiments, it outperforms iteration-matched single-task evolution in most settings, with \emph{STA Best-Local} strongest in-distribution and \emph{STA Best-Shared} more robust on held-out task sizes. ARC and time-series case studies further suggest that shared evolution can reduce overfitting in low-evidence settings, although our comparisons match evolutionary iterations rather than total computational cost. {EMO-STA requires related tasks with a common executable interface; future work could automate task grouping and allocate shared search by relatedness.}
EMO-STA is a framework for LLM-guided evolutionary program discovery that
evolves a shared archive across related tasks and uses it to initialize
task-local adaptation. Across our experiments, it outperforms
iteration-matched single-task evolution in most settings, with
\emph{STA Best-Local} strongest in-distribution and
\emph{STA Best-Shared} more robust on held-out task sizes. ARC and
time-series case studies further suggest that shared evolution can
reduce overfitting in low-evidence settings, consistent with shared
search favoring programs that generalize across related tasks. These
comparisons match evolutionary iterations rather than total
computational cost.

EMO-STA currently assumes a common executable interface across tasks,
while the benefits of sharing depend on task relatedness. Our controlled
K-module experiment (Section~\ref{sec:task-relatedness}) illustrates that
conflicting objectives can induce negative transfer even when this
interface requirement is satisfied. Future work could automate task grouping and allocate shared versus
task-local search based on relatedness. A broader question is whether
LLMs can infer reusable program structure to guide these decisions,
extending EMO-STA beyond predefined task families while identifying
when independent search is preferable.

\section*{Acknowledgements}

We thank Konstantinos D. Polyzos, Nikhil Vaidyanathan, and Xihe Gu for helpful discussions and feedback. This work is supported in part by the NSF grants CCF-2046816, CCF-2403075, and CCF-2212426; the ONR grants N00014-24-1-2289 and N00014-22-1-2363; the Eric and Wendy Schmidt AI in Science Postdoctoral Fellowship; the NSF TILOS AI Institute; and the UCSD Centers for Machine Intelligence, Computing, and Security (MICS). Computational aspects of the research are generously supported by an Amazon Research Award on Foundation Model Development.
\bibliography{references}
\bibliographystyle{colm2026_conference}

\newpage
\appendix
\clearpage
\section*{Appendix}

This appendix is organized into five parts. Appendix~\ref{app:emo-sta-pseudocode} gives pseudocode for EMO-STA, including the three archive-initialization variants. {Appendix~\ref{app:additional-results} provides expanded in-distribution results, representative evolution trajectories, iteration-budget and OOD evaluations, and additional ablations and baseline comparisons.} Appendix~\ref{app:experimental-details} describes task-family construction, scoring, the COVID Deaths and ARC-AGI case studies, the controlled task-relatedness experiment, and limitations. Appendix~\ref{app:prompts-and-examples} includes representative prompts and program-level examples. Appendix~\ref{sec:further_related} provides further related work on multi-task learning and shared representations, adaptation and test-time learning, and warm-start/transfer methods in black-box optimization.

\section{EMO-STA Algorithm Pseudocode}
\label{app:emo-sta-pseudocode}

\definecolor{warmbg}{RGB}{228,245,233}    % Warmstart
\definecolor{sharedbg}{RGB}{221,235,247}  % Best-Shared
\definecolor{localbg}{RGB}{247,236,221}   % Best-Local

\newcommand{\warmbox}[1]{%
  \begingroup\setlength{\fboxsep}{1pt}\colorbox{warmbg}{\strut #1}\endgroup}
\newcommand{\sharedbox}[1]{%
  \begingroup\setlength{\fboxsep}{1pt}\colorbox{sharedbg}{\strut #1}\endgroup}
\newcommand{\localbox}[1]{%
  \begingroup\setlength{\fboxsep}{1pt}\colorbox{localbg}{\strut #1}\endgroup}

Algorithm~\ref{alg:emo-sta} summarizes EMO-STA. The shared phase is identical
across the three adaptation variants; the only difference is how the
task-local archive is initialized from the shared archive. We therefore color
the three initialization choices separately: \textit{STA Warmstart} (green),
\textit{STA Best-Shared} (blue), and \textit{STA Best-Local} (orange).

\begin{algorithm}[!h]
\caption{EMO-STA (Shared-Then-Adapt)}
\label{alg:emo-sta}
\small
\begin{algorithmic}[1]
\Require Task family $\mathfrak{T}=\{\mathcal{T}_1,\ldots,\mathcal{T}_K\}$,
shared budget $S$, per-task adaptation budget $A$, variant
$v \in \{\textsc{Warmstart},\textsc{Best-Shared},\textsc{Best-Local}\}$
\Ensure Task-specific solutions $\{p_i^\star\}_{i=1}^{K}$

\State Initialize shared archive $\mathcal{A}^{\mathrm{shared}} \gets \varnothing$

\Statex \textbf{Shared evolution}
\For{$t=1,\ldots,S$}
    \State Propose or edit a candidate program $p$ using the LLM and shared-archive context
    \For{$i=1,\ldots,K$}
        \State Evaluate $p$ on task $\mathcal{T}_i$ and cache the task score $s_i(p)$
    \EndFor
    \State Compute the shared score
    \[
    s_{\mathrm{shared}}(p) \gets \frac{1}{K}\sum_{i=1}^{K} s_i(p)
    \]
    \State Insert $p$ into $\mathcal{A}^{\mathrm{shared}}$ together with
    $\{s_i(p)\}_{i=1}^{K}$ and $s_{\mathrm{shared}}(p)$
\EndFor

\Statex \textbf{Task-local adaptation}
\Statex \textit{Projection.} $\Pi_i$ converts a shared archive entry into a
task-local archive entry for $\mathcal{T}_i$, reusing cached task scores when available.

\For{$i=1,\ldots,K$}
    \If{$v=\textsc{Warmstart}$}
        \State \warmbox{$\mathcal{A}^{\mathrm{init}}_i \gets \{\Pi_i(p): p \in \mathcal{A}^{\mathrm{shared}}\}$}
        \Comment{STA Warmstart}
    \ElsIf{$v=\textsc{Best-Shared}$}
        \State \sharedbox{$p_{\mathrm{bs}}^\star \gets \arg\max_{p \in \mathcal{A}^{\mathrm{shared}}} s_{\mathrm{shared}}(p)$}
        \Comment{STA Best-Shared}
        \State \sharedbox{$\mathcal{A}^{\mathrm{init}}_i \gets \{\Pi_i(p_{\mathrm{bs}}^\star)\}$}
    \Else
        \State \localbox{$p_{i,\mathrm{bl}}^\star \gets \arg\max_{p \in \mathcal{A}^{\mathrm{shared}}} s_i(p)$}
        \Comment{STA Best-Local}
        \State \localbox{$\mathcal{A}^{\mathrm{init}}_i \gets \{\Pi_i(p_{i,\mathrm{bl}}^\star)\}$}
    \EndIf

    \State Set the task-local archive $\mathcal{A}_i \gets \mathcal{A}^{\mathrm{init}}_i$
    \For{$a=1,\ldots,A$}
        \State Propose or edit a candidate program using the LLM and task-local archive context
        \State Evaluate on $\mathcal{T}_i$, score with $s_i(\cdot)$, and update $\mathcal{A}_i$
    \EndFor
    \State $p_i^\star \gets \arg\max_{p \in \mathcal{A}_i} s_i(p)$
\EndFor

\State \Return $\{p_i^\star\}_{i=1}^{K}$
\end{algorithmic}
\end{algorithm}

\section{Additional Results}
\label{app:additional-results}

\subsection{Detailed In-Distribution Results}
\label{app:detailed-results}

Tables~\ref{tab:mt-sts-appendix-seed-adaptation-continuous} and
\ref{tab:mt-sts-appendix-seed-adaptation-modeling} provide expanded versions of
the main results, separating each method into its own row, including the
pre-adaptation shared score, and reporting the \textit{Shared / Adapt / Total}
budget configuration for each family. \textit{STA Best-Shared (Before Adaptation)}
denotes the best program produced by the shared evolution before any task-specific
adaptation, so it is the initial program used by \textit{STA Best-Shared}.

\begin{table*}[!h]
\centering
\caption{\small{Comparison of standard single-task and EMO-STA optimization for continuous optimization families. The budget row reports \textit{Shared / Adapt / Total} iterations, where Total is computed as Shared plus the per-task adaptation budget times the number of tasks in the family.}}
\label{tab:mt-sts-appendix-seed-adaptation-continuous}
\scriptsize
\setlength{\tabcolsep}{4pt}
\renewcommand{\arraystretch}{1.08}
\setlength{\aboverulesep}{0.5ex}
\setlength{\belowrulesep}{0.5ex}

\resizebox{\textwidth}{!}{%
\begin{tabular}{llcccc}
\toprule
Model & Method
& Function minimization
& Circle packing
& Circle packing rectangles
& Heilbronn triangle \\
\midrule
\multicolumn{2}{l}{Budget (Shared / Adapt / Total)}
& $40 / 15 / 100$
& $60 / 15 / 120$
& $60 / 15 / 120$
& $60 / 15 / 120$ \\
\midrule

\multirow{5}{*}{\textbf{Haiku-4.5}}
& STA Best-Shared (Before Adaptation) & $.887 \pm .06$ & $.902 \pm .05$ & $.832 \pm .01$ & $.523 \pm .03$ \\
& STA Best-Local & $\mathbf{.952 \pm .04}$ & $.934 \pm .03$ & $.861 \pm .02$ & $\mathbf{.650 \pm .06}$ \\
& STA Warmstart & $.949 \pm .05$ & $.926 \pm .03$ & $\mathbf{.865 \pm .02}$ & $.628 \pm .05$ \\
& STA Best-Shared & $.941 \pm .06$ & $\mathbf{.940 \pm .02}$ & $.845 \pm .01$ & $.628 \pm .06$ \\
& Single-task & $.888 \pm .05$ & $.865 \pm .03$ & $.832 \pm .01$ & $.547 \pm .03$ \\
\midrule

\multirow{5}{*}{\textbf{Sonnet-4.5}}
& STA Best-Shared (Before Adaptation) & $.862 \pm .03$ & $.938 \pm .03$ & $.875 \pm .04$ & $.472 \pm .08$ \\
& STA Best-Local & $\mathbf{.925 \pm .02}$ & $\mathbf{.965 \pm .02}$ & $\mathbf{.898 \pm .03}$ & $\mathbf{.622 \pm .04}$ \\
& STA Warmstart & $.917 \pm .02$ & $.964 \pm .02$ & $.890 \pm .03$ & $.596 \pm .05$ \\
& STA Best-Shared & $.904 \pm .03$ & $.947 \pm .03$ & $.892 \pm .04$ & $.619 \pm .05$ \\
& Single-task & $.891 \pm .05$ & $.927 \pm .02$ & $.840 \pm .02$ & $.548 \pm .04$ \\
\midrule

\multirow{5}{*}{\textbf{Opus-4.5}}
& STA Best-Shared (Before Adaptation) & $.877 \pm .07$ & $.901 \pm .01$ & $.935 \pm .01$ & $.608 \pm .05$ \\
& STA Best-Local & $\mathbf{.969 \pm .03}$ & $\mathbf{.940 \pm .01}$ & $\mathbf{.951 \pm .01}$ & $\mathbf{.741 \pm .03}$ \\
& STA Warmstart & $.942 \pm .07$ & $.926 \pm .01$ & $.943 \pm .01$ & $.732 \pm .04$ \\
& STA Best-Shared & $.941 \pm .09$ & $.930 \pm .01$ & $.943 \pm .01$ & $.704 \pm .04$ \\
& Single-task & $.914 \pm .05$ & $.912 \pm .01$ & $.912 \pm .01$ & $.622 \pm .06$ \\
\midrule

\multirow{5}{*}{\textbf{Sonnet-4.6}}
& STA Best-Shared (Before Adaptation) & $.946 \pm .03$ & $.994 \pm .00$ & $.980 \pm .02$ & $.711 \pm .05$ \\
& STA Best-Local & $.988 \pm .02$ & $\mathbf{.997 \pm .00}$ & $\mathbf{.986 \pm .01}$ & $.862 \pm .04$ \\
& STA Warmstart & $.973 \pm .03$ & $\mathbf{.997 \pm .00}$ & $.985 \pm .01$ & $.809 \pm .04$ \\
& STA Best-Shared & $\mathbf{.991 \pm .02}$ & $\mathbf{.997 \pm .00}$ & $.985 \pm .01$ & $\mathbf{.865 \pm .07}$ \\
& Single-task & $.901 \pm .02$ & $.957 \pm .03$ & $.967 \pm .02$ & $.678 \pm .05$ \\
\midrule

\multirow{5}{*}{\textbf{Opus-4.6}}
& STA Best-Shared (Before Adaptation) & $.922 \pm .04$ & $.960 \pm .02$ & $.941 \pm .01$ & $.784 \pm .03$ \\
& STA Best-Local & $\mathbf{.945 \pm .03}$ & $\mathbf{.984 \pm .01}$ & $\mathbf{.967 \pm .01}$ & $.863 \pm .03$ \\
& STA Warmstart & $.943 \pm .03$ & $.972 \pm .02$ & $.957 \pm .01$ & $.844 \pm .03$ \\
& STA Best-Shared & $.932 \pm .04$ & $.979 \pm .02$ & $.962 \pm .01$ & $\mathbf{.877 \pm .03}$ \\
& Single-task & $.895 \pm .04$ & $.963 \pm .01$ & $.944 \pm .01$ & $.744 \pm .04$ \\

\bottomrule
\end{tabular}%
}
%\vspace{-5pt}
\end{table*}

\begin{table*}[!h]
\centering
\caption{\small{Comparison of standard single-task and EMO-STA optimization for modeling and algorithmic optimization families. The budget row reports \textit{Shared / Adapt / Total} iterations, where Total is computed as Shared plus the per-task adaptation budget times the number of tasks in the family.}}
\label{tab:mt-sts-appendix-seed-adaptation-modeling}
\scriptsize
\setlength{\tabcolsep}{4pt}
\renewcommand{\arraystretch}{1.08}
\setlength{\aboverulesep}{0.5ex}
\setlength{\belowrulesep}{0.5ex}

\resizebox{\textwidth}{!}{%
\begin{tabular}{llcccc}
\toprule
Model & Method
& Signal processing
& SLDBench-3D
& Rust adaptive sort
& K-module \\
\midrule
\multicolumn{2}{l}{Budget (Shared / Adapt / Total)}
& $60 / 10 / 100$
& $60 / 10 / 80$
& $60 / 10 / 100$
& $40 / 20 / 120$ \\
\midrule

\multirow{5}{*}{\textbf{Haiku-4.5}}
& STA Best-Shared (Before Adaptation) & $.568 \pm .04$ & $.936 \pm .02$ & $.509 \pm .03$ & $.392 \pm .05$ \\
& STA Best-Local & $\mathbf{.600 \pm .05}$ & $\mathbf{.958 \pm .02}$ & $.533 \pm .02$ & $.567 \pm .06$ \\
& STA Warmstart & $.584 \pm .06$ & $.953 \pm .02$ & $.535 \pm .02$ & $.567 \pm .04$ \\
& STA Best-Shared & $.597 \pm .04$ & $.949 \pm .02$ & $.509 \pm .03$ & $\mathbf{.575 \pm .07}$ \\
& Single-task & $.569 \pm .01$ & $.951 \pm .01$ & $\mathbf{.539 \pm .02}$ & $.550 \pm .03$ \\
\midrule

\multirow{5}{*}{\textbf{Sonnet-4.5}}
& STA Best-Shared (Before Adaptation) & $.559 \pm .02$ & $.955 \pm .02$ & $.457 \pm .03$ & $.367 \pm .02$ \\
& STA Best-Local & $\mathbf{.587 \pm .01}$ & $\mathbf{.976 \pm .01}$ & $.481 \pm .03$ & $.617 \pm .03$ \\
& STA Warmstart & $.578 \pm .02$ & $.971 \pm .01$ & $.484 \pm .03$ & $\mathbf{.650 \pm .02}$ \\
& STA Best-Shared & $.582 \pm .02$ & $.971 \pm .02$ & $.457 \pm .03$ & $.567 \pm .06$ \\
& Single-task & $.576 \pm .01$ & $.959 \pm .01$ & $\mathbf{.528 \pm .01}$ & $.617 \pm .05$ \\
\midrule

\multirow{5}{*}{\textbf{Opus-4.5}}
& STA Best-Shared (Before Adaptation) & $.612 \pm .03$ & $.959 \pm .02$ & $.483 \pm .05$ & $.442 \pm .02$ \\
& STA Best-Local & $.620 \pm .03$ & $\mathbf{.983 \pm .00}$ & $.515 \pm .05$ & $.617 \pm .03$ \\
& STA Warmstart & $\mathbf{.635 \pm .03}$ & $.972 \pm .01$ & $\mathbf{.520 \pm .05}$ & $\mathbf{.675 \pm .03}$ \\
& STA Best-Shared & $.625 \pm .02$ & $.981 \pm .00$ & $.483 \pm .05$ & $.592 \pm .03$ \\
& Single-task & $.568 \pm .01$ & $.973 \pm .01$ & $.497 \pm .02$ & $.567 \pm .05$ \\
\midrule

\multirow{5}{*}{\textbf{Sonnet-4.6}}
& STA Best-Shared (Before Adaptation) & $.607 \pm .05$ & $.959 \pm .01$ & $.656 \pm .01$ & $.383 \pm .05$ \\
& STA Best-Local & $\mathbf{.628 \pm .04}$ & $\mathbf{.969 \pm .01}$ & $.659 \pm .01$ & $.617 \pm .09$ \\
& STA Warmstart & $.626 \pm .04$ & $.968 \pm .01$ & $\mathbf{.663 \pm .01}$ & $\mathbf{.700 \pm .07}$ \\
& STA Best-Shared & $.613 \pm .05$ & $\mathbf{.969 \pm .01}$ & $.656 \pm .01$ & $.575 \pm .03$ \\
& Single-task & $.608 \pm .03$ & $.955 \pm .01$ & $.616 \pm .03$ & $.675 \pm .05$ \\
\midrule

\multirow{5}{*}{\textbf{Opus-4.6}}
& STA Best-Shared (Before Adaptation) & $.652 \pm .04$ & $.958 \pm .02$ & $.612 \pm .02$ & $.450 \pm .03$ \\
& STA Best-Local & $.713 \pm .05$ & $\mathbf{.975 \pm .01}$ & $.616 \pm .02$ & $.725 \pm .02$ \\
& STA Warmstart & $.707 \pm .04$ & $.973 \pm .01$ & $\mathbf{.625 \pm .02}$ & $\mathbf{.800 \pm .05}$ \\
& STA Best-Shared & $\mathbf{.716 \pm .04}$ & $.967 \pm .01$ & $.612 \pm .02$ & $.692 \pm .05$ \\
& Single-task & $.648 \pm .03$ & $.964 \pm .02$ & $.531 \pm .05$ & $.758 \pm .08$ \\

\bottomrule
\end{tabular}%
}
%\vspace{-5pt}
\end{table*}

\subsection{Representative EMO-STA Trajectories}
\label{app:representative-emosta-trajectories}

We include representative trajectories to show how shared evolution and task-local adaptation contribute over time. These examples are not intended to be exhaustive; they illustrate distinct adaptation modes, from light geometric calibration to broad task-specific improvement and targeted correction of a weak subtask. The \emph{Shared / Adapt / Total} column reports \(S/A/\text{Total}\), where \(S\) is the shared budget, \(A\) is the per-task adaptation budget, and \(\text{Total}=S+KA\) is the matched family-level iteration budget.

\begin{table}[!h]
\centering
\small
\setlength{\tabcolsep}{4pt}
\renewcommand{\arraystretch}{1.10}
\resizebox{\linewidth}{!}{%
\begin{tabular}{llcccc}
\toprule
Example & Model / method & Shared / Adapt / Total & Shared & Adapt & Single-task \\
\midrule
Circle packing
& Haiku-4.5, STA Best-Local
& \(60/15/120\) & \(.833\) & \(.903 \rightarrow .925\) & \(.865\) \\
Circle-packing rectangles
& Sonnet-4.5, STA Best-Shared
& \(60/15/120\) & \(.894\) & \(.894 \rightarrow .924\) & \(.840\) \\
Heilbronn triangle
& Sonnet-4.6, STA Best-Shared
& \(60/15/120\) & \(.750\) & \(.750 \rightarrow .905\) & \(.678\) \\
Signal processing
& Opus-4.6, STA Best-Local
& \(60/10/100\) & \(.619\) & \(.635 \rightarrow .685\) & \(.648\) \\
\bottomrule
\end{tabular}%
}
\vspace{0.4em}
\caption{\small{Representative EMO-STA trajectory examples. The Shared column reports the final family-average score at the end of shared evolution. The Adapt column reports the average score at the start and end of task-local adaptation. Single-task reports the mean over five independent single-task runs for the same family, model, and matched family-level total iteration budget.}}
\label{tab:representative-emosta-trajectories}
\end{table}

\textbf{Discussion.}
The trajectories in \Cref{tab:representative-emosta-trajectories} illustrate two complementary benefits of EMO-STA. In the circle-packing examples, shared evolution already discovers a reusable geometric solver scaffold, and adaptation mainly calibrates it to the selected task sizes or rectangle geometry. In the Heilbronn triangle example, adaptation has a larger effect, improving all subtasks and raising the family average far above the matched single-task baseline. The signal-processing example shows a more targeted correction: the shared program is already competitive on several signal types, while adaptation mainly repairs the step-change task, raising it from \(.694\) to \(.883\) and lifting the family average above the single-task reference. Overall, these trajectories suggest that EMO-STA helps both when shared evolution finds a strong general program that needs light retuning and when adaptation must make a focused task-local correction to a shared scaffold.

\begin{figure}[!h]
    \centering
    \includegraphics[width=\linewidth]{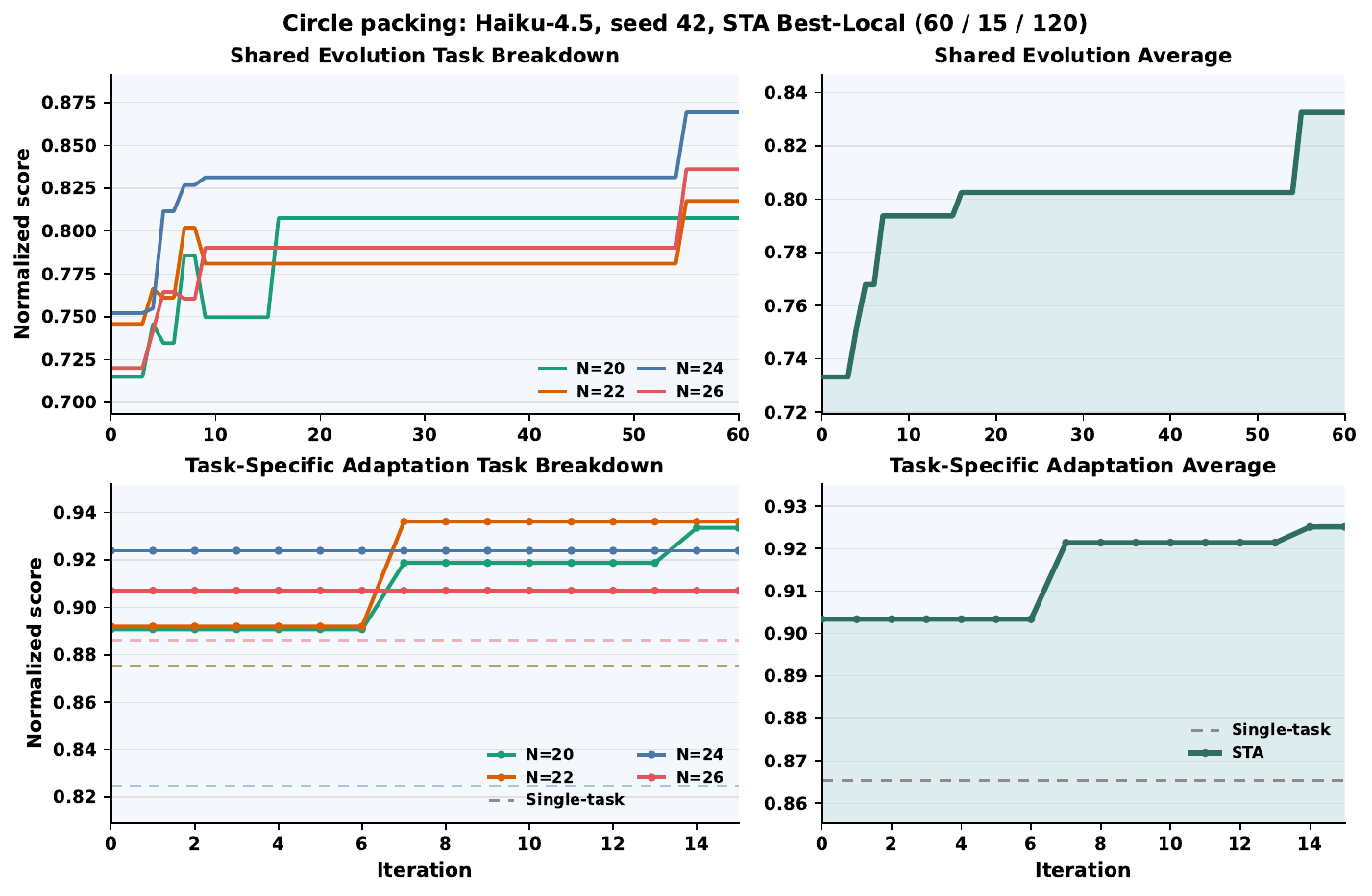}
    \caption{\small{Circle-packing trajectory for Haiku-4.5, seed 42, using \textit{STA Best-Local} with a \(60/15/120\) \(S/A/\mathrm{Total}\) setting. The adapted family average increases from \(.903\) to \(.925\), compared with a five-run Single-task average of \(.865\).}}
    \label{fig:traj-circle-packing-haiku45-bestlocal}
\end{figure}

\begin{figure}[!h]
    \centering
    \includegraphics[width=\linewidth]{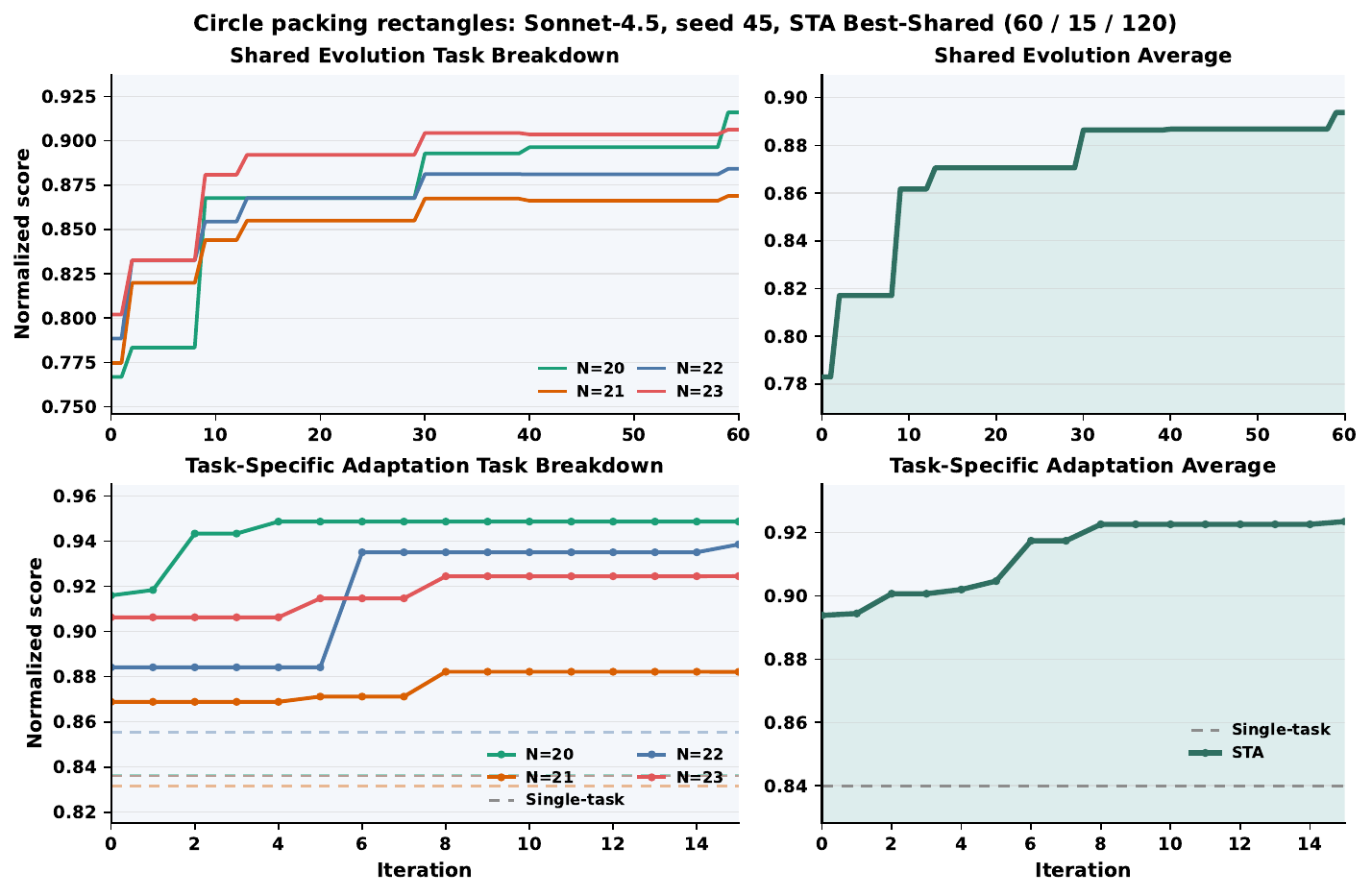}
    \caption{\small{Circle-packing-rectangles trajectory for Sonnet-4.5, seed 45, using \textit{STA Best-Shared} with a \(60/15/120\) \(S/A/\mathrm{Total}\) setting. Adaptation improves the average score from \(.894\) to \(.924\), above the five-run Single-task average of \(.840\).}}
    \label{fig:traj-circle-packing-rectangles-sonnet45-bestshared}
\end{figure}

\begin{figure}[!h]
    \centering
    \includegraphics[width=\linewidth]{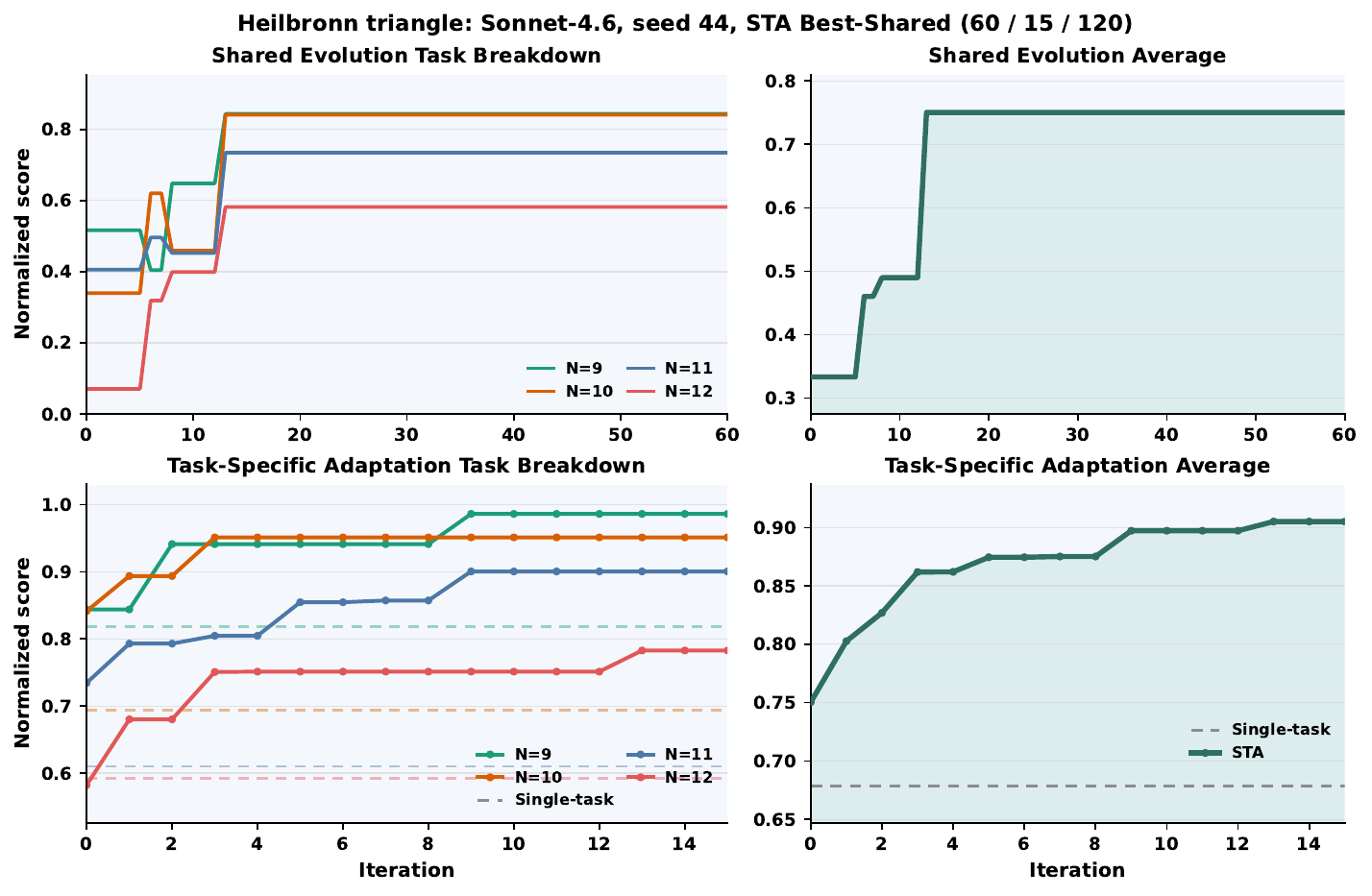}
    \caption{\small{Heilbronn-triangle trajectory for Sonnet-4.6, seed 44, using \textit{STA Best-Shared} with a \(60/15/120\) \(S/A/\mathrm{Total}\) setting. This example shows broad task-specific improvement, with the average score increasing from \(.750\) to \(.905\) versus a five-run Single-task average of \(.678\).}}
    \label{fig:traj-heilbronn-sonnet46-bestshared}
\end{figure}

\begin{figure}[!h]
    \centering
    \includegraphics[width=\linewidth]{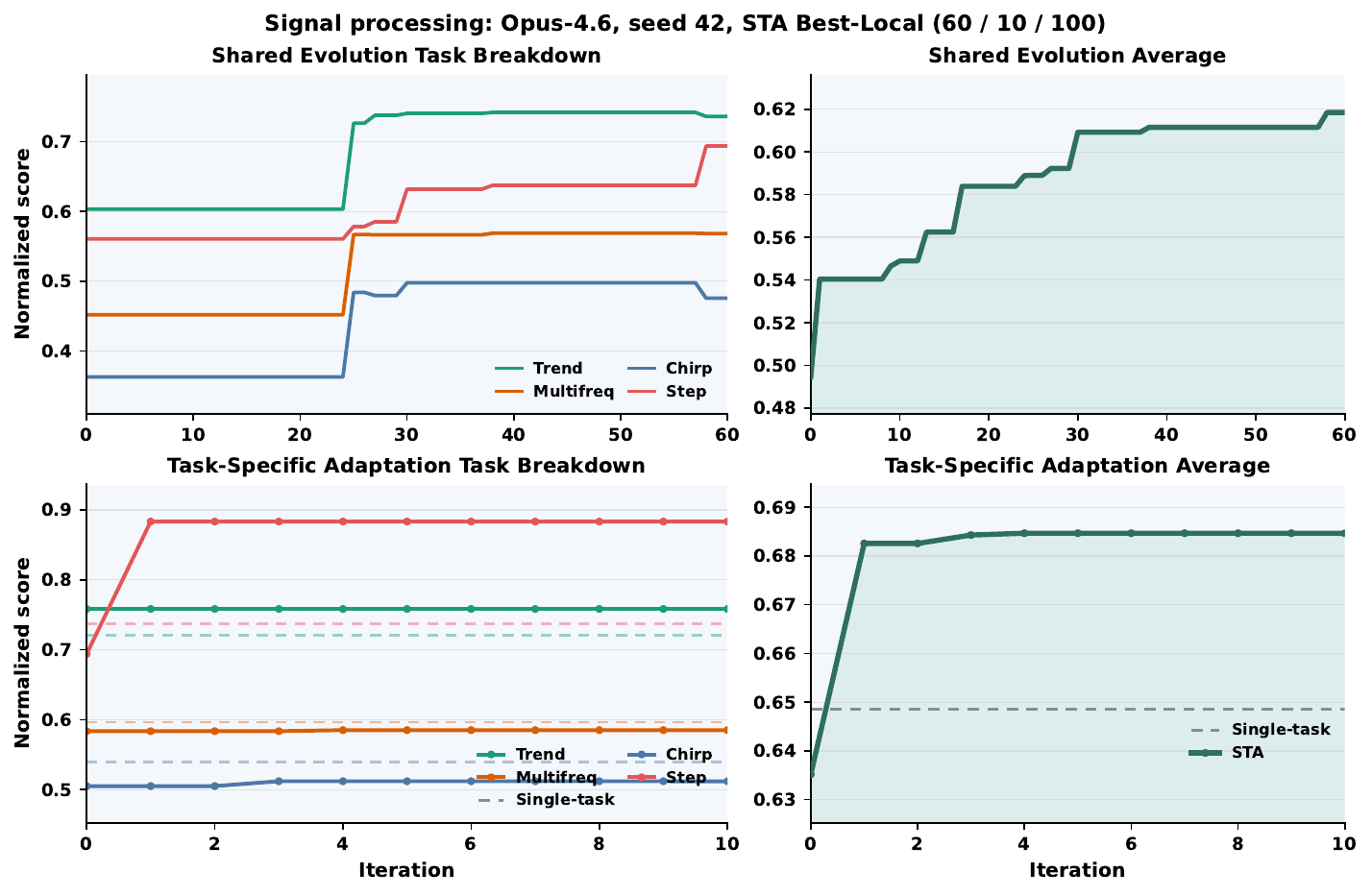}
    \caption{\small{Signal-processing trajectory for Opus-4.6, seed 42, using \textit{STA Best-Local} with a \(60/10/100\) \(S/A/\mathrm{Total}\) setting. Adaptation is concentrated on the step-change task, which improves from \(.694\) to \(.883\), raising the family average from \(.635\) to \(.685\) above the five-run Single-task average of \(.648\).}}
    \label{fig:traj-signal-processing-opus46-bestlocal}
\end{figure}

\subsection{Additional Out-of-Distribution Evaluation Results}
\label{app:additional-ood-evaluation-details}
We include the full OOD holdout results for the three geometric optimization families. All programs are selected by their in-distribution runs and then evaluated on held-out task sizes without rerunning evolution.

\begin{figure}[!h]
    \centering
    \includegraphics[width=0.74\linewidth]{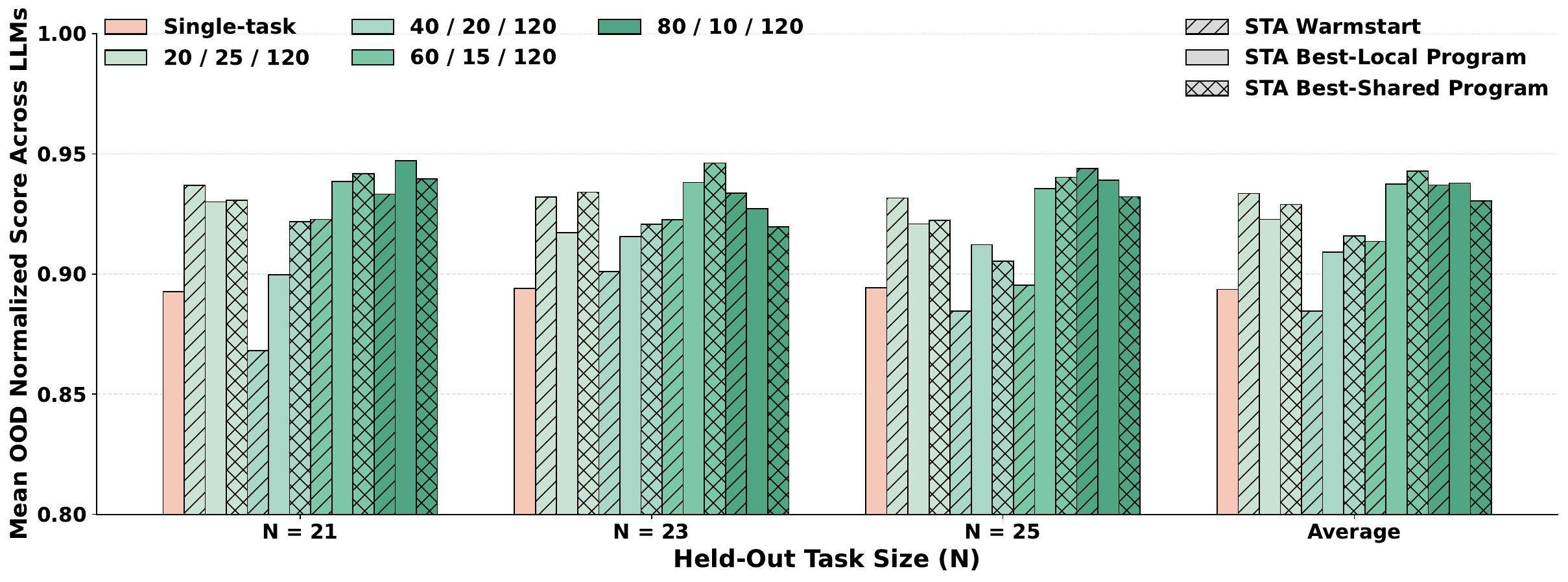}
    \caption{OOD holdout evaluation for circle packing across EMO-STA budget allocations with the single-task baseline fixed at 120 total iterations. The x-axis shows held-out task sizes plus the average across holdouts. The peach bars show the fixed single-task baseline, green colors denote the \textit{Shared / Per-task adaptation / Total} budget allocation, and hatch patterns denote the STA adaptation variant. Bars report mean OOD normalized score across LLMs.}
    \label{fig:circle-packing-ood-b30-holdout-seed-adaptation}
\end{figure}

\begin{figure}[!h]
    \centering
    \includegraphics[width=0.74\linewidth]{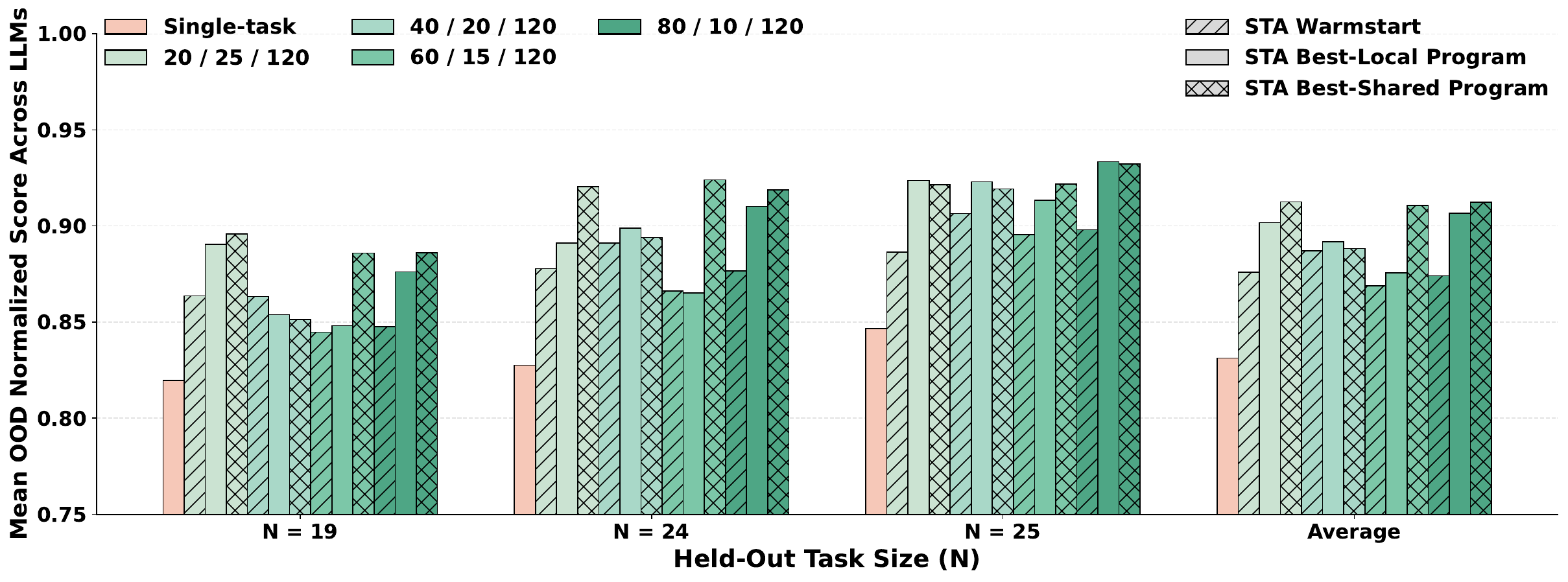}
    \caption{OOD holdout evaluation for circle packing in rectangles across EMO-STA budget allocations with the single-task baseline fixed at 120 total iterations. The x-axis shows held-out task sizes plus the average across holdouts. The peach bars show the fixed single-task baseline, green colors denote the \textit{Shared / Per-task adaptation / Total} budget allocation, and hatch patterns denote the STA adaptation variant. Bars report mean OOD normalized score across LLMs.}
    \label{fig:circle-packing-rectangle-ood-b30-holdout-seed-adaptation}
\end{figure}

\begin{figure}[!h]
    \centering
    \includegraphics[width=0.74\linewidth]{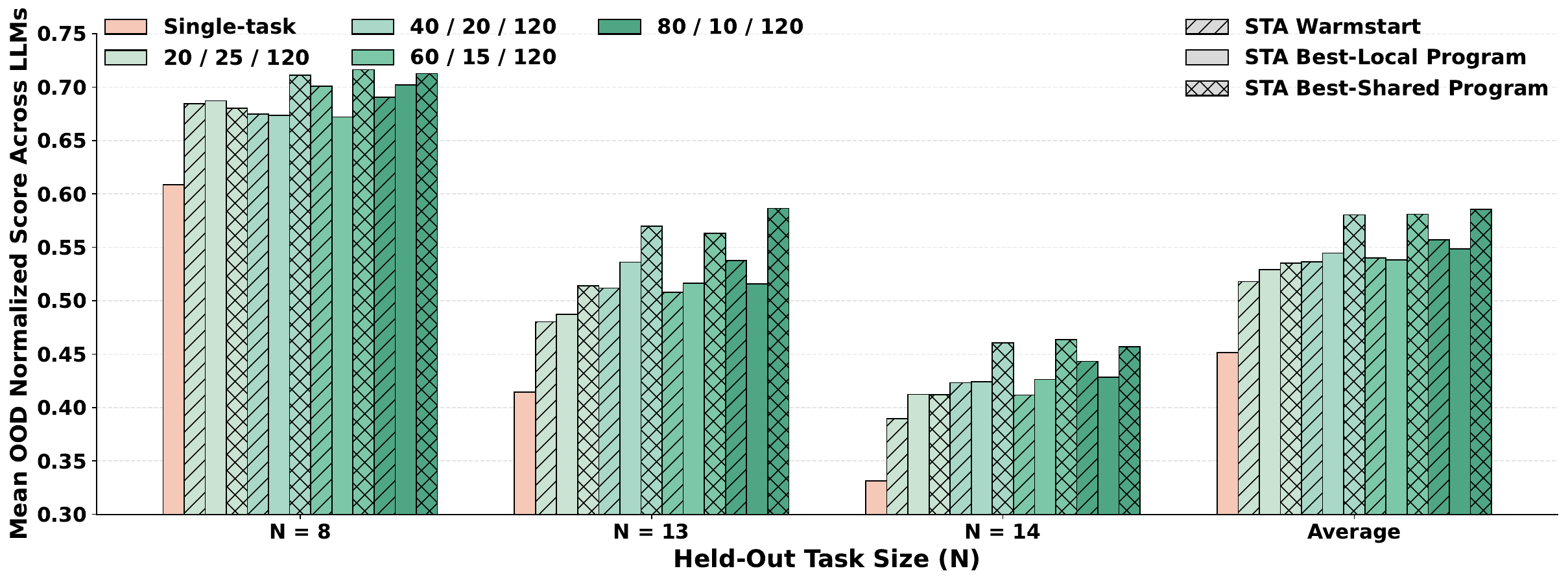}
    \caption{OOD holdout evaluation for the Heilbronn triangle task across EMO-STA budget allocations with the single-task baseline fixed at 120 total iterations. The x-axis shows held-out task sizes plus the average across holdouts. The peach bars show the fixed single-task baseline, green colors denote the \textit{Shared / Per-task adaptation / Total} budget allocation, and hatch patterns denote the STA adaptation variant. Bars report mean OOD normalized score across LLMs.}
    \label{fig:heilbronn-triangle-ood-b30-holdout-seed-adaptation}
\end{figure}

\textbf{Discussion on the additional results.} As shown in \Cref{fig:circle-packing-ood-b30-holdout-seed-adaptation,fig:circle-packing-rectangle-ood-b30-holdout-seed-adaptation,fig:heilbronn-triangle-ood-b30-holdout-seed-adaptation}, held-out transfer is robust across iteration-budget allocation settings, but that the three adaptation strategies have different OOD profiles. Across all three geometric families, EMO-STA variants generally outperform the direct single-task baseline on held-out task sizes, indicating that the shared phase learns reusable geometric structure rather than only improving the in-distribution tasks. \emph{STA Best-Shared} is the most reliable OOD strategy overall, especially for circle packing in rectangles and Heilbronn triangle, where the globally best shared program transfers better than task-specialized adaptations across most held-out sizes. This suggests that, when the held-out task may differ from any one training task, the shared-average solution can preserve more family-level robustness. \emph{STA Warmstart} is particularly competitive for unit-square circle packing, where retaining a diverse archive of layouts helps across neighboring circle counts. \emph{STA Best-Local} remains strong in several settings, but its advantage is more tied to proximity between the adaptation task and the held-out size, so it is most useful when the target task is close to the evaluation size. Together, these patterns indicate that shared evolution improves transfer beyond the training task sizes.

\subsection{Additional Iteration-Budget Allocation Results}
\label{app:additional-compute-allocation-results}

We report additional iteration-budget allocation results for Heilbronn triangle and the two circle-packing families, complementing the main-text results on Heilbronn triangle and function minimization.

\begin{figure}[!h]
  \centering

  \begin{center}
  \small
  \legendboxsingle\ Single-task
  \hspace{1.0em}
  \legendboxwarmstart\ STA Warmstart
  \hspace{1.0em}
  \legendboxbestlocal\ STA Best-Local
  \hspace{1.0em}
  \legendboxbestshared\ STA Best-Shared
  \end{center}

  \includegraphics[width=0.7\linewidth]{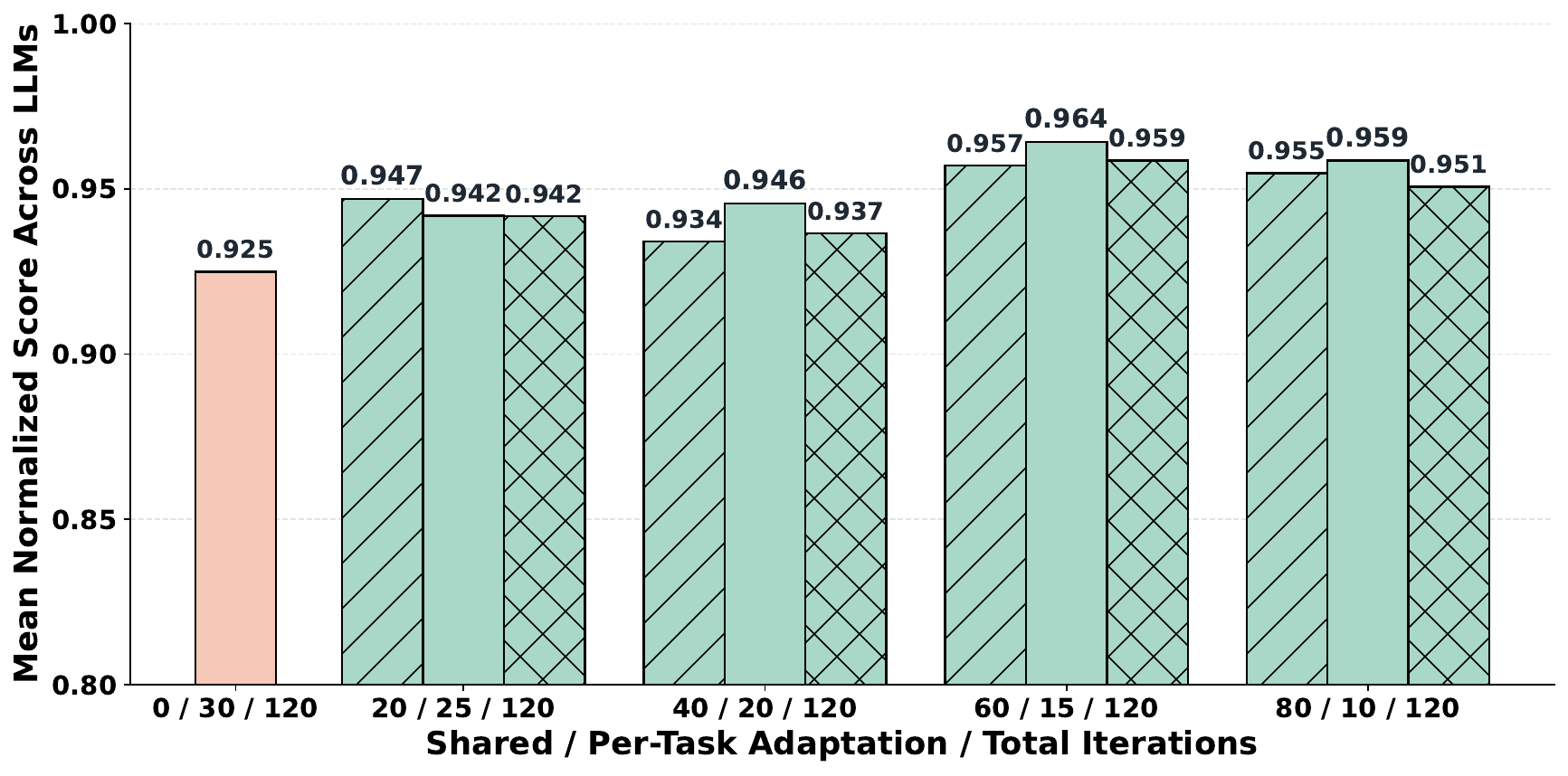}
  \caption{
  Iteration-budget allocation results for EMO-STA on \textit{circle packing}, with the single-task baseline fixed at \(B=30\) per task, corresponding to \(KB=120\) total iterations.
  Grouped bars show \textit{STA Warmstart}, \textit{STA Best-Local}, and \textit{STA Best-Shared} under different \emph{Shared / Per-task Adapt / Total} allocations. Here, \emph{Shared} is the family-level shared budget \(S\), \emph{Adapt} is the per-task adaptation budget \(A\), and \emph{Total} is \(S + K A\).
  The leftmost bar is the direct single-task baseline, with allocation \(0 / 30 / 120\).
  Bars report the mean score averaged over the five models Claude Haiku-4.5, Sonnet-4.5, Sonnet-4.6, Opus-4.5, and Opus-4.6.
  }
  \label{fig:circle-packing-fixed-b30-seed-adaptation-budget-sweep}
\end{figure}

\begin{figure}[!h]
    \centering

    \begin{center}
    \small
    \legendboxsingle\ Single-task
    \hspace{1.0em}
    \legendboxwarmstart\ STA Warmstart
    \hspace{1.0em}
    \legendboxbestlocal\ STA Best-Local
    \hspace{1.0em}
    \legendboxbestshared\ STA Best-Shared
    \end{center}

    \includegraphics[width=0.7\linewidth]{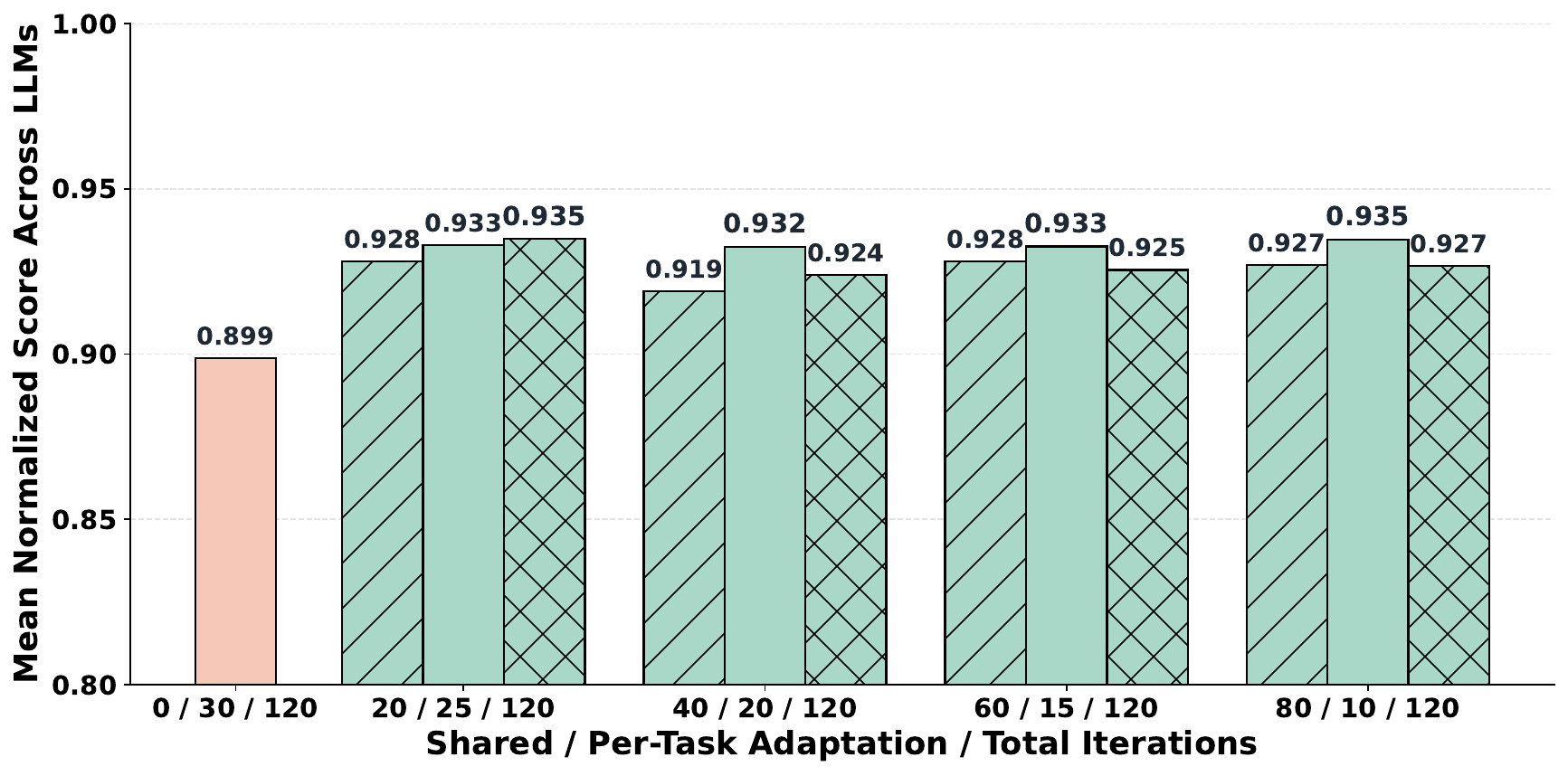}
    \caption{
    Iteration-budget allocation results for EMO-STA on \textit{circle packing in rectangles}, with the single-task baseline fixed at \(B=30\) per task, corresponding to \(KB=120\) total iterations.
    Grouped bars show \textit{STA Warmstart}, \textit{STA Best-Local}, and \textit{STA Best-Shared} under different \emph{Shared / Per-task Adapt / Total} allocations. Here, \emph{Shared} is the family-level shared budget \(S\), \emph{Adapt} is the per-task adaptation budget \(A\), and \emph{Total} is \(S + K A\).
    The leftmost bar is the direct single-task baseline, with allocation \(0 / 30 / 120\).
    Bars report the mean score averaged over the five models Claude Haiku-4.5, Sonnet-4.5, Sonnet-4.6, Opus-4.5, and Opus-4.6.
    }
    \label{fig:circle-packing-rectangle-fixed-b30-seed-adaptation-budget-sweep}
\end{figure}

\begin{figure}[!h]
    \centering

    \begin{center}
    \small
    \legendboxsingle\ Single-task
    \hspace{1.0em}
    \legendboxwarmstart\ STA Warmstart
    \hspace{1.0em}
    \legendboxbestlocal\ STA Best-Local
    \hspace{1.0em}
    \legendboxbestshared\ STA Best-Shared
    \end{center}

    \includegraphics[width=0.82\linewidth]{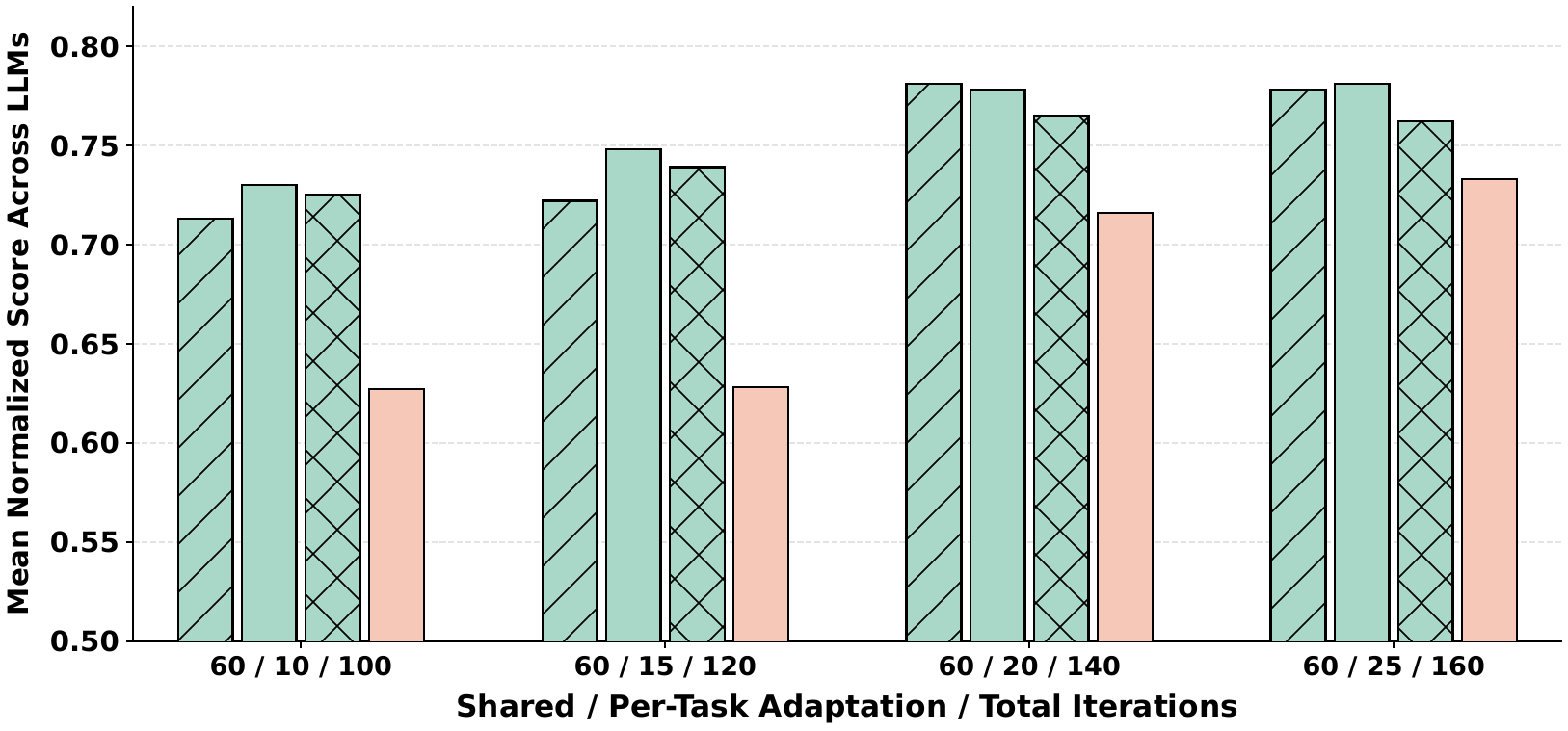}
    \caption{
    Iteration-budget allocation results for the Heilbronn triangle family with the shared budget fixed at \(S=60\).
    The x-axis reports \textit{Shared / Per-task Adapt / Total} iterations. For each total budget, the direct single-task baseline uses the corresponding matched per-task budget \(B\), so that \(S + K A = K B\).
    Bars report mean normalized score across LLMs and seeds for \textit{STA Warmstart}, \textit{STA Best-Local}, \textit{STA Best-Shared}, and direct single-task optimization.
    }
    \label{fig:heilbronn-public-task-budget-scaling-s60}
\end{figure}

\textbf{Discussion on the additional results.}
As shown in \Cref{fig:heilbronn-public-task-budget-scaling-s60,fig:circle-packing-fixed-b30-seed-adaptation-budget-sweep,fig:circle-packing-rectangle-fixed-b30-seed-adaptation-budget-sweep}, shared evolution remains beneficial across both ways of varying the iteration budget: increasing the total budget after a fixed shared phase, and changing the shared/adaptation split under a fixed total budget. For Heilbronn triangle, increasing the total budget improves the direct single-task baseline, but the EMO-STA variants remain consistently stronger; the gap is largest at lower budgets, where the fixed shared phase provides useful geometric structure before task-specific adaptation. The two circle-packing families show the same qualitative advantage under a fixed total iteration budget. The unit-square circle-packing family benefits most from a balanced allocation, with the strongest result at \(60/15/120\), while circle packing in rectangles is less sensitive to the exact allocation, suggesting that shared evolution quickly discovers reusable geometry-aware structure and adaptation mainly refines it for each task size and rectangle aspect ratio. Across these results, \emph{STA Best-Local} is usually strongest or competitive, while \emph{STA Warmstart} and \emph{STA Best-Shared} also remain above the single-task baseline, reinforcing that the gain comes from shared evolution rather than a single initialization choice.

\begingroup
\subsection{Ablations and Additional Baselines}
\label{app:search-ablations}

We examine whether shared evolution followed by adaptation improves
over using the full budget for shared search or adapting programs
found through single-task search. These experiments use Qwen3.8-27B
on signal processing and the original fixed-target K-module family
described in Appendix~\ref{app:task-family-design}. The K-module
setup is separate from the task-relatedness experiment in
Appendix~\ref{app:k-module-relatedness}.

\begin{table}[htbp]
\centering
\small
\caption{
Ablations and additional baselines with Qwen3.8-27B using the native
OpenEvolve workflow. Results are mean task scores $\pm$ sample
standard deviation across three search seeds (42, 43, and 44),
after averaging four tasks within each seed. Higher is better.
Matched family-level budgets are 100 iterations for signal
processing and 120 for K-module, per seed.
One-then-adapt also averages uniformly over four source choices
within each seed; its budget is per source choice.
Bold denotes the highest matched-budget mean.
}
\label{tab:search-ablations}
\begin{tabular}{lcc}
\toprule
Strategy & Signal processing & K-module \\
\midrule
Single-task
& $0.5170 \pm 0.0059$ & $0.5139 \pm 0.0636$ \\
Shared-only, Best-Shared
& $0.5317 \pm 0.0081$ & $0.4583 \pm 0.0000$ \\
Shared-only, Best-Local
& $0.5401 \pm 0.0029$ & $0.5556 \pm 0.0241$ \\
One-then-adapt
& $0.5283 \pm 0.0070$ & $0.5347 \pm 0.0514$ \\
Single-task pool then adapt
& $0.5434 \pm 0.0261$ & $0.4861 \pm 0.0481$ \\
STA Best-Local
& $\mathbf{0.5550 \pm 0.0088}$ & $0.6111 \pm 0.0481$ \\
STA Warmstart
& $0.5443 \pm 0.0065$ & $\mathbf{0.6250 \pm 0.0417}$ \\
\bottomrule
\end{tabular}
\end{table}

\noindent\textbf{Shared-then-adapt.}
STA Best-Local and STA Warmstart begin from the same shared
checkpoint after \(S\) iterations. Best-Local initializes each
target search with the program that has the highest target-task
score in that checkpoint. Warmstart instead transfers the retained
shared population into a new task-specific database, replacing
shared scores with target-task scores. Cached task scores are
reused, and missing scores are reevaluated. Each target then
receives \(A\) adaptation iterations.

\noindent\textbf{Shared evolution without adaptation.}
The shared-only strategies continue the same shared checkpoint
until the full matched budget has been spent on shared search.
Best-Shared selects one program with the highest average task
score, while Best-Local selects the highest-scoring program for
each task separately. Neither strategy performs further evolution
after selection.

\noindent\textbf{Adaptation from single-task searches.}
One-then-adapt first evolves programs on one source task for
\(S\) iterations. We then evaluate the retained programs on every
target, select the best program for each target, and adapt it
for \(A\) iterations. We repeat this procedure for all four source
choices and average their results uniformly rather than selecting
the best source.

Single-task pool then adapt instead runs four independent
single-task searches, each for \(S/4\) iterations, and pools their
retained programs. It then follows the same target-specific
selection and adaptation procedure. Both baselines initialize each
adaptation from one program, making STA Best-Local the direct
comparison with the same initialization size and adaptation budget.

\noindent\textbf{Iteration budgets.}
Signal processing uses \(S=60\) shared or source-task iterations
and \(A=10\) adaptation iterations per task; K-module uses
\(S=40\) and \(A=20\). The total budgets are therefore
\(S+4A=100\) and \(120\), respectively.
Single-task uses \(25\) and \(30\) iterations per task, while
the pool's initial searches use \(15\) and \(10\) iterations
per task before adaptation.

Each source choice in One-then-adapt receives the full matched
family-level budget. Budgets count native evolution iterations.
Candidate selection and cross-task evaluation require no additional
program generations, but evaluation costs and token usage are
not equalized.

\noindent\textbf{Results.}
Table~\ref{tab:search-ablations} shows that STA achieves the highest
matched-budget mean in both families: Best-Local on signal
processing and Warmstart on K-module. STA Best-Local has a higher
mean than both single-task transfer baselines while using the same
single-program initialization and per-task adaptation budget.
This suggests that shared search provides better starting programs
than transferring candidates found through independent searches
in these settings. Both STA variants also achieve higher means
than spending the full matched budget on shared search alone.
\par
\endgroup

\section{Experimental Details}
\label{app:experimental-details}

\subsection{Task-Family Construction, Scoring, and Iteration Budgets}
\label{app:task-family-design}

For every task family, we converted an existing OpenEvolve example or closely related benchmark setup into a small collection of related subtasks that share one evolving artifact, one evaluator family, and the same shared-then-adapt workflow~\citep{openevolve}. In the shared phase, the evolving program is optimized against the average score across the family. We then initialize task-specific continuations from the shared archive and compare them against direct single-task baselines with matched interfaces and family-level iteration budgets. The main design goal is to expose reusable structure that can be discovered during shared optimization, while still leaving enough task-specific variation for adaptation to matter. Below, we first describe shared implementation and budget details, then give the task-specific construction and scoring rules for each benchmark family.

\subsubsection{Shared implementation and budget details}

\textbf{OpenEvolve configuration.} The underlying OpenEvolve hyperparameters and run configurations mostly follow the original OpenEvolve example settings, while EMO-STA introduces task-family-specific interfaces, evaluators, and shared/adaptation budget schedules. The exact configuration files used for each experiment are included in the supplementary code.

\noindent \textbf{LLM access and hardware.} Claude-family models were accessed through Amazon Bedrock, and Gemini models through the Gemini Developer API via Google AI Studio rather than Vertex AI. Qwen3.8-27B was run locally on NVIDIA L40S GPUs. Full model identifiers and sampling parameters are provided in the supplementary code.

\subsubsection{Task families}

\textbf{Function minimization.}
We adapted the original standalone function-minimization example from OpenEvolve~\citep{openevolve} into a four-task family of public two-dimensional objectives commonly used in global-optimization benchmarks~\citep{jamil2013benchmark}.
\begin{itemize}
  \item \textbf{Oscillatory Basin} (original example) preserves the spirit of the original sinusoidal landscape, with smooth periodic structure and multiple local basins.
  \item \textbf{Ackley} adds a broad, nearly flat outer region with a sharp central basin.
  \item \textbf{Rastrigin} introduces a highly multimodal landscape with many regularly spaced local optima.
  \item \textbf{Rosenbrock} tests narrow-valley optimization with strong variable coupling.
\end{itemize}
Rather than allowing the evolving code to adapt to a single named landscape, the EMO-STA version requires one generic derivative-free optimizer that receives only an opaque \texttt{objective\_fn} and \texttt{bounds} from the evaluator. The benchmark functions are translated but not rescaled, and the task name and optimum are hidden from the candidate. Bounds are \([-5,5]^2\) for Oscillatory Basin and Ackley, \([-5.12,5.12]^2\) for Rastrigin, and \([-3,3]^2\) for Rosenbrock. The full evaluator calls each candidate with \texttt{iterations=200} and five deterministic seeds \((0,\ldots,4)\), while the cheaper cascade stage uses \texttt{iterations=50} and two seeds. For each seed, the evaluator recomputes the true objective value at the returned point and rejects points outside bounds. The task score is
\[
0.50(1+\Delta f)^{-1}+0.35(1+d)^{-1}+0.15\rho,
\]
where \(\Delta f\) is the nonnegative gap to the task optimum, \(d\) is Euclidean distance to the translated optimum, and \(\rho\) is the successful-trial fraction. 

\textbf{Signal processing.}
We adapted the original OpenEvolve signal-processing benchmark, which evaluated a single algorithm across several synthetic signal types, into four explicit EMO-STA subtasks~\citep{openevolve}:
\begin{itemize}
  \item \textbf{Trend+sine} combines a smooth global trend with a periodic component.
  \item \textbf{Multifrequency} superposes multiple sinusoidal components at different frequencies.
  \item \textbf{Chirp} uses a sinusoid whose frequency changes over time.
  \item \textbf{Step changes} contains abrupt piecewise shifts in the underlying signal level.
\end{itemize}
All tasks use the same causal interface, \texttt{process\_signal(noisy\_signal, window\_size)}, with \(\texttt{window\_size}=20\). The task lengths are \(500,600,700,800\), with Gaussian noise standard deviations \(0.2,0.3,0.4,0.5\), respectively. Full evaluation uses three fixed noisy realizations per task, with seeds \(0,1,2\), while the cascade stage uses seed \(0\). The clean signal and noisy observation are deterministic functions of the task and seed, so evaluation is fixed across runs. The candidate observes only the noisy input signal and must return a one-dimensional filtered signal of length \(\texttt{len(noisy)}-\texttt{window\_size}+1\); it never sees the clean target, task identifier, or generating formula. The evaluator measures slope reversals, recent lag error, average tracking error against the aligned noisy signal, false reversals against the clean signal, correlation with the clean signal, and noise reduction. The final task score is a normalized composite in \([0,1]\), combining denoising quality, smoothness, clean-signal correlation, noise reduction, and success rate; failures or timeouts receive zero. We excluded the random-walk case from the EMO-STA family so that the shared tasks remain closely related while still exhibiting distinct denoising and trend-recovery behavior.

\textbf{Circle packing.}
We adapted the unit-square circle-packing benchmark with fixed \(n=26\), used in AlphaEvolve and implemented in OpenEvolve, into a task family with similar circle counts~\citep{novikov2025alphaevolve,openevolve}. The EMO-STA training tasks use \(n \in \{20,22,24,26\}\), all through a single \texttt{construct\_packing(n)} or \texttt{run\_packing(n)} program interface, so the evolved code must implement a reusable packing strategy rather than adapt to one fixed problem size. The evaluator expects centers of shape \((n,2)\) and radii of shape \((n,)\), and independently validates finite nonnegative radii, containment in \([0,1]^2\), and pairwise non-overlap with tolerance \(10^{-6}\). Invalid packings, execution failures, and timeouts receive score zero. Valid packings are scored by the normalized ratio \(\texttt{sum\_radii}/\texttt{target\_sum\_radii}\), using  targets \(2.301,2.420,2.530,2.635\) for \(n=20,22,24,26\), taken from public best-known circle-packing tables~\citep{friedman_circles_in_squares}. The ratio is not capped, so a packing exceeding the reference can score above one. We also define evaluation-only holdouts at \(n \in \{21,23,25\}\), with targets \(2.362,2.478,2.587\), to test whether the shared representation transfers to unseen but nearby circle counts. This normalization is necessary because absolute radius sums are not directly comparable across different \(n\): larger circle counts naturally allow larger total sums, so averaging raw sums would bias shared evolution toward the larger-\(n\) tasks. 

\textbf{Circle packing in rectangles.}
We also define a second circle-packing EMO-STA family that keeps the same shared constructive structure but changes the container geometry to the perimeter-4 rectangle setting~\citep{friedman_circles_in_rectangles}. In this family, the evolving program still implements \texttt{construct\_packing(n)} or \texttt{run\_packing(n)}, but it must return centers, radii, and a rectangle width \(\alpha\). The evaluator sets the height to \(2-\alpha\), so the rectangle perimeter is \(4\), and requires \(0<\alpha\leq 1\), finite nonnegative radii, containment in \([0,\alpha]\times[0,2-\alpha]\), and pairwise non-overlap. The four public training tasks use \(n \in \{20,21,22,23\}\), with target sums \(2.305,2.365,2.425,2.484\). Valid packings are scored by the uncapped ratio between the summed radius and the task-specific target. Evaluation-only OOD rectangle tasks use \(n\in\{19,24,25\}\), with targets \(2.241,2.535,2.592\). As in unit-square circle packing, normalization is essential because attainable absolute sums vary with both \(n\) and the geometry of the best rectangle, so raw summed radii do not provide a useful family-level signal for shared evolution.

\textbf{Heilbronn triangle.}
We adapted the Heilbronn triangle benchmark into a four-task EMO-STA family over nearby point counts inside one fixed canonical unit-area triangle with vertices \((0,0)\), \((2,0)\), and \((0,1)\), using known small-\(n\) reference values for normalization~\citep{weisstein_heilbronn_triangle}. The evolving code uses a generic \texttt{construct\_points(n)} or \texttt{run\_heilbronn(n)} interface and must maximize the minimum triangle area induced by all triples of points. The evaluator validates shape \((n,2)\), finiteness, and containment in the canonical triangle, and then exactly computes the minimum area over all point triples. Invalid outputs receive score zero. Valid outputs are scored as \(\texttt{min\_triangle\_area}/\texttt{target\_min\_area}\), using references \(0.0548469,0.0433767,0.0360927,0.0310048\) for \(n=9,10,11,12\). Evaluation-only OOD tasks use \(n\in\{8,13,14\}\), with references \(0.0677891,0.0245643,0.0237758\). The normalization is important because the attainable optimum changes substantially with \(n\): as more points are packed into the same canonical triangle, the best achievable minimum area decreases, so averaging raw areas would bias the shared objective toward easier smaller-\(n\) tasks. The normalized objective instead puts the training tasks on a common scale and encourages the shared phase to learn reusable geometric placement structure across nearby sizes.

\textbf{K-module.}
We adapted the original public OpenEvolve 4-module, 5-option K-module problem into a harder hidden-family EMO-STA benchmark~\citep{openevolve}. The EMO-STA version uses six named modules with six opaque options each, giving \(6^6\) possible configurations, and defines four hidden target tasks. A candidate must return one complete configuration through \texttt{run\_pipeline()} or \texttt{configure\_pipeline()}; malformed configurations or invalid option names receive zero. Each hidden task has a fixed target configuration, and the task score is the fraction of modules matched, \(\texttt{correct\_modules}/6\). Thus, a task-local score of \(2/3\) means four of six modules match the hidden target. The four hidden targets are deterministic and fixed across all seeds and runs. They are included in the supplementary evaluator code for reproducibility, but the prompts and public artifacts expose only the module names and option counts. The main comparisons use the \emph{balanced} construction: each module has exactly one agreeing task pair, with the remaining two tasks using distinct target options, and each of the six task pairs agrees on exactly one module. This yields zero pairwise score correlation over all configurations, despite partial overlap among task solutions. Choosing the agreeing option at every module gives a shared consensus with score \(1/2\) on each task, whereas each task-specific optimum has score \(1\). Appendix~{\ref{app:k-module-relatedness}} describes the related and conflicting variants.

\textbf{SLDBench-3D.}
We adapted SLDBench, a benchmark for automated scaling-law discovery, into a two-task EMO-STA subset containing \texttt{vocab\_scaling\_law} and \texttt{data\_constrained\_scaling\_law}~\citep{lin2026languagemodelsdiscoverscaling}. Both tasks are loaded from \texttt{pkuHaowei/sldbench} and use the provided train/test splits. They are 3D scalar-loss scaling-law tasks with a 7-parameter cap, making them close enough for transfer while still non-identical. We canonicalize both tasks to the same three-column schema, \([\texttt{model\_size\_like}, \texttt{diversity\_like}, \texttt{total\_data\_like}]\): vocabulary scaling uses \([\texttt{non\_vocab\_parameters}, \texttt{vocab\_size}, \texttt{num\_characters}]\), while data-constrained scaling maps \(\texttt{params}\), \(\texttt{unique\_tokens}\), and \(\texttt{tokens}\) into the canonical schema. For each group, the evaluator fits coefficients on the train split by calling \texttt{fit\_scaling\_law}, and then evaluates \texttt{scaling\_law\_func} on the corresponding held-out test group. The fitted parameter vector must contain between one and seven finite parameters, so EMO-STA shares the law form and fitting routine but refits coefficients locally for each group. The score is \(1/(1+\mathrm{NMSE})\), where NMSE is test MSE normalized by the variance of the held-out targets. This preserves the intended scaling-law structure while making the family amenable to shared optimization.

\textbf{Rust adaptive sort.}
We adapted the original standalone OpenEvolve Rust sorting benchmark into four deterministic regimes: random, nearly sorted, reverse sorted, and duplicates~\citep{openevolve}. The EMO-STA evaluator copies each candidate into a temporary Cargo project, compiles it once in release mode, and then benchmarks the compiled binary on the selected task regime, so one \texttt{adaptive\_sort} implementation is reused across all tasks. Compile failures, runtime failures, malformed output, and timeouts receive zero. Random, nearly sorted, and duplicate tasks use array sizes \(1000\) and \(10000\) with seeds \(0,1,2\); reverse-sorted tasks use the same sizes without random seeds. Nearly sorted arrays use \(5\%\) random swaps, and duplicate arrays use 10 unique values at size \(1000\) and 100 unique values at size \(10000\). Each dataset is checked for exact sorted correctness against Rust \texttt{sort\_unstable}; any task with less than \(100\%\) correctness receives score zero. For correct tasks, the evaluator compares candidate median runtime to the median runtime of \texttt{sort\_unstable}, measured in the same binary after one warmup and five benchmark repetitions. The score is \(0.8s+0.2c\), where \(s=\texttt{mean\_speedup}/(1+\texttt{mean\_speedup})\) rewards speed and \(c=1/(1+\mathrm{CV})\) rewards consistency across datasets. This forces the shared phase to discover broadly useful control logic for switching among sorting behaviors, while still allowing task-specific adaptation to calibrate the implementation to one regime. We intentionally omitted the original \texttt{partially\_sorted} regime from the initial family and reserve it as a possible holdout or generalization check.

\phantomsection
\label{app:iteration-accounting}
\textbf{Iteration-budget accounting and evaluation overhead.}
{Our main comparisons match family-level evolutionary-iteration budgets, \(S+KA=KB\).} An iteration proposes or edits one candidate program and
evaluates it; it is not a fixed unit of total computational cost. In
nominal full-evaluation accounting, excluding initialization, archive
projection, retries, and evaluation-only holdouts, the numbers of
candidate--task evaluations are
\[
N_{\mathrm{STA}}=KS+KA,\qquad
N_{\mathrm{Single\text{-}task}}=KB.
\]
The shared phase evaluates each candidate on \(K\) tasks, whereas a local
iteration evaluates only one task. Thus, equal iteration budgets do not
imply equal evaluator work. These are candidate--task counts, not counts
of all internal trials or objective-function calls. Cascaded evaluation,
compilation reuse, batching, and candidate runtimes further affect
execution cost.

During shared evolution, per-task scores are cached.
\emph{STA Warmstart} can reuse them when initializing task-local archives,
and \emph{STA Best-Local} can select candidates from their cached
target-task scores. Reusing cached scores can avoid redundant evaluations;
any rescoring during archive projection adds evaluator work but no new
candidate-generation iterations.

We do not assume that LLM generation dominates evaluation cost, nor that
matched iterations imply matched LLM-call counts, token usage, monetary
cost, or wall-clock time. The reported comparisons establish solution
quality under an iteration-matched protocol; assessing total-cost
efficiency would additionally require measuring these resources.

\subsection{ARC-AGI Case Study Details}
\label{app:arc_setting}

Using the ARC-AGI example from the OpenEvolve repository as our implementation base~\citep{openevolve}, we evaluate EMO-STA on two model-specific sets of failed single-task ARC runs from the ARC Prize 2025 ARC-AGI evaluation split, selecting the first 20 failed run--task instances in evaluation-split order for Gemini-3.1-Pro-Preview and 20 for Claude Opus 4.6. This appendix describes how we construct transformed task families, match budgets, and diagnose overfitting in the single-task baseline.

\textbf{Task-family construction.}
For each base ARC task, we construct a five-task augmented family consisting of the original task, a \(90^\circ\) rotation, a left-right flip, an up-down flip, and a fixed color permutation, following~\citet{akyürek2025surprising}. Each transformation is applied consistently to every input grid and every target output grid, so the surface representation changes while the underlying reasoning rule is preserved up to the corresponding transformation. For transformed variants, exact-match evaluation is performed in the transformed task representation. Thus, the augmented family provides related task diversity without using the held-out test output for optimization or introducing task-specific solution templates.

\textbf{Color permutation.}
ARC colors are integer labels from 0 to 9, so a color transformation can be implemented as a permutation of these labels. We use the fixed remapping
\begin{lstlisting}[style=pythoncode]
permutation = np.array([0, 2, 3, 4, 5, 6, 7, 8, 9, 1], dtype=np.int64)
return permutation[np.asarray(grid, dtype=np.int64)]
\end{lstlisting}
This leaves color 0 unchanged, which is useful because ARC often uses 0 as the background color, and cyclically remaps the remaining colors. Since the same remapping is applied to both inputs and outputs, the task logic is unchanged; only the color names differ.

\textbf{Evolution objective and OpenEvolve configuration.}
All ARC runs use diff-based program evolution with seed 42 and either Gemini-3.1-Pro-Preview or Claude Opus 4.6 as the primary model. Programs are optimized using ARC train-set pass@2: each program emits two candidate output grids for each training input, and an example is counted as solved if either candidate exactly matches the target grid. Held-out test grids are used only for evaluation and overfitting diagnosis, not for optimization. Unless otherwise specified, the evolution configuration uses population size 60, archive size 30, four islands, MAP-Elites features over score and diversity with 10 bins, migration every 15 iterations, and migration rate 0.15. The exact run configurations are included in the supplementary code.

\textbf{Iteration-budget matching.}
For each augmented ARC family, EMO-STA first runs a shared evolutionary phase for 60 iterations across all five variants, then specializes separately to each variant for 15 iterations. The direct single-task baseline evolves each variant independently for 27 iterations. This matches the family-level iteration budget:
\[
60 + 5 \times 15 = 5 \times 27 = 135.
\]
Thus, the EMO-STA and single-task comparisons use the same total number of evolution iterations across the five related variants, not the same number of task-level evaluations or the same total cost. We evaluate \emph{STA Warmstart}, \emph{STA Best-Shared}, and \emph{STA Best-Local} under this same matched iteration budget; the main-text solve rates are measured on the original base tasks after adaptation, using their held-out test inputs.

\textbf{Failed-case selection and overfitting diagnosis.}
For each primary model, we first run the standard single-task evolution baseline on tasks from the ARC Prize 2025 ARC-AGI evaluation split using the same train-set pass@2 objective. A run is marked as failed if its best evolved program does not solve the held-out test input. Because evolutionary search is stochastic, failure is defined at the run level rather than as a permanent property of a task; when needed, we repeat the same evaluation protocol to obtain enough failed runs. We then select the first 20 failed run--task instances per model in evaluation-split order, separately for Gemini-3.1-Pro-Preview and Claude Opus 4.6, and use the corresponding base tasks for the augmented multi-task experiments. To diagnose overfitting, we inspect whether the best program from each failed run nevertheless solves all provided training examples under pass@2. When this happens, we classify the failure as overfitting: the program fits the sparse training pairs but fails to generalize to the held-out test grid.

The exact sampled failed run--task instances are listed in \Cref{tab:arc-failed-case-ids}.

\begin{table}[!h]
\centering
\scriptsize
\setlength{\tabcolsep}{5pt}
\renewcommand{\arraystretch}{1.05}
\begin{tabular}{cc@{\qquad}cc}
\toprule
\multicolumn{2}{c}{Gemini-3.1-Pro-Preview}
&
\multicolumn{2}{c}{Claude Opus 4.6} \\
\cmidrule(r){1-2}
\cmidrule(l){3-4}
Dataset index & ARC task ID & Dataset index & ARC task ID \\
\midrule
0   & \texttt{0934a4d8} & 0   & \texttt{0934a4d8} \\
5   & \texttt{16b78196} & 5   & \texttt{16b78196} \\
8   & \texttt{195c6913} & 6   & \texttt{16de56c4} \\
9   & \texttt{1ae2feb7} & 9   & \texttt{1ae2feb7} \\
11  & \texttt{20a9e565} & 10  & \texttt{20270e3b} \\
14  & \texttt{247ef758} & 11  & \texttt{20a9e565} \\
18  & \texttt{291dc1e1} & 14  & \texttt{247ef758} \\
41  & \texttt{581f7754} & 18  & \texttt{291dc1e1} \\
46  & \texttt{62593bfd} & 28  & \texttt{3a25b0d8} \\
52  & \texttt{6ffbe589} & 38  & \texttt{4e34c42c} \\
63  & \texttt{800d221b} & 41  & \texttt{581f7754} \\
67  & \texttt{88e364bc} & 58  & \texttt{7b3084d4} \\
69  & \texttt{898e7135} & 69  & \texttt{898e7135} \\
70  & \texttt{8b7bacbf} & 70  & \texttt{8b7bacbf} \\
90  & \texttt{b5ca7ac4} & 82  & \texttt{a32d8b75} \\
98  & \texttt{cbebaa4b} & 85  & \texttt{a6f40cea} \\
103 & \texttt{db0c5428} & 90  & \texttt{b5ca7ac4} \\
106 & \texttt{dd6b8c4b} & 103 & \texttt{db0c5428} \\
110 & \texttt{e3721c99} & 108 & \texttt{dfadab01} \\
115 & \texttt{eee78d87} & 110 & \texttt{e3721c99} \\
\bottomrule
\end{tabular}
\vspace{0.4em}
\caption{Selected failed ARC run--task instances used in the ARC-AGI case study. Each entry is reported as the task's position in the ARC Prize 2025 ARC-AGI evaluation split and its ARC task ID. The selected cases are the first 20 observed failures for each model in evaluation-split order; rows are paired only for compact presentation and do not imply correspondence between the two model-specific sets.}
\label{tab:arc-failed-case-ids}
\end{table}

\subsection{COVID Deaths Time-Series Case Study}
\label{app:covid-deaths-case-study}

We use the EFE-Time time-series feature engineering
setting~\citep{taga2026evolutionaryfeatureengineeringstructured}
to study overfitting in evolutionary optimization for data-science tasks. EFE-Time evolves
time-series preprocessing programs rather than forecasting models. Each candidate implements
\texttt{fit}, \texttt{transform}, and \texttt{inverse\_transform}: the program is fit on
the available history, transforms the input series, passes the transformed sequence to a
fixed Chronos-2 forecaster~\citep{ansari2025chronos}, and maps the forecast back to the
original scale before evaluation. The evolutionary objective is therefore downstream
forecasting performance, which makes this setting a natural stress test for overfitting:
with short or noisy series, evolution can select transformations that exploit validation
windows without improving held-out forecasts.

The dataset is GIFT-Eval's COVID Deaths task, a daily healthcare benchmark with 266
univariate series and a 30-step prediction horizon~\citep{aksu2024giftevalbenchmarkgeneraltime}.
The target is confirmed COVID-19 deaths. These series are heterogeneous across countries
and noisy due to underreporting, reporting delays, day-of-week effects, and differences in
reporting conventions. Thus, a transformation evolved separately
for one series may fit local reporting artifacts rather than reusable forecasting structure.

In our EMO formulation, each COVID Deaths series is treated as one task. The single-task
baseline runs EFE-Time independently for each series, evolving a separate transformation
from scratch. STA Best-Shared instead performs one shared evolution over the full family of
266 series, selects the transformation with the best family-level objective, and applies it
directly to each series with no task-local adaptation. This corresponds to the native
EFE-Time no-adaptation setting~\citep{taga2026evolutionaryfeatureengineeringstructured}. To match evolutionary-iteration budgets, we run each single-task evolution for 5 iterations, giving $266 \times 5 = 1330$ total iterations, and run the shared STA Best-Shared evolution for 1330 iterations. This matches candidate-generation iterations only: each shared candidate is evaluated across all 266 series, whereas each single-task candidate is evaluated on one series. The comparison therefore does not equalize forecasting-evaluation workload or total computational cost.

The results in Figure~\ref{fig:covid-deaths} indicate that the single-task baseline
overfits. It improves validation MASE and wQL by 12.32\% and 12.99\%, respectively, but
these validation gains largely vanish on held-out test windows: test MASE worsens by
1.12\% and test wQL improves by only 2.71\%. STA Best-Shared without adaptation obtains
smaller or comparable validation gains but much stronger test gains, improving test MASE
by 13.24\% and test wQL by 32.56\%. This pattern supports the regularization view of
shared evolution: because the selected transformation must perform well across many noisy,
heterogeneous series, it is less likely to encode idiosyncratic artifacts from any single
series. In preliminary runs, adding task-local adaptation after the shared phase could
reintroduce overfitting on this dataset, so we report the no-adaptation STA Best-Shared
variant for this case study.

\begingroup

\subsection{K-module Task Relatedness Experiment Details}
\label{app:k-module-relatedness}

\textbf{Task construction.} We construct four K-module tasks, denoted \(A,B,C,D\), each with six modules and six options per module. A candidate returns one fixed configuration, and its task score is the fraction of module choices matching that task's hidden target. We vary only the overlap among the four target configurations while preserving the search space, scoring rule, per-task score distribution, and optimal score.

In the \emph{related} condition, two modules have the same target option across all four tasks; for the remaining four modules, the target is shared by task pairs \(AB\), \(BC\), \(CD\), and \(DA\), respectively. In the \emph{balanced} condition, each module has one agreeing task pair, with each of the six task pairs agreeing on exactly one module. In the \emph{conflicting} condition, all four tasks have different target options at every module. Exhaustive evaluation of all \(6^6=46{,}656\) configurations gives mean pairwise Pearson score correlations of \(\rho=1/3\), \(0\), and \(-1/5\) for the related, balanced, and conflicting conditions, respectively.

The balanced condition matches the target-overlap structure of the main K-module benchmark. The relatedness sweep is a separate experiment using \emph{STA Best-Local} and the model and reporting protocol below.

To avoid artifacts from particular module positions or option identities, we randomly permute module positions and option labels for each instance; the resulting hidden targets remain fixed throughout search. Shared and Single-task searches use the same initial configuration, chosen to score zero on every task.

\textbf{Search protocol.} Using Qwen3.8-27B~{\citep{qwen38_27b}}, \emph{STA Best-Local} performs 40 shared iterations maximizing the mean task score, followed by 20 adaptation iterations per task initialized from its highest-scoring shared candidate. Single-task search uses 30 iterations per task, matching the family-level evolutionary-iteration budget: \(40+4\times20=4\times30=120\). This matches iterations, not task-level evaluation counts. Hidden targets are inaccessible to the model, and search receives only scalar score feedback.

\textbf{Reporting.} For each method and condition, we average the four task scores within each run, then compute the mean and sample standard deviation across three runs. In Figure~\ref{fig:k-module-relatedness}, diamonds and error bars report these statistics for EMO-STA. 
\par
\endgroup

\subsection{Scope and Limitations}
\label{app:limitations}

EMO-STA is designed for task families that can be expressed through a shared executable interface, and is therefore most appropriate when tasks share meaningful program structure. Our experiments focus on such integrable families, so future work could extend the evaluation to larger task collections, more general forms of task relatedness, and automated criteria for deciding when shared evolution should be applied.

\section{Prompts and Program Examples}
\label{app:prompts-and-examples}

\subsection{Sample Prompts for Shared Program Interfaces}
\label{app:shared-interface-prompts}

We include representative prompts illustrating how each task family is framed around a common entry-point function while encouraging reusable solutions that transfer across subtasks.

\begin{lstlisting}[
  style=prompttext,
  basicstyle=\ttfamily\footnotesize,
  columns=fullflexible,
  keepspaces=true,
  frame=single,
  breaklines=true,
  caption={Example system prompt for the EMO-STA circle-packing task family.},
  label={lst:circle-packing-prompt}
]
You are an expert mathematician specializing in circle packing and computational geometry.

Your task is to improve a generic constructor/optimizer for packing n circles of varying radii inside the unit square [0,1] x [0,1], maximizing the sum of their radii.

The evolving code must preserve these exact function signatures:

def construct_packing(n: int):
    ...
    return centers, radii, sum_radii

def run_packing(n: int):
    return construct_packing(n)

The evaluator will call your code for multiple related tasks with:
n in {20, 22, 24, 26}

Public reference target sums used for scoring:
- n=20: 2.301
- n=22: 2.420
- n=24: 2.530
- n=26: 2.635

Constraints:
- centers must have shape (n, 2)
- radii must have shape (n,)
- all circles must lie fully inside the unit square
- circles must not overlap
- keep all outputs finite
- keep the algorithm deterministic

Guidance:
- Favor reusable geometric construction and optimization routines parameterized by n
- Avoid implementing four completely separate hardcoded exact layouts as the main strategy
- Avoid implementing a lookup table of exact layouts only for the seen n values
- Explicit constructive geometry, symmetry-breaking, ring/grid hybrids, corner utilization, and local continuous optimization are all reasonable
- Computing radii from fixed centers is acceptable; jointly optimizing centers and radii is also acceptable
- SciPy-based constrained optimization is allowed if it remains reliable enough for the timeout budget
- Good initial placements matter
- Edge effects are important in a square container
- Variable-sized circles are likely beneficial
- Focus on valid packings first, then improve sum of radii
- Good solutions should remain reasonable for nearby unseen values of n

Do not use plotting code inside the evolved block.
Do not print debugging output from the evolved block.
Write all improvements only between # EVOLVE-BLOCK-START and # EVOLVE-BLOCK-END.
\end{lstlisting}

\begin{lstlisting}[
  style=prompttext,
  basicstyle=\ttfamily\footnotesize,
  columns=fullflexible,
  keepspaces=true,
  frame=single,
  breaklines=true,
  caption={Example system prompt for the EMO-STA signal-processing task family.},
  label={lst:signal-processing-prompt}
]
You are improving a generic causal 1D signal-processing / denoising
algorithm for a multi-task shared-then-specialize workflow.

Important requirements:
- Preserve the function signatures exactly:
  - process_signal(noisy_signal, window_size=20)
  - run_signal_processing(noisy_signal, window_size=20)
- The same evolving program must work across multiple signal families, not
  just one.
- Do not hardcode any single benchmark family, task ID, signal formula, or
  task-specific constants beyond generic windowed processing.
- The evaluator passes the actual noisy signal into the candidate; the
  algorithm must filter that direct input instead of regenerating data.
- Focus on causal smoothing, trend preservation, low lag, bounded memory,
  deterministic behavior, and stable runtime.
- Useful generic ideas include weighted moving averages, Savitzky-Golay
  style local fitting, adaptive smoothing, robust local regression, and
  trend-aware smoothing, but no single method is required.
- Use only Python, NumPy, and SciPy.
\end{lstlisting}

\begin{lstlisting}[
  style=prompttext,
  basicstyle=\ttfamily\footnotesize,
  columns=fullflexible,
  keepspaces=true,
  frame=single,
  breaklines=true,
  caption={Example system prompt for the EMO-STA Heilbronn-triangle task family.},
  label={lst:heilbronn-triangle-prompt}
]
You are an expert mathematician specializing in combinatorial and continuous geometry.

Your task is to improve a generic constructor/optimizer for the Heilbronn triangle problem in a unit-area triangle.

The evolving code must preserve these exact function signatures:

def construct_points(n: int):
    ...
    return points, min_area

def run_heilbronn(n: int):
    return construct_points(n)

The evaluator uses one fixed canonical unit-area triangle:
A = (0.0, 0.0)
B = (2.0, 0.0)
C = (0.0, 1.0)

A point (x, y) is inside the triangle iff:
- x >= 0
- y >= 0
- x / 2 + y <= 1

The evaluator will call your code for multiple related tasks with:
n in {9, 10, 11, 12}

Public pinned scoring anchors:
- n=9: 0.0548469387755102
- n=10: 0.04337673349889024
- n=11: 0.03609267801015405
- n=12: 0.03100478174352528

Constraints:
- points must have shape (n, 2)
- all points must be finite
- all points must lie on or inside the canonical triangle
- keep the algorithm deterministic

Guidance:
- Favor reusable geometric construction and optimization routines parameterized by n
- Avoid implementing four completely separate hardcoded exact configurations as the main strategy
- Boundary-heavy configurations, symmetry, barycentric parameterizations, deterministic local improvement, and maximizing the worst triple area are all reasonable ideas
- Duplicates or collinear triples are feasible but will produce zero score, so try to avoid them
- Because all unit-area triangles are affinely equivalent, it is acceptable to optimize only in this canonical triangle
- Focus on valid point sets first, then improve the minimum triangle area

Do not use plotting code inside the evolved block.
Do not print debugging output from the evolved block.
Write all improvements only between # EVOLVE-BLOCK-START and # EVOLVE-BLOCK-END.
\end{lstlisting}

\begin{lstlisting}[
  style=prompttext,
  basicstyle=\ttfamily\footnotesize,
  columns=fullflexible,
  keepspaces=true,
  frame=single,
  breaklines=true,
  caption={Example system prompt for the EMO-STA circle-packing rectangle task family.},
  label={lst:circle-packing-rectangle-prompt}
]
You are an expert mathematician specializing in circle packing and computational geometry.

Your task is to improve a generic constructor/optimizer for packing n circles of varying radii inside a rectangle of perimeter 4, maximizing the sum of their radii.

The evolving code must preserve these exact function signatures:

def construct_packing(n: int):
    ...
    return centers, radii, alpha, sum_radii

def run_packing(n: int):
    return construct_packing(n)

In this family:
- alpha is the rectangle width
- rectangle height is 2 - alpha
- alpha must satisfy 0 < alpha <= 1
- the evaluator will call your code for multiple related tasks with:
  n in {20, 21, 22, 23}

Pinned public scoring anchors:
- n=20: 2.305
- n=21: 2.365
- n=22: 2.425
- n=23: 2.484

Constraints:
- centers must have shape (n, 2)
- radii must have shape (n,)
- all circles must lie fully inside the rectangle [0, alpha] x [0, 2 - alpha]
- circles must not overlap
- keep all outputs finite
- keep the algorithm deterministic

Guidance:
- Favor reusable geometric construction and optimization routines parameterized by n
- The rectangle aspect ratio is also part of the search through alpha
- Avoid implementing four completely separate hardcoded exact layouts as the main strategy
- Explicit constructive geometry, symmetry-breaking, corner/edge utilization, ring/grid hybrids, and local continuous optimization are all reasonable
- Computing radii from fixed centers is acceptable; jointly optimizing centers, radii, and alpha is also acceptable
- SciPy-based constrained or unconstrained optimization is allowed if it remains reliable enough for the timeout budget
- Good initial placements matter
- The shorter side is alpha, so radii cannot exceed alpha / 2
- Focus on valid packings first, then improve sum of radii

Do not use plotting code inside the evolved block.
Do not print debugging output from the evolved block.
Write all improvements only between # EVOLVE-BLOCK-START and # EVOLVE-BLOCK-END.
\end{lstlisting}

\subsection{Illustrative EMO-STA Program Examples}
\label{app:emo-sta-program-examples}

These examples are intended to illustrate how EMO-STA changes program structure, not to present minimal or globally optimal implementations. We selected circle packing and signal processing because they show two distinct adaptation modes: numerical/geometric refinement and behavioral trade-off adjustment. In both cases, adaptation does not discard the shared solution and start over; it retains the main computational template discovered during shared evolution, then reallocates search effort, modifies heuristics, or tunes decision thresholds for the target task.

Table~\ref{tab:emosta-program-examples} summarizes the two examples. We compare the program obtained after shared evolution (\emph{STA Best-Shared (Before Adaptation)}) with the final program obtained after \emph{STA Best-Local} adaptation, emphasizing qualitative changes in representation and search behavior in addition to the final score. The single-task score is reported as a reference average over
the five independent single-task runs from the same iteration-matched setting.

\begin{table}[!h]
\centering
\small
\setlength{\tabcolsep}{5pt}
\begin{tabular}{llcccc}
\toprule
\multirow{2}{*}{Family}
& \multirow{2}{*}{Task}
& \multirow{2}{*}{Model}
& \multirow{2}{*}{Budget}
& \multicolumn{1}{c}{STA Best-Shared}
& \multirow{2}{*}{STA Best-Local} \\
&
&
&
& \multicolumn{1}{c}{(Before Adaptation)}
& \\
\midrule
Circle packing
& $n=24$
& \texttt{claude-opus-4-6}
& $60/15/120$
& $0.9643$
& $\mathbf{0.9995}$ \\
Signal processing
& Step changes
& \texttt{claude-opus-4-6}
& $60/10/100$
& $0.6938$
& $\mathbf{0.8834}$ \\
\bottomrule
\end{tabular}
\vspace{4pt}
\caption{Representative program-level examples of EMO-STA adaptation. Budgets
are reported as Shared / Adapt / Total iterations, where Total is the matched
family-level iteration budget \(S + K A = K B\). The corresponding per-task single-task
baseline budgets are \(B=30\) for circle packing and \(B=25\) for signal
processing. The STA Best-Shared (Before Adaptation) column reports the best
shared program evaluated before target-specific adaptation.}
\label{tab:emosta-program-examples}
\end{table}

\noindent\textbf{Circle packing.}
The circle-packing example uses \texttt{claude-opus-4-6} with a $60/15/120$
Shared / Adapt / Total budget on the $n=24$ task, corresponding to a matched
single-task baseline budget of \(B=30\) per task. The shared program already
captures a useful family-level idea: circle packing should be treated as a
geometric optimization problem rather than as only a fixed constructive layout.
It proposes plausible geometric arrangements, relaxes center positions, assigns
feasible radii through constrained optimization, and then improves the layout
through local search and restarts. This shared solution scores $0.9643$ on
$n=24$, comparable to the five-run single-task average of $0.9567 \pm 0.0194$.
Best-Local adaptation keeps this same algorithmic representation, but makes the
search more specialized for the target circle count. The adapted program adds
more diverse layout proposals and spends more of its computation jointly
adjusting center positions and radii. The score rises to $0.9995$. This is the
intended EMO-STA behavior: shared evolution discovers the general solver
structure, while specialization sharpens that solver for one target instance.

\begin{lstlisting}[style=pythoncode,caption={Selected code snippets from the circle-packing shared and adapted programs.},label={lst:circle-packing-program-example}]
# Shared program: hexagonal layouts, relaxation, LP radii, local search.

import numpy as np
from scipy.optimize import linprog
from scipy.spatial.distance import pdist, squareform
import time

def compute_max_radii_lp(centers):
    n = len(centers)
    c_obj = -np.ones(n)
    dists = squareform(pdist(centers))

    pairs = []
    for i in range(n):
        for j in range(i + 1, n):
            pairs.append((i, j, dists[i, j]))

    A_ub = np.zeros((len(pairs) + n, n))
    b_ub = np.zeros(len(pairs) + n)

    for idx, (i, j, d) in enumerate(pairs):
        A_ub[idx, i] = 1.0
        A_ub[idx, j] = 1.0
        b_ub[idx] = d - 1e-10

    for i in range(n):
        x, y = centers[i]
        A_ub[len(pairs) + i, i] = 1.0
        b_ub[len(pairs) + i] = max(
            min(x, y, 1.0 - x, 1.0 - y), 0.0
        ) - 1e-10

    res = linprog(c_obj, A_ub=A_ub, b_ub=b_ub,
                  bounds=[(0, None)] * n, method="highs")
    if res.success:
        return np.maximum(res.x, 0.0)

    return np.zeros(n)

def gen_hex(n, cols, rows, shift=0.0):
    dx = 1.0 / (cols + 1)
    dy = 1.0 / (rows + 1)
    pts = []
    for r in range(rows):
        off = (0.5 * dx if r % 2 else 0.0) + shift
        for c in range(cols):
            if len(pts) >= n:
                break
            pts.append([
                np.clip(dx * (c + 1) + off, 0.01, 0.99),
                np.clip(dy * (r + 1), 0.01, 0.99),
            ])
    return np.array(pts[:n]) if len(pts) >= n else None

def construct_packing(n: int):
    n = int(n)
    t0 = time.time()
    best_c, best_s = None, -1.0

    sq = int(np.ceil(np.sqrt(n)))
    for cols in range(max(2, sq - 2), min(n + 1, sq + 6)):
        rows = int(np.ceil(n / cols))
        for shift in [0.0, 0.02, -0.02]:
            centers = gen_hex(n, cols, rows, shift)
            if centers is None:
                continue
            centers = force_relax(centers, n, iters=80)
            score = float(np.sum(compute_max_radii_lp(centers)))
            if score > best_s:
                best_s, best_c = score, centers.copy()

    best_c, best_s = optimize_cyclic(
        best_c, n, t0, 18, steps=[0.04, 0.02, 0.01]
    )

    rng = np.random.RandomState(42)
    for it in range(50):
        if time.time() - t0 > 46:
            break
        scale = max(0.003, 0.09 * (0.78 ** it))
        trial = np.clip(best_c + rng.randn(n, 2) * scale, 0.005, 0.995)
        trial = force_relax(trial, n, iters=40)
        trial, score = optimize_cyclic(
            trial, n, t0, min(time.time() - t0 + 3, 46),
            steps=[0.02, 0.008],
        )
        if score > best_s:
            best_s, best_c = score, trial.copy()

    radii = compute_max_radii_lp(best_c)
    return best_c, radii, float(np.sum(radii))


# Adapted program: same solver structure, plus joint SLSQP refinement
# and more diverse layout proposals.

from scipy.optimize import minimize
from scipy.sparse import csr_matrix

def joint_constraints(x, n):
    centers = x[:2 * n].reshape(n, 2)
    radii = x[2 * n:]
    cons = []

    for i in range(n):
        cons.append(centers[i, 0] - radii[i])
        cons.append(centers[i, 1] - radii[i])
        cons.append(1.0 - centers[i, 0] - radii[i])
        cons.append(1.0 - centers[i, 1] - radii[i])

    for i in range(n):
        for j in range(i + 1, n):
            d = np.linalg.norm(centers[i] - centers[j])
            cons.append(d - radii[i] - radii[j])

    return np.array(cons)

def joint_optimize(centers, n, maxiter=300):
    radii = compute_max_radii_lp(centers)
    x0 = np.concatenate([centers.flatten(), radii])
    bounds = [(0.005, 0.995)] * (2 * n) + [(1e-4, 0.5)] * n

    res = minimize(
        lambda x, nn: -np.sum(x[2 * nn:]),
        x0,
        args=(n,),
        method="SLSQP",
        bounds=bounds,
        constraints={"type": "ineq", "fun": joint_constraints, "args": (n,)},
        options={"maxiter": maxiter, "ftol": 1e-12},
    )

    if np.all(joint_constraints(res.x, n) >= -1e-6):
        centers_new = res.x[:2 * n].reshape(n, 2)
        radii_new = np.maximum(res.x[2 * n:], 0.0)
        return centers_new, radii_new, float(np.sum(radii_new))

    return centers, radii, float(np.sum(radii))

def gen_sunflower(n):
    golden = (1.0 + np.sqrt(5.0)) / 2.0
    pts = []
    for i in range(n):
        r = np.sqrt((i + 0.5) / n) * 0.45
        theta = 2.0 * np.pi * i / golden**2
        pts.append([
            np.clip(0.5 + r * np.cos(theta), 0.01, 0.99),
            np.clip(0.5 + r * np.sin(theta), 0.01, 0.99),
        ])
    return np.array(pts)

def construct_packing(n: int):
    n = int(n)
    t0 = time.time()
    candidates = []

    sq = int(np.ceil(np.sqrt(n)))
    for cols in range(max(2, sq - 2), min(n + 1, sq + 4)):
        rows = int(np.ceil(n / cols))
        centers = gen_hex_grid(n, cols, rows)
        if centers is not None:
            candidates.append(centers)

    candidates.append(gen_sunflower(n))
    optimized_grid = gen_optimized_grid(n)
    if optimized_grid is not None:
        candidates.append(optimized_grid)

    best_c, best_s = None, -1.0
    for centers in candidates:
        centers = force_relax(centers, n, iters=80)
        score = float(np.sum(compute_max_radii_lp(centers)))
        if score > best_s:
            best_s, best_c = score, centers.copy()

    best_c, best_r, best_s = joint_optimize(best_c, n, maxiter=500)
    best_c, best_s = optimize_cyclic(
        best_c, n, t0, 15, steps=[0.04, 0.02, 0.01]
    )
    best_c, best_r, best_s = joint_optimize(best_c, n, maxiter=800)

    if time.time() - t0 < 58:
        best_c, best_s = optimize_cyclic(
            best_c, n, t0, 59, steps=[0.0003, 0.0001]
        )

    radii = compute_max_radii_lp(best_c)
    return best_c, radii, float(np.sum(radii))
\end{lstlisting}

\noindent\textbf{Signal processing.}
The signal-processing example uses \texttt{claude-opus-4-6} with a $60/10/100$
Shared / Adapt / Total budget on the step-change task. The shared program
learns a general causal filtering strategy: estimate local structure from a
sliding window, combine a trend-following estimate with a smoother estimate, and
then apply adaptive smoothing over time. This kind of program is useful across
the signal-processing family because the tasks contain different mixtures of
noise, oscillation, trend, and abrupt change. The shared program scores
$0.6938$ on the step-change task. Best-Local adaptation keeps the same
multi-stage causal-filtering structure, but changes the behavior of the filter
for the target task. In particular, it becomes more conservative about reacting
to small fluctuations while still allowing larger changes to pass through. This
raises the score to $0.8834$, above the five-run single-task average of
$0.7373 \pm 0.0653$. This example shows a different kind of specialization from
circle packing: rather than refining a geometric optimizer, adaptation refines
the behavioral tradeoff encoded in the algorithm.

\begin{lstlisting}[style=pythoncode,caption={Selected code snippets from the signal-processing shared and adapted programs.},label={lst:signal-processing-program-example}]
# Shared program: local trend estimation plus adaptive smoothing.

import numpy as np
from numpy.lib.stride_tricks import sliding_window_view

def process_signal(noisy_signal, window_size):
    signal = np.asarray(noisy_signal, dtype=float)
    w = max(3, int(window_size))
    n = len(signal)

    if n < w:
        return signal.copy()

    t = np.arange(w, dtype=float)
    t_end = float(w - 1)
    windows = sliding_window_view(signal, w)

    sigma = w * 0.45
    weights = np.exp(-0.5 * ((t - t_end) / sigma) ** 2)

    X_quad = np.column_stack([np.ones(w), t, t * t])
    WX_quad = X_quad * weights[:, None]
    kernel_quad = np.array([1.0, t_end, t_end * t_end]) @ np.linalg.solve(
        WX_quad.T @ X_quad + 1e-10 * np.eye(3),
        WX_quad.T,
    )

    X_lin = np.column_stack([np.ones(w), t])
    WX_lin = X_lin * weights[:, None]
    kernel_lin = np.array([1.0, t_end]) @ np.linalg.solve(
        WX_lin.T @ X_lin + 1e-10 * np.eye(2),
        WX_lin.T,
    )

    local_quad = windows @ kernel_quad
    local_lin = windows @ kernel_lin

    noise_est = np.median(np.abs(np.diff(signal))) * 1.4826 + 1e-12
    curvature = np.abs(local_quad - local_lin)
    blend = np.clip(curvature / (3.0 * noise_est), 0.0, 1.0)

    filtered = blend * local_quad + (1.0 - blend) * local_lin

    for alpha_base, threshold, sharpness in [
        (0.03, 1.2, 5.0),
        (0.04, 0.5, 6.0),
        (0.06, 0.35, 7.0),
    ]:
        smoothed = np.empty_like(filtered)
        smoothed[0] = filtered[0]
        for i in range(1, len(filtered)):
            g = abs(filtered[i] - smoothed[i - 1]) / noise_est
            alpha = alpha_base + (1.0 - alpha_base) / (
                1.0 + np.exp(-sharpness * (g - threshold))
            )
            smoothed[i] = alpha * filtered[i] + (1.0 - alpha) * smoothed[i - 1]
        filtered = smoothed

    return filtered


# Adapted program: same filtering template, but more conservative around
# small fluctuations and more selective about when to react.

def process_signal(noisy_signal, window_size):
    signal = np.asarray(noisy_signal, dtype=float)
    w = max(3, int(window_size))
    n = len(signal)

    if n < w:
        return signal.copy()

    t = np.arange(w, dtype=float)
    t_end = float(w - 1)
    windows = sliding_window_view(signal, w)

    noise_est = (
        np.median(np.abs(np.diff(signal))) * 1.4826 / np.sqrt(2.0) + 1e-12
    )

    sigma = w * 0.4
    weights = np.exp(-0.5 * ((t - t_end) / sigma) ** 2)

    X_lin = np.column_stack([np.ones(w), t])
    WX_lin = X_lin * weights[:, None]
    kernel_lin = np.array([1.0, t_end]) @ np.linalg.solve(
        WX_lin.T @ X_lin + 1e-10 * np.eye(2),
        WX_lin.T,
    )

    local_lin = windows @ kernel_lin
    local_mean = windows @ (weights / weights.sum())

    slope_mag = np.abs(local_lin - local_mean)
    blend = np.clip(slope_mag / (2.0 * noise_est), 0.0, 1.0)
    filtered = blend * local_lin + (1.0 - blend) * local_mean

    dead_zone = 0.3 * noise_est

    for alpha_base, threshold, sharpness, zone_scale in [
        (0.02, 1.5, 6.0, 1.0),
        (0.02, 0.8, 8.0, 0.5),
        (0.03, 0.5, 10.0, 0.3),
    ]:
        smoothed = np.empty_like(filtered)
        smoothed[0] = filtered[0]

        for i in range(1, len(filtered)):
            diff = filtered[i] - smoothed[i - 1]
            g = abs(diff) / noise_est
            alpha = alpha_base + (1.0 - alpha_base) / (
                1.0 + np.exp(-sharpness * (g - threshold))
            )

            if abs(diff) < dead_zone * zone_scale:
                alpha = min(alpha, alpha_base * 0.5)

            smoothed[i] = alpha * filtered[i] + (1.0 - alpha) * smoothed[i - 1]

        filtered = smoothed

    return filtered
\end{lstlisting}

\section{Further Related Work}\label{sec:further_related}

\begingroup

\noindent \textbf{LLM-based automatic heuristic design.}
Beyond the methods discussed in the main text, recent work has proposed several complementary improvements to LLM-based automatic heuristic design. ReEvo uses reflective feedback to guide heuristic evolution, MCTS-AHD replaces population-based evolution with tree-structured search, LLaMEA-HPO combines LLM-generated algorithm structures with numerical hyperparameter optimization, and LHNS uses LLM-guided neighborhood search~\citep{reevo,mctsahd,llameahpo,lhns}. Other systems modify the broader search or generator dynamics: ThetaEvolve studies test-time learning for open optimization problems, DeltaEvolve represents program edits through semantic deltas, AdaEvolve adapts exploration during evolution, and EvoX meta-evolves the search strategy~\citep{wang2025thetaevolve,jiang2026deltaevolve,cemri2026adaevolve,liu2026evox}. These methods primarily improve how candidate algorithms are generated, searched, or refined, and are largely complementary to EMO-STA's shared-then-adapt organization across related task objectives.
\par
\endgroup

\noindent \textbf{Multi-task learning and shared representations.}
Multi-task learning studies how related tasks can improve one another by sharing an inductive bias, representation, or set of parameters~\citep{caruana1997multitask,baxter2000model,argyriou2008convex,ruder2017overview,zhang2022survey}. Classical and modern formulations often use a shared representation with task-specific predictors. A recurring theme in this literature is that sharing is beneficial only when tasks are sufficiently aligned: task relatedness, interference, competition, and grouping can determine whether a shared representation improves individual tasks or causes negative transfer~\citep{ruder2017overview,zhang2022survey,pmlr-v238-ildiz24a,li2025identifying}. EMO-STA shares the same broad motivation of leveraging common structure across related tasks, but transfers executable programs rather than learned feature representations: it evolves a shared program archive and uses it to initialize task-local evolutionary search.

\noindent \textbf{Transfer in Bayesian optimization.}
Bayesian optimization offers an adjacent perspective on transfer across related black-box optimization problems. Multi-task BO transfers observations through a multi-task Gaussian-process surrogate~\citep{NIPS2013_f33ba15e}, while warm-start BO models related optimization problems jointly~\citep{poloczek2016warmstartingbayesianoptimization}. Other work studies practical transfer learning for Bayesian optimization~\citep{feurer2022practicaltransferlearningbayesian}. The distinction for EMO-STA is the object of transfer: it reuses executable program candidates to initialize task-local evolution, rather than sharing information through a surrogate model.

\noindent \textbf{Adaptation and test-time learning.}
EMO-STA also connects to the broader adaptation literature, where a general model, representation, or procedure is adapted to a target task, domain, or data distribution. In our setting, adaptation occurs at the level of executable programs: after shared evolution identifies reusable candidates, task-local evolution adapts them under each target task's evaluator. Recent work on LLM adaptation studies related goals through model-side updates, data- and domain-specific adaptation, parameter-efficient fine-tuning, retrieval- or tool-augmented adaptation, and real-time updating mechanisms~\citep{ke2025adaptation,prottasha2025peft}. Test-time adaptation is another instance of this line of work, where adaptation happens during evaluation after target inputs or target-domain data become available; recent surveys organize such methods by whether they modify model parameters, inference procedures, normalization statistics, samples, or prompts~\citep{xiao2024modeladaptationtesttime}. In LLMs and foundation models, recent work explores test-time training for in-context learning and ARC-style reasoning~\citep{akyürek2025surprising,gozeten2026testtime}, test-time learning from unlabeled target-domain inputs~\citep{hu2025testtime}, and test-time learning for open optimization problems through evolving program databases and reinforcement learning~\citep{wang2025thetaevolve}.

%%%%%%%%%%%%%%%%%%%%%%%%%%%%%%%%%%%%%%%%%%%%%%%%%%%%%%%%%%%%

\newpage

\end{document}